\documentclass[runningheads]{llncs}

\usepackage[T1]{fontenc}
\usepackage{graphicx}
\usepackage{comment}
\usepackage{amsmath,amssymb}
\usepackage{color}
\usepackage{url}
\usepackage{hyperref}

\usepackage{bm}
\usepackage{acro}
\usepackage{xspace}
\usepackage{array}
\usepackage{booktabs}
\usepackage{fontawesome}
\usepackage{xcolor,colortbl}
\definecolor{Gray}{gray}{0.9}
\definecolor{Textgray}{gray}{0.4}
\definecolor{notetext}{rgb}{0.7,0,0}
\definecolor{notetext}{rgb}{0.7,0,0}

\usepackage[nointegrals]{wasysym}
\usepackage{makecell}
\usepackage{balance}
\usepackage{multirow}
\usepackage{overpic}
\usepackage{bbm}
\usepackage{dsfont}
\usepackage[font=small,labelfont=bf]{caption}
\usepackage{color}
\usepackage{pifont}
\usepackage{etoolbox}
\usepackage{float}
\usepackage{tikz}
\usepackage{svg}
\usepackage{times}
\usepackage{epsfig}
\usepackage{graphicx}
\usepackage{amsmath}
\usepackage{amssymb}
\usepackage{comment}
\usepackage{soul}
\usepackage{listings}
\usepackage{lscape}
\usepackage{rotating}
\usepackage{enumitem}
\usepackage[utf8]{inputenc}
\usepackage{pgfplots}
\usepackage{import}
\usepackage{xifthen}
\usepackage{pdfpages}
\usepackage{transparent}
\usepackage{tabularx}
\usepackage{tabu}
\usepackage{xcolor}
\usepackage{arydshln}
\DeclareUnicodeCharacter{2212}{−}
\usepgfplotslibrary{groupplots,dateplot}
\usetikzlibrary{patterns,shapes.arrows}
\pgfplotsset{compat=newest}

\usepackage{subcaption}
\usepackage{xfrac}
\usepackage{tcolorbox}

\usepackage{svg}
\newcommand{\orcid}[1]{\href{https://orcid.org/#1}{\includegraphics[width=10pt]{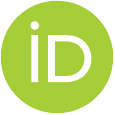}}}

\definecolor{LightCyan}{rgb}{0.87,0.92,0.96}
\definecolor{m_green}{RGB}{212, 251, 121}
\definecolor{m_darkgreen}{HTML}{d5e9d5}
\definecolor{m_green}{HTML}{d5e9d5}
\definecolor{m_orange}{HTML}{fff2cd}
\definecolor{m_red}{HTML}{fccecd}
\definecolor{m_violet}{HTML}{e2d6e8}
\definecolor{m_blue}{HTML}{d8e9fc}

\newcommand{\reffig}[1]{Fig.\,\ref{fig:#1}}
\newcommand{\reftab}[1]{Tab.\,\ref{tab:#1}}

\lstdefinestyle{mypython}{
  language=python,
  breaklines=true,
  basicstyle=\fontsize{8}{12}\selectfont\ttfamily,
  keywordstyle=\bfseries\color{my_blue},
  linewidth=.99\textwidth,
}

\newcommand{\colorsquare}[1]{{\color{#1}$\blacksquare$}\hspace{-7.78pt}$\square$}

\newcommand{\parag}[1]{\vskip8pt \noindent \textbf{#1}}
\newcommand{\PAR}[1]{\vskip4pt \noindent {\bf #1~}}
\newcommand{\PARbegin}[1]{\noindent {\bf #1~}}

\usepackage[capitalize]{cleveref}
\crefname{section}{Sec.}{Secs.}
\Crefname{section}{Section}{Sections}
\Crefname{table}{Table}{Tables}
\crefname{table}{Tab.}{Tabs.}

\definecolor{darkgreen}{RGB}{0,255,0}
\definecolor{linkgreen}{RGB}{52,130,48}

\newcommand{\cmark}{\ding{51}}%
\newcommand{\xmark}{\ding{55}}%

\newcolumntype{Y}{>{\centering\arraybackslash}X}
\newcolumntype{Z}{>{\raggedleft\arraybackslash}X}

\newcommand*{\eg}{\emph{e.g.}\@\xspace}
\newcommand*{\ie}{\emph{i.e.}\@\xspace}

\newcommand*{\etal}{\emph{et al.}\@\xspace}

\newif\ifmynotes
\mynotestrue
\definecolor{notetext}{rgb}{0.7,0,0}

\newcommand\ArrowDown[1]{
\hspace{-12px}\rotatebox[origin=c]{270}{$\curvearrowright$}{\hspace{2px}#1}
}

\makeatletter
\def\adl@drawiv#1#2#3{%
        \hskip.5\tabcolsep
        \xleaders#3{#2.5\@tempdimb #1{1}#2.5\@tempdimb}%
                #2\z@ plus1fil minus1fil\relax
        \hskip.5\tabcolsep}
\newcommand{\cdashlinelr}[1]{%
  \noalign{\vskip\aboverulesep
           \global\let\@dashdrawstore\adl@draw
           \global\let\adl@draw\adl@drawiv}
  \cdashline{#1}
  \noalign{\global\let\adl@draw\@dashdrawstore
           \vskip\belowrulesep}}
\makeatother

\newif\ifreview
\reviewfalse

\ifreview
	\usepackage{lineno}

	\linenumbers
\fi

\begin{document}

\def\SubNumber{}

\def\GCPRTrack{}

\title{UGainS: Uncertainty Guided Anomaly Instance Segmentation}

\ifreview
\else
    \author{Alexey Nekrasov\inst{1}\orcid{0000-0002-7230-0294} \and
Alexander Hermans\inst{1}\orcid{0000-0003-2127-0782} \and
Lars Kuhnert\inst{2}\orcid{0009-0008-4341-957X} \and
Bastian Leibe\inst{1}\orcid{0000-0003-4225-0051}}

\authorrunning{A. Nekrasov et al.}

\institute{Visual Computing Institute, RWTH Aachen University, Aachen, Germany
\email{nekrasov@vision.rwth-aachen.de} \and
Ford Motor Company, Dearborn, Michigan, USA}

\fi

\maketitle              %

\begin{abstract}
    A single unexpected object on the road can cause an accident or may lead to injuries.
    To prevent this, we need a reliable mechanism for finding anomalous objects on the road.
    This task, called anomaly segmentation, can be a stepping stone to safe and reliable autonomous driving.
    Current approaches tackle anomaly segmentation by assigning an anomaly score to each pixel and by grouping anomalous regions using simple heuristics.
    However, pixel grouping is a limiting factor when it comes to evaluating the segmentation performance of individual anomalous objects.
    To address the issue of grouping multiple anomaly instances into one, we propose an approach that produces accurate anomaly instance masks.
    Our approach centers on an out-of-distribution segmentation model for identifying uncertain regions and a strong generalist segmentation model for anomaly instances segmentation.
    We investigate ways to use uncertain regions to guide such a segmentation model to perform segmentation of anomalous instances.
    By incorporating strong object priors from a generalist model we additionally improve the per-pixel anomaly segmentation performance.
    Our approach outperforms current pixel-level anomaly segmentation methods, achieving an AP of $80.08\%$ and $88.98\%$ on the Fishyscapes Lost and Found and the RoadAnomaly validation sets respectively.
    Project page: \url{https://vision.rwth-aachen.de/ugains}
\end{abstract}

\section{Introduction}
\label{sec:intro}
Current semantic segmentation approaches~\cite{cheng2022maskedattention,xie2021segformer,chen2018encoderdecoder} achieve impressive performance for known classes on driving datasets.
However, in the context of autonomous driving, the real world is more complex than the test scenarios captured in today's benchmarks~\cite{valdenegro-toro2021find}. 
In particular, a wide variety of objects, \eg wild animals or debris on the road, can pose a threat to vehicles or passengers when encountered on the road. 
Since individual objects may occur very rarely, learning to reliably detect such potentially dangerous objects in the long tail of the data distribution is a major challenge.
Even datasets with millions of images may only contain few frames with anomalous objects~\cite{li2022coda}.
Since modern roads are built according to well-defined rules, it creates a model bias towards regular road context.
To make matters worse, deep networks tend to make overconfident predictions and often fail on out-of-distribution cases~\cite{guo2017calibration,kendall2017what}.
Together, rare occurrences of anomalous instances and model overconfidence, creates a need to address the problem of detecting anomalous road objects.

Most anomaly segmentation approaches focus on predicting a per-pixel outlier score to create an uncertainty map of the visible scene.
Available approaches in this category are based on classification uncertainty \cite{lakshminarayanan2017simple,hendrycks2018baseline,hendrycks2019deep}, reconstruction errors \cite{xia2020synthesize,lis2019detecting}, or they rely on data augmentation with synthetic outliers to regularize models \cite{grcic2022densehybrid,chan2021entropy,bevandic2019simultaneous}. 

In this paper, we propose an approach that addresses the anomaly instance segmentation problem.
Our approach uses uncertainty maps to detect image structures that are not well captured by the given training data~\cite{nayal2023rba}.
However, we then use the extracted outlier pixels as point guidance for a class-agnostic segmentation pipeline~\cite{kirillov2023segment} to segment anomalous instances.
\begin{figure}[t]
  \centering
  \setlength{\tabcolsep}{0pt}
  \newcommand{\figcrop}[1]{\includegraphics[trim={5cm 0cm 0cm 0cm},clip, width=0.246\textwidth]{figures/teaser_qualitative/#1.jpg}}
  \begin{tabularx}{\textwidth}{YYYY}%
    \figcrop{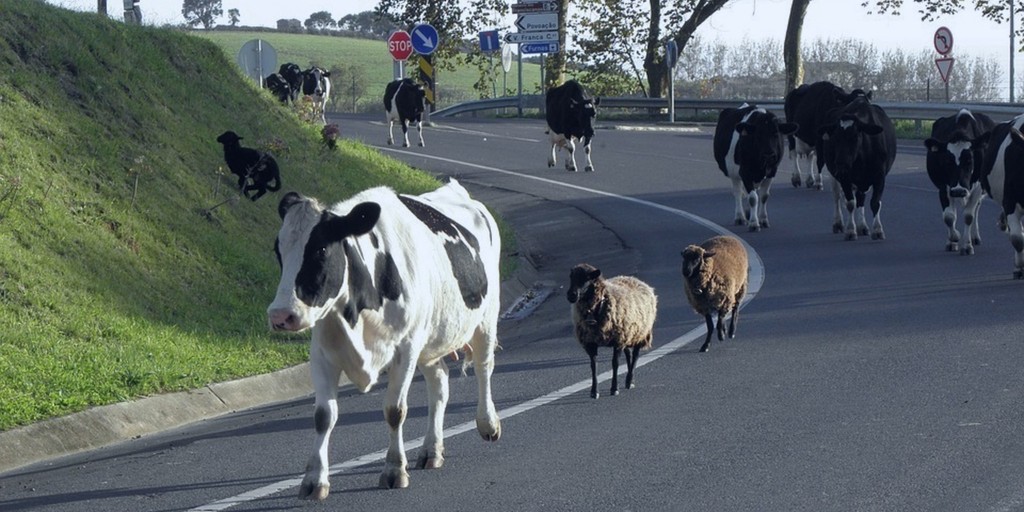}&\figcrop{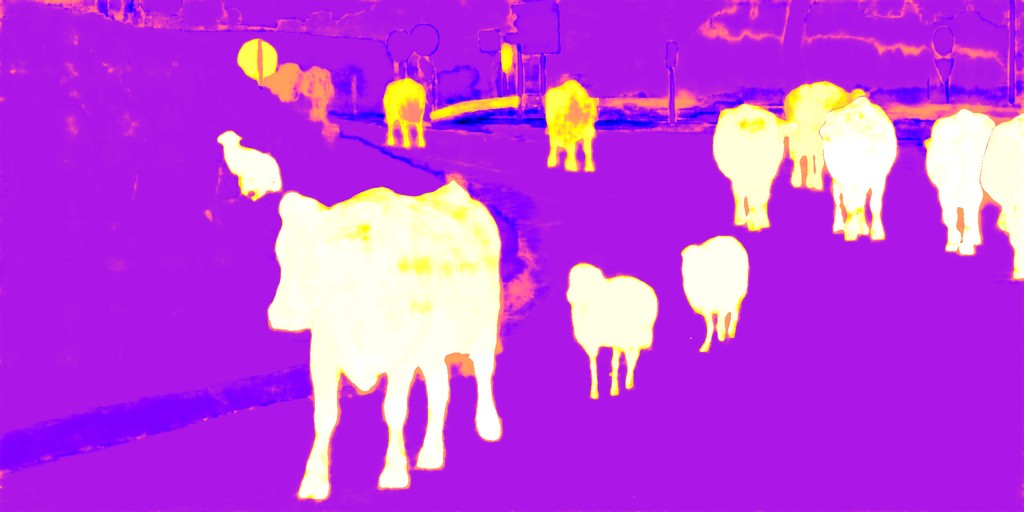}&
  \figcrop{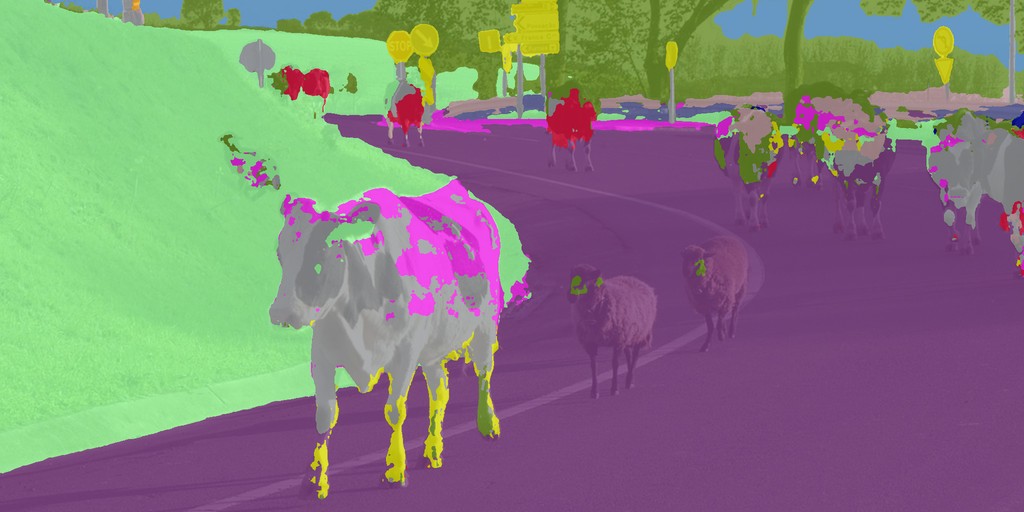}&\figcrop{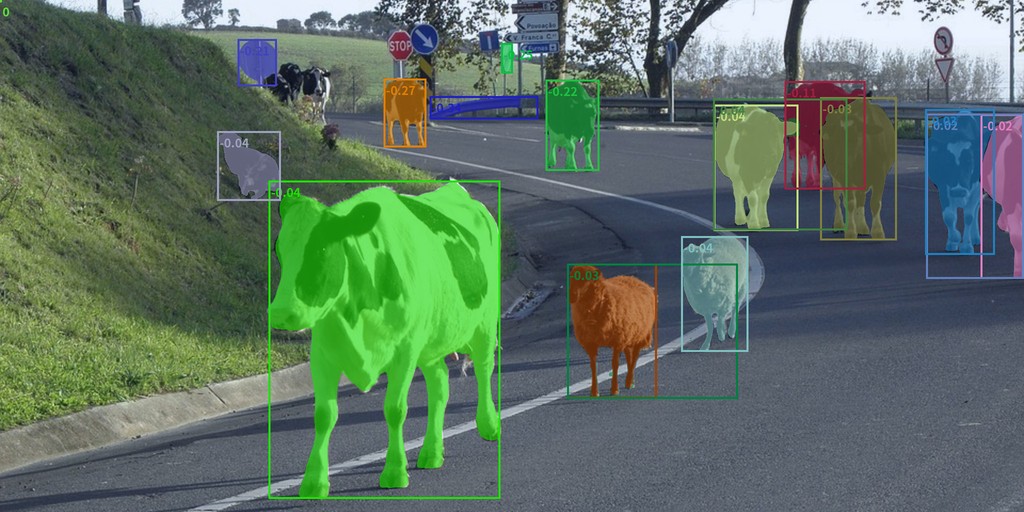}\\[-5pt]%
  \scriptsize RGB input & \scriptsize Uncertainty & \scriptsize Semantic Segmentation & \scriptsize Anomaly Instances
  \end{tabularx}
  \caption{
      \textbf{Instance prediction of anomalous objects.}
      We propose UGainS, an approach for instance segmentation of anomalous objects.
      UGainS combines a strong generalist segmentation model (SAM \cite{kirillov2023segment}), and a strong out-of-distribution method (RbA \cite{nayal2023rba}).
      This allows us to predict an anomaly instance segmentation, refined uncertainty maps as well as an in-distribution semantic segmentation.
  }
  \label{fig:teaser}
\end{figure}

The resulting out-of-distribution instance segmentations (see \reffig{teaser}) can then be used for mining rare objects~\cite{jiang2022improving}, which could be added to the training set to improve future model performance.
When using such a point-guidance scheme, the key challenge is to decide how to sample and prioritize the guidance points in order to achieve a good coverage of the relevant anomalous regions with as few proposals as possible.
We systematically explore this design space and present a detailed experimental evaluation of the different design choices.

Current anomaly segmentation benchmarks evaluate methods based on per-pixel or component-level metrics~\cite{chan2021segmentmeifyoucan}.
Per-pixel metrics make it possible to evaluate performance under large class imbalance, but they favor large objects. %
Component-level metrics are a proxy for object-level segmentation and evaluate the intersection between predicted and ground truth anomaly regions.
To generate object instances, benchmarks often use the connected components algorithm in the metric formulation, which is a proxy for instance-level performance \cite{chan2021segmentmeifyoucan}. 
However, current benchmarks do not provide a clear way to evaluate instance prediction performance within anomaly regions and focus on finding the entire region.
We argue that finding individual instances within large anomalous regions can be beneficial for downstream tasks such as object mining~\cite{li2022coda}, tracking~\cite{liu2022opening}, or model introspection. %
Therefore we propose an anomaly instance segmentation evaluation measure (\emph{instance AP} or \emph{iAP}) and present experimental results using this measure, in addition to also evaluating on the standard per-pixel measures. 
Our proposed approach achieves state-of-the-art per-pixel performance on the RoadAnomaly and Fishyscapes Lost and Found benchmark datasets.
In addition, we demonstrate the utility of the extracted anomaly instances by visualizing cases where previous methods would not be able to recover separate anomalies.

\PAR{Our contributions} are summarized as follows:
1) We propose UGainS, a novel approach for predicting anomalous object instances based on point-guided segmentation with points sampled from an epistemic uncertainty map.
2) We evaluate the design choices of performing uncertainty guidance for a generalist segmentation model and provide detailed ablations to highlight the influence of the different steps of our approach.
3) UGainS achieves state-of-the-art anomaly segmentation performance on the Fishyscapes Lost and Found~\cite{blum2021fishyscapes} validation set with $80.08\%$ pixel-level average precision, and on the RoadAnomaly~\cite{lis2019detecting} dataset with $88.98\%$ pixel-level average precision. Notably, UGainS improves per-pixel performance compared with previous anomaly segmentation methods, while at the same time extracting interpretable instance segmentation masks.

\section{Related Work}

\subsection{Pixel-Wise Anomaly Segmentation}
The anomaly segmentation task has attracted attention in recent years, and is derived from the out-of-distribution classification task.
Early approaches used probabilistic modeling directly to obtain outlier scores and did not require any special training procedures.
These methods assume that anomalies should have high uncertainty values.
Methods have approximated the epistemic uncertainty of predictions using deep ensembles~\cite{lakshminarayanan2017simple}, Monte-Carlo dropout~\cite{mukhoti2019evaluating}, or directly using SoftMax probabilities~\cite{hendrycks2018baseline}.
Liang~\etal~\cite{liang2022gmmseg} recently proposed to use Gaussian Mixture Models in a segmentation model to capture class conditional densities. 
In another line of work, methods focus on directly improving the predicted outlier scores~\cite{chan2021entropy,jung2021standardized}.

Previous studies have relied on embedding space estimation~\cite{lee2018simple,blum2021fishyscapes} or direct classification of anomalous regions~\cite{blum2021fishyscapes}.
Reconstruction-based methods rely on re-synthesis of input images~\cite{lis2019detecting,lis2021detecting,dibiase2021pixelwise,xia2020synthesize,ohgushi2020road} and the assumption that anomalies cannot be reconstructed well.
Other approaches train or fine-tune on images with artificial anomalies \cite{grcic2022densehybrid,grcic2021dense,bevandic2019simultaneous,hendrycks2018baseline,chan2021segmentmeifyoucan}. 
Following early methods derived from classification \cite{liang2018enhancing}, recent methods focus on using adversarial examples on an unmodified in-distribution model~\cite{lee2018simple,besnier2021triggering}.
While the majority of previous approaches use per-pixel prediction models~\cite{chen2018encoderdecoder}, the most recent approach, Rejected by All (RbA), focuses on acquiring anomaly scores from a set of predicted mask-class pairs~\cite{nayal2023rba}.
It assumes that pixels which are not covered by any masks correspond to anomalous objects.
We follow this line of work and extend it to instance segmentation.

\subsection{Anomaly Instance Segmentation}
While the majority of anomaly segmentation approaches focus on a per-pixel anomaly classification, and thus ignore any instance information, several recent methods have attempted to address anomaly instance segmentation.
MergeNet~\cite{gupta2018mergenet} uses two networks to predict anomalous regions.
LiDAR-guided small object segmentation focuses on retrieving small objects \cite{singh2020lidar}.
Xue \etal \cite{xue2020tiny} uses occlusion edges to represent anomalous instances and a sliding window approach to generate object proposals.
These methods use low-level heuristics, based on connected components or over-segmentation, to retrieve anomalous instances.
The SegmentMeIfYouCan \cite{chan2021segmentmeifyoucan} benchmark suggests using component-level metrics to evaluate object segmentation performance, grouping anomalous pixels with connected components and evaluating component-level intersection over union.
However, grouping multiple spatially close objects may hinder downstream tasks such as tracking, data collection, or model introspection. %
Treating multiple connected instances as a single object may furthermore be limiting for future advances in anomaly instance segmentation.

A few works tackle the problem of finding anomalous instances without such heuristics.
OSIS~\cite{wong2019identifying} is an open-set instance segmentation work on 3D point clouds (which was adapted to images by Gasperini~\etal~\cite{gasperini2022holistic}).
It learns class-agnostic spatial instance embeddings that are clustered to predict an object. %
A more recent work, EOPSN~\cite{hwang2021exemplarbased}, relies on existing unknown instances in the unannotated regions during training, which is a limiting factor for autonomous driving setups as training sets are often close to densely annotated.
The concurrent Holistic Segmentation U3HS approach \cite{gasperini2022holistic} also uses uncertainty scores to obtain anomaly instance segmentations, however U3HS uses explicit clustering of learned instance features.
In contrast, we use an uncertainty guided point sampling scheme with a generalist instance segmentation model to provide anomaly instance segmentation, which achieves a stronger instance segmentation performance in practice.

\subsection{Promptable Segmentation Models}
Recent advances in large-scale text-guided training for classification~\cite{jia2021scaling,radford2021learning} are sparking interest in large-scale open-vocabulary and open-world segmentation~\cite{ding2023openvocabulary,rao2022denseclip,xu2022groupvit}.
XDecoder~\cite{zou2022generalized} addresses multiple segmentation tasks by combining the Mask2Former decoder~\cite{cheng2022maskedattention} with text supervision.
A recent extension, SEEM~\cite{zou2023segment}, extends the work to open-set segmentation using multiple prompt types.
Our method uses the recent Segment Anything Model (SAM)~\cite{kirillov2023segment}.
Kirillov~\etal propose the large-scale segmentation dataset SA-1B with class-agnostic mask annotations and SAM for interactive segmentation.
SAM accepts prompts of different types, such as points, masks, boxes, or text.
Several other works extend SAM to medical image segmentation~\cite{ma2023segment,wu2023medical}, image inpainting~\cite{yu2023inpaint}, camouflaged object detection~\cite{tang2023can}, industrial anomaly segmentation~\cite{cao2023segment}, and other applications~\cite{ji2023segment}.
The base SAM model is limited to prompted prediction and lacks query interaction for separate objects. 
As such it is difficult to use for general instance segmentation with a fixed set of classes.
By first predicting general outlier regions, we can sample points and utilize SAM for anomaly instance segmentation though.

\section{Method}
\label{sec:method}
\begin{figure}[t]
  \centering
  \includegraphics[trim={0.1cm 0.7cm 0.8cm 0.3cm},clip,width=\textwidth]{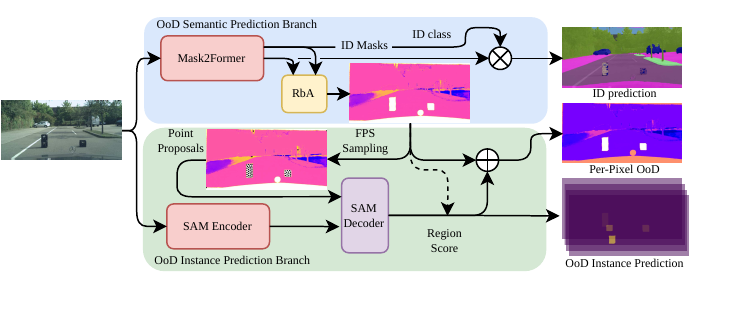}
  \caption{
      \textbf{Method Overview.}
      UGainS consists of two main blocks: an OoD Semantic Prediction Branch (\colorsquare{m_blue}) and OoD Instance Prediction Branch (\colorsquare{m_darkgreen}).
      We use the Mask2Former model \cite{cheng2022maskedattention} to get class and mask predictions for in-distribution (ID) classes.
      From these predictions we compute a per-pixel uncertainty map using RbA \cite{nayal2023rba} .
      Using farthest point sampling (FPS) on the uncertainty map we generate point proposals for anomalous objects.
      These point proposals are then used as input for the Segment Anything Model (SAM) \cite{kirillov2023segment} to get anomaly instance predictions.
      The final prediction is an in-distribution semantic segmentation, instance-refined per-pixel anomaly scores, and a set of out-of-distribution anomaly instance masks.
  }
  \label{fig:main}
\end{figure}

UGainS combines a strong in-distribution segmentation model with a strong generalist segmentation model (see \reffig{main}).
At the core of our method is a Mask2Former~\cite{cheng2022maskedattention} regularized to avoid predicting objects in unknown regions~\cite{nayal2023rba}, and the Segment Anything Model (SAM)~\cite{kirillov2023segment}.
The semantic branch predicts masks for known semantic classes, such as cars or pedestrians, and avoids prediction in regions occupied by out-of-distribution objects.
Based on the Reject by All (RbA) score definition~\cite{nayal2023rba} these unknown regions will have higher uncertainty values.
We obtain point proposals for the out-of-distribution branch by sampling points in these regions.
For each sampled point, the Segment Anything Model generates an instance mask.
We average the anomaly scores under the mask regions to obtain the final prediction.
To this end, we propose to tackle the problem of anomaly segmentation as segmenting \emph{everything} we know and prompting to segment \emph{everywhere} where we do not know.
In this section, we provide a detailed description of each component of our approach.

\PAR{OoD Semantic Prediction Branch.}
With the out-of-distribution (OoD) semantic prediction branch we predict in-distribution (ID) semantic segmentation and OoD per-pixel scores (see \reffig{main}, \colorsquare{m_blue}).

\label{semantic}
Here we use the RbA adaptation of the Mask2Former model.
The Mask2Former model consists of a backbone with a pixel decoder and a transformer decoder.
For an image $\bm{x}$, the Swin-B~\cite{liu2021swin} backbone produces latent features at a coarse scale, and in combination with the pixel decoder, produces an $L$-layer feature pyramid $\{f_i(\bm{x})\}_{i=1}^L$.
The transformer decoder processes these multi-resolution $C$-dimensional features together with the $N$ learned object queries $\mathbf{Q}_{0} \in \mathbb{R}^{N \times C}$.
The object queries are sequentially updated by cross-attending the multi-resolution features in a round-robin fashion.
The final set of object queries $\mathbf{Q}_{i}$ is then multiplied with the high-resolution feature map $f_L(\bm{x})$, generating $\bm{\mathcal{M}} \in \mathbb{R}^{N \times H \times W}$ binary masks for an image of the size $H \times W$.
In addition, class probabilities $\bm{\mathcal{L}} \in \mathbb{R}^{N \times (K + 1)}$ are predicted from queries, where $(K+1)$ are closed-set classes with an additional class to signal a query deactivation.
During training we use the RbA approach, which regularizes Mask2Former to predict no in-distribution class or mask in an anomalous region.
RbA uses a per-pixel negative sum of mask predictions to score anomalous regions, where every pixel in a mask is assigned a sum of predicted class posterior probabilities:
\[
\bm{U}(\bm{x}) = - \sum_{k=1}^{K} \sum_{n=1}^{N} \bm{\mathcal{L}}_{k,n}(\bm{x}) \cdot \bm{\mathcal{M}}_{k,n}(\bm{x})
\label{eq:rba}
.\]
This results in per-pixel RbA scores $\bm{U}(\bm{x}) \in [-N, 0]^{H \times W}$, where a low score indicates that a pixel is contained in one or multiple predicted masks for in-distribution classes. %
A high score indicates that the pixel belongs to no masks or all masks have a very low confidence for the pixel.
We use the RbA method because it predicts low anomaly scores on semantic boundaries, \ie, road-to-sidewalk boundaries (\reffig{teaser}), and since it generally has a strong anomaly segmentation performance.
However, in principle we could use any model that predicts a per-pixel anomaly score.
For a detailed overview of Mask2Former and RbA we refer the interested reader to the respective papers~\cite{cheng2022maskedattention,nayal2023rba}.

\PAR{OoD Instance Prediction Branch.} 
To get instance predictions we use an instance prediction branch that is based on semantic outlier scores  (see \reffig{main}, \colorsquare{m_darkgreen}).
A naive approach to obtain instances would be to apply the connected components algorithm on a binary mask from outlier scores.
However, as discussed earlier, this approach does not work when objects have an overlap or an intersection with known classes (see \reffig{concon}).
We address this problem with a novel approach using a segmentation model promptable with point proposals.
To convert the per-pixel anomaly predictions into instances, we prompt SAM  to predict instance masks for a set of sampled anomalous points.
However, there is no obvious way to create a prompt from a noisy and partially incorrect anomaly mask.
While we could use all anomaly points as prompts, this would result in increased computational complexity and diminishing results (see Sec.~\ref{sec:ablations}).
We therefore resort to farthest point sampling (FPS) which we run on the 2D coordinates of the thresholded anomaly points.
(Ablations on hyperparameters and other sampling strategies can be found in Section~\ref{sec:ablations}.)

We then prompt SAM with the resulting anomaly point proposals to predict instance masks.
SAM consists of an image encoder, a prompt encoder, and a lightweight mask decoder.
The image encoder is a plain ViT-H~\cite{dosovitskiy2021vit} which extracts latent dense image features.
During the prompting, we obtain a prompt embedding for each sampled coordinate, comprised of a positional and learned embedding.
Finally, the lightweight mask decoder from SAM processes image features together with prompt embeddings to predict object masks.
The mask decoder maps each prompt to a hierarchy of three masks with foreground probabilities.
Based on the predicted foreground probability, the best of the three masks is selected to remove the ambiguity.
To provide each mask with anomaly scores, we use an average of the uncertainty values in the mask region.
After obtaining individual scores, we remove duplicate masks using non-maximum suppression (NMS).

Since we obtain high-quality object masks, we further combine them with the initial per-pixel anomaly predictions.
Similar to the computation of the per-pixel anomaly scores in the semantic branch, we obtain the anomaly scores as a sum of individual scores.
We add these new instance-based anomaly scores to the existing RbA scores to produce new per-pixel scores, which can be seen as a form of RbA ensembling.
Effectively, we improve the per-pixel scores by using strong object priors from the instance predictions.
In summary, UGainS predicts an in-distribution semantic segmentation, a set of out-of-distribution anomaly instance masks, and an improved per-pixel anomaly score.

\section{Experiments}
\label{sec:experiments}

\PARbegin{Datasets.}
We use the Cityscapes dataset \cite{cordts2016cityscapes} for in-distribution training.
We use $2975$ training and $500$ validation images with $19$ labeled classes with instance and semantic labels.

To validate the quality of anomaly predictions we use the Fishyscapes Lost and Found (FS L\&F) \cite{blum2021fishyscapes} and RoadAnomaly \cite{lis2019detecting} datasets.
Fishyscapes L\&F \cite{blum2021fishyscapes} is a filtered and re-annotated version of the Lost and Found dataset \cite{pinggera2016lost}.
The original Lost and Found dataset (L\&F) was created by placing anomalous objects on the road in front of a vehicle.
Lost and Found contains both semantic and instance annotations for anomalous objects, as well as a \textit{road} annotation; it labels the rest as \textit{ignore}.
However, it contains noisy annotations, \eg labeling bicycles and children as anomalies \cite{blum2021fishyscapes}.
Thus, the FS L\&F dataset has become a standard benchmark in the anomaly segmentation community.
Fishyscapes L\&F contains $100$ images in the validation set and $275$ images in the test set.
However, the FS L\&F dataset does not originally contain instance labels. 
To evaluate instance segmentation performance, we re-label objects in the validation set  by applying connected components on semantic annotations.
This works for FS L\&F, since objects in the dataset are far apart.
For comparability with pixel-level evaluation, we only use the newly generated instance for the instance-level evaluation.

The RoadAnomaly~\cite{lis2019detecting} dataset is an earlier version of the RoadAnomaly21 dataset, which is part of the SegementMeIfYouCan~\cite{chan2021segmentmeifyoucan} benchmark.
SegmentMeIfYouCan does not provide annotations for instance masks in the validation set.
The RoadAnomaly dataset contains $60$ images with semantic labels and partial instance annotations, as such we only evaluate pixel-level metrics on this dataset.

\PAR{Dataset-Specific Adjustments.}
The RbA method has strong responses for the car hood and rectification artifacts at the image boundaries, since these areas are not labeled in the Cityscapes dataset.
We create a dataset-specific ignore mask and do not sample points from the masked regions.
While this is a heuristics, we argue such an ignore mask can easily be constructed during a calibration process and does not harm the generality of our method.
Furthermore, common evaluation protocols exclude predictions from ignore regions, so we do not gain an unfair advantage over other methods.
\begin{figure}[t]
  \centering
  \begin{tabular}{cccc}
    \includegraphics[trim={200px 75px 225px 100px},clip,width=0.243\textwidth]{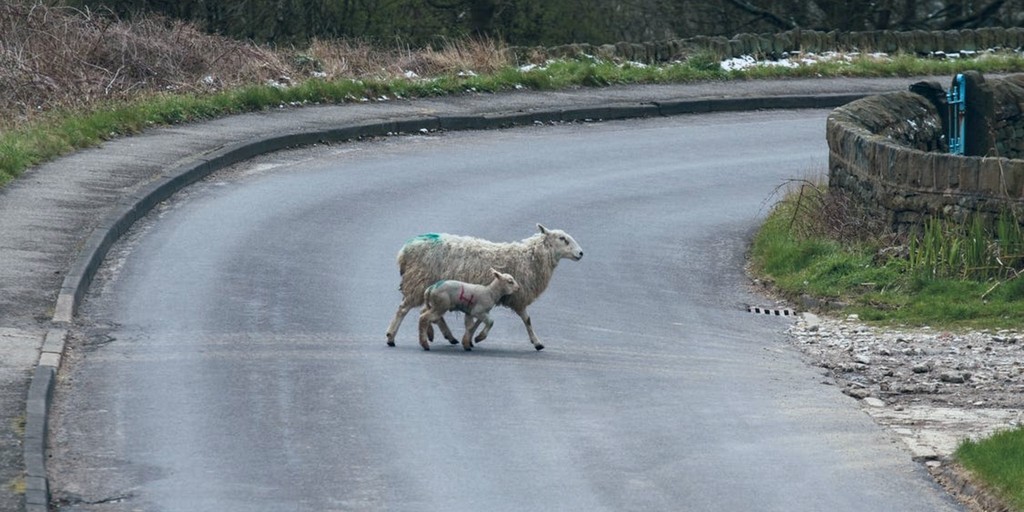}&%
    \includegraphics[trim={200px 75px 225px 100px},clip,width=0.243\textwidth]{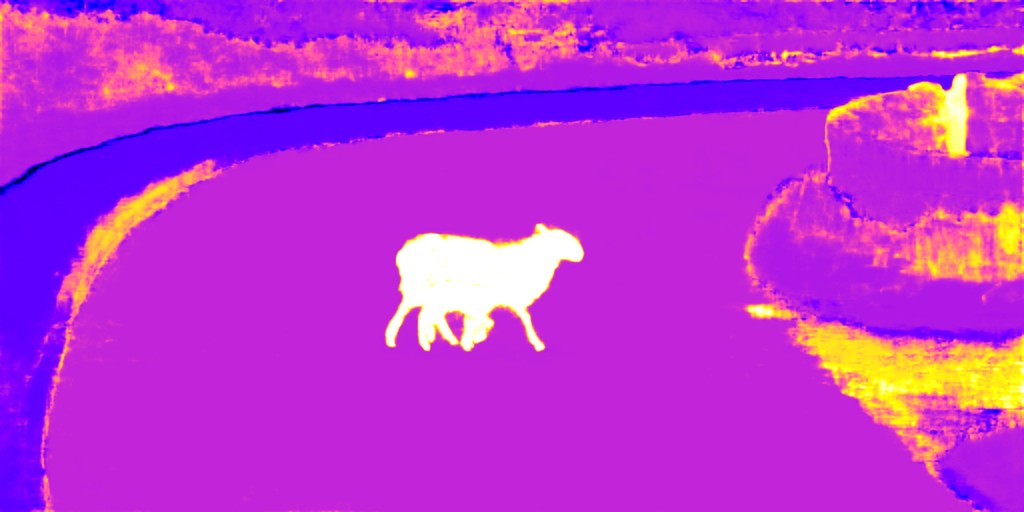}&%
    \includegraphics[trim={200px 75px 225px 100px},clip,width=0.243\textwidth]{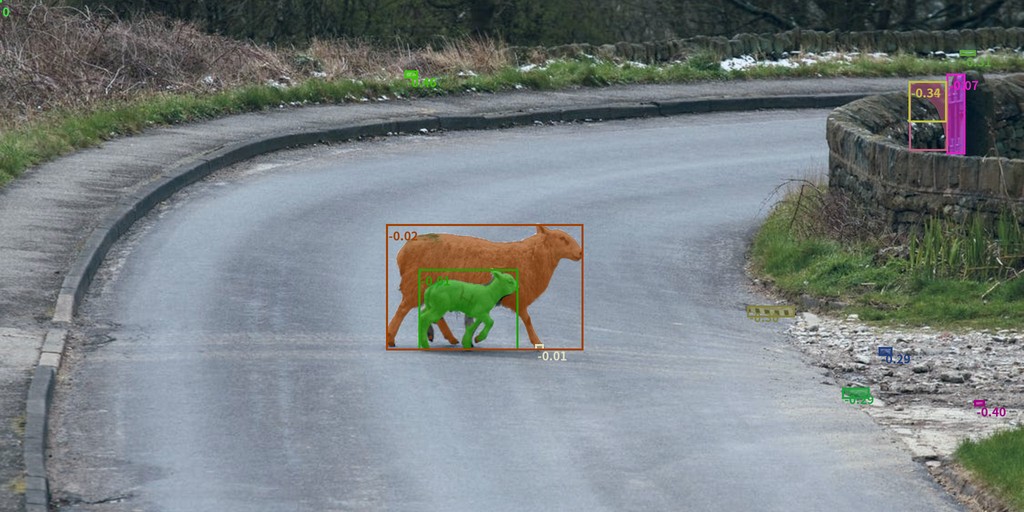}&%
    \includegraphics[trim={200px 75px 225px 100px},clip,width=0.243\textwidth]{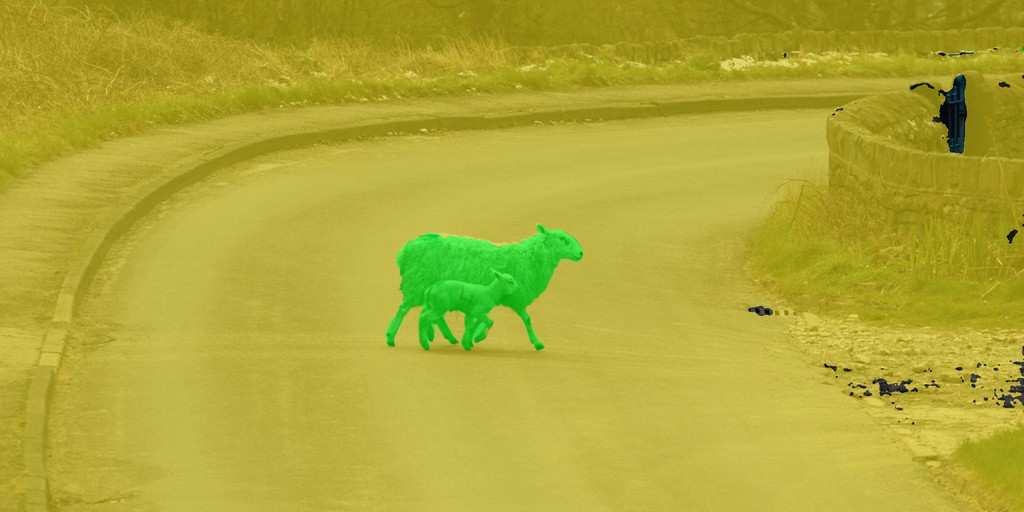}\\[-2pt]%
    \scriptsize Image & \scriptsize Anomaly scores & \scriptsize Our instance prediction & \scriptsize Connected components
  \end{tabular}
  \vspace{-9pt}
  \caption{
      \textbf{Groups of anomalous pixels as instances.}
      Current evaluation of methods do not distinguish object groups and individual objects to evaluate the quality of object retrieval.
      We argue that for purposes such as tracking, data mining, or model introspection, it is important to recover individual instances rather than groups of objects.
      Methods or metrics that rely on connected components are therefore not well suited for instance segmentation in practice.
  }
  \label{fig:concon}
\end{figure}

\PAR{Evaluation Metrics.}
To evaluate the out-of-distribution performance we use the standard community metrics \emph{average precision} (AP) and \emph{false positive rate at $95\%$ true positive rate} (FPR$_{95}$).
To distinguish these pixel-level metrics from our later instance-level metrics, we prepend a p: pAP and pFPR$_{95}$.
Evaluating anomalous instances on the other hand is not trivial.
One approach is to use a component-level metric, such as the sIoU \cite{chan2021segmentmeifyoucan} evaluation protocol.
In the sIoU formulation, the connected components method creates instances from predicted uncertainty maps and then scores them using the intersection-over-union metric.
However, it does not penalize grouping of objects or prediction that overlaps with an \emph{ignore} region. %
We argue that it is still important to segment individual instances for downstream tasks, such as data mining or model introspection.
If an anomalous object appears in front of another anomalous region, the computation of component-level metrics will ignore grouping (see \reffig{concon}).
For example, in the Cityscapes dataset tunnels or bridges are labeled as a \textit{void} class, resulting in high anomaly scores.
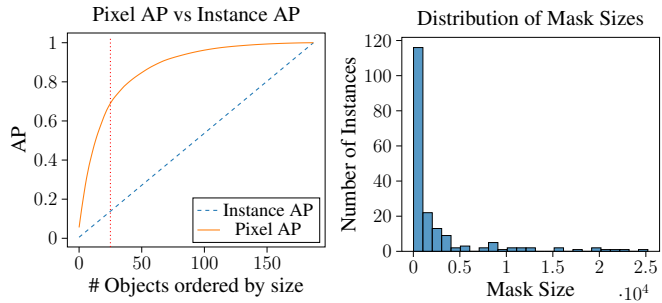
\begin{figure}[t]
  \begin{center}
    \begin{tikzpicture}[thick, scale=0.5, transform shape]

\definecolor{darkgray176}{RGB}{176,176,176}
\definecolor{darkorange25512714}{RGB}{255,127,14}
\definecolor{steelblue31119180}{RGB}{31,119,180}

\begin{axis}[
tick align=outside,
tick pos=left,
title={\fontsize{16}{16}\selectfont Pixel AP vs Instance AP},
x grid style={darkgray176},
xlabel={\fontsize{16}{16}\selectfont \# Objects ordered by size},
xmin=-9.35, xmax=196.35,
xtick style={color=black},
y grid style={darkgray176},
ylabel={\fontsize{16}{16}\selectfont AP},
tick label style={font=\fontsize{14}{14}\selectfont},
ymin=-0.0444148933049291, ymax=1.04973404253833,
ytick style={color=black},
legend style={at={(0.98,0.02)}, anchor=south east, font=\fontsize{14}{14}\selectfont}
]

\addplot [thick, dashed, steelblue31119180]
table {%
0 0.00531914923340082
1 0.0106382984668016
2 0.015957448631525
3 0.0212765969336033
4 0.0265957470983267
5 0.0319148972630501
6 0.0372340455651283
7 0.0425531938672066
8 0.0478723421692848
9 0.0531914941966534
10 0.0585106387734413
11 0.0638297945261002
12 0.0691489353775978
13 0.0744680911302567
14 0.0797872319817543
15 0.0851063877344131
16 0.0904255285859108
17 0.0957446843385696
18 0.101063825190067
19 0.106382988393307
20 0.111702129244804
21 0.117021277546883
22 0.12234041839838
23 0.1276595890522
24 0.132978722453117
25 0.138297870755196
26 0.143617004156113
27 0.148936182260513
28 0.15425531566143
29 0.159574463963509
30 0.164893597364426
31 0.170212775468826
32 0.175531908869743
33 0.180851057171822
34 0.186170190572739
35 0.191489368677139
36 0.196808516979218
37 0.202127650380135
38 0.207446813583374
39 0.212765976786613
40 0.218085095286369
41 0.223404258489609
42 0.228723406791687
43 0.234042555093765
44 0.239361718297005
45 0.244680836796761
46 0.25
47 0.255319178104401
48 0.260638296604156
49 0.265957444906235
50 0.271276593208313
51 0.276595741510391
52 0.281914889812469
53 0.287234008312225
54 0.292553186416626
55 0.297872364521027
56 0.303191483020782
57 0.308510631322861
58 0.313829779624939
59 0.319148927927017
60 0.324468076229095
61 0.329787194728851
62 0.335106372833252
63 0.340425550937653
64 0.345744669437408
65 0.351063817739487
66 0.356382966041565
67 0.361702114343643
68 0.367021262645721
69 0.372340381145477
70 0.377659618854523
71 0.382978737354279
72 0.388297885656357
73 0.393617033958435
74 0.398936182260513
75 0.404255300760269
76 0.409574449062347
77 0.414893627166748
78 0.420212805271149
79 0.425531953573227
80 0.430851072072983
81 0.436170190572739
82 0.441489368677139
83 0.446808516979218
84 0.452127635478973
85 0.457446813583374
86 0.462765991687775
87 0.468085110187531
88 0.473404258489609
89 0.478723436594009
90 0.484042555093765
91 0.489361673593521
92 0.494680821895599
93 0.5
94 0.505319237709045
95 0.510638356208801
96 0.515957474708557
97 0.521276593208313
98 0.526595771312714
99 0.531914889812469
100 0.53723406791687
101 0.542553186416626
102 0.547872364521027
103 0.553191483020782
104 0.558510661125183
105 0.563829779624939
106 0.569148898124695
107 0.574468016624451
108 0.579787254333496
109 0.585106372833252
110 0.590425610542297
111 0.595744729042053
112 0.601063847541809
113 0.606382966041565
114 0.611702144145966
115 0.617021262645721
116 0.622340440750122
117 0.627659559249878
118 0.632978737354279
119 0.638297855854034
120 0.643617033958435
121 0.648936152458191
122 0.654255270957947
123 0.659574389457703
124 0.664893627166748
125 0.670212745666504
126 0.675531983375549
127 0.680851101875305
128 0.686170220375061
129 0.691489338874817
130 0.696808516979218
131 0.702127635478973
132 0.707446813583374
133 0.71276593208313
134 0.718085110187531
135 0.723404228687286
136 0.728723406791687
137 0.734042525291443
138 0.739361643791199
139 0.744680762290955
140 0.75
141 0.755319237709045
142 0.760638356208801
143 0.765957474708557
144 0.771276593208313
145 0.776595771312714
146 0.781914889812469
147 0.78723406791687
148 0.792553186416626
149 0.797872364521027
150 0.803191483020782
151 0.808510601520538
152 0.813829779624939
153 0.819148898124695
154 0.824468016624451
155 0.829787254333496
156 0.835106372833252
157 0.840425610542297
158 0.845744729042053
159 0.851063907146454
160 0.856382966041565
161 0.861702144145966
162 0.867021262645721
163 0.872340381145477
164 0.877659618854523
165 0.882978737354279
166 0.888297855854034
167 0.893617033958435
168 0.898936092853546
169 0.904255270957947
170 0.909574389457703
171 0.914893627166748
172 0.920212745666504
173 0.925531983375549
174 0.930851101875305
175 0.936170220375061
176 0.941489398479462
177 0.946808516979218
178 0.952127635478973
179 0.957446873188019
180 0.962765872478485
181 0.968085110187531
182 0.973404228687286
183 0.978723347187042
184 0.984042525291443
185 0.989361643791199
186 0.994680762290955
187 1
};
\addlegendentry{Instance AP}

\addplot [thick, darkorange25512714]
table {%
0 0.0554462105237343
1 0.102470770599308
2 0.148532919034711
3 0.190956274652723
4 0.23232118623263
5 0.272542059680621
6 0.30891746188958
7 0.341031948355398
8 0.372641220901647
9 0.399607535773841
10 0.425504968303807
11 0.450765664282251
12 0.475137016749387
13 0.497199416344444
14 0.51851860868592
15 0.538021536109938
16 0.556111117155658
17 0.57375811410655
18 0.591071085325494
19 0.608217043678466
20 0.625125008697426
21 0.639984978447389
22 0.654759354103542
23 0.667345861215445
24 0.679241352594385
25 0.690339357538303
26 0.70041649633896
27 0.710151258764371
28 0.718493551419732
29 0.726510168986444
30 0.734011134329465
31 0.741499573707537
32 0.748971311799012
33 0.756248897433794
34 0.763486817512906
35 0.770580688995117
36 0.777545125506199
37 0.783292455756501
38 0.789027260041855
39 0.794711960467417
40 0.800177456506389
41 0.805486377983512
42 0.81063663723796
43 0.815634497252207
44 0.820496659312851
45 0.825223123419892
46 0.829724120157869
47 0.834189626661827
48 0.838592503341045
49 0.842874295692433
50 0.847078844593307
51 0.851277130511708
52 0.855433663213616
53 0.859588108254699
54 0.863654871541145
55 0.867414748686367
56 0.871164187527464
57 0.874901100403614
58 0.87835409140761
59 0.881646332528106
60 0.88492396002283
61 0.888147308336112
62 0.89136856898857
63 0.894397764845159
64 0.897420697719273
65 0.900343422873804
66 0.903239008437614
67 0.906086577802456
68 0.908666926581742
69 0.910971704132172
70 0.913263955717654
71 0.91553115537324
72 0.917637605145326
73 0.919723178309167
74 0.921685579484351
75 0.923597876799744
76 0.925499735811014
77 0.927401594822283
78 0.929174018862424
79 0.930929741615967
80 0.93268546436951
81 0.934437011801404
82 0.936184383911648
83 0.937908791752821
84 0.939581008073378
85 0.941221909481564
86 0.942839846620679
87 0.944405592239178
88 0.94595046124943
89 0.94747654131226
90 0.94899844605344
91 0.950407617110089
92 0.951806349862614
93 0.953198819632666
94 0.954580851098594
95 0.955942005956275
96 0.957213391398496
97 0.958447198945873
98 0.959666392867477
99 0.960885586789081
100 0.962083904102439
101 0.963204977965284
102 0.964280123289986
103 0.965351093293039
104 0.966417887974443
105 0.967461718386775
106 0.96848884751251
107 0.969463785117629
108 0.970384443541306
109 0.971300926643334
110 0.972186094832992
111 0.973062912379351
112 0.973931379282411
113 0.974783144898875
114 0.975545141099877
115 0.976277910049334
116 0.976975188764773
117 0.977668292158561
118 0.978359307891525
119 0.979027359355418
120 0.979674534211064
121 0.980315446084236
122 0.980941744331636
123 0.981511675736769
124 0.982071168837779
125 0.98262648661714
126 0.983179716735676
127 0.983720420889264
128 0.98424651141708
129 0.984766338962421
130 0.985265289899516
131 0.985762153175786
132 0.986252753469583
133 0.986737090780905
134 0.987219340431402
135 0.9877015900819
136 0.988177576749923
137 0.988638949792174
138 0.989087796869477
139 0.989532468625131
140 0.989964614415837
141 0.990373795937471
142 0.990745399564261
143 0.991114915530227
144 0.991465642548771
145 0.991816369567314
146 0.992165008925033
147 0.992509472961103
148 0.992851849336348
149 0.993181699746645
150 0.993509462496117
151 0.993830962263116
152 0.994150374369289
153 0.994461435832164
154 0.994747445365143
155 0.995029279576473
156 0.99530067548368
157 0.995551194782639
158 0.995785012795002
159 0.99601256782489
160 0.996238035193954
161 0.996463502563018
162 0.996686882271257
163 0.996899823675373
164 0.997108589757839
165 0.997315268179481
166 0.997503157653701
167 0.997688959467096
168 0.997868498298017
169 0.998045949468114
170 0.998221312977386
171 0.998394588825833
172 0.998557426370157
173 0.998716088592831
174 0.998864312511382
175 0.999008361108284
176 0.999131533096939
177 0.999250529763945
178 0.999365351109302
179 0.999478084793834
180 0.999578292513418
181 0.99965971128558
182 0.999734867075268
183 0.999803759882481
184 0.999860126724747
185 0.999912318245364
186 0.999958246783507
187 1
};
\addlegendentry{Pixel AP}

\addplot [thick, dotted, red] coordinates {(25, -0.0444148933049291) (25, 1.02)};

\end{axis}

\end{tikzpicture}
    \begin{tikzpicture}[scale=0.5]

\definecolor{darkgray176}{RGB}{176,176,176}
\definecolor{steelblue31119180}{RGB}{31,119,180}

\begin{axis}[
tick align=outside,
tick pos=left,
title={\fontsize{16}{16}\selectfont Distribution of Mask Sizes},
x grid style={darkgray176},
xlabel={\fontsize{16}{16}\selectfont Mask Size},
xmin=-1239.9, xmax=26477.9,
xtick style={color=black},
y grid style={darkgray176},
ylabel={\fontsize{16}{16}\selectfont Number of Instances},
ymin=0, ymax=121.8,
tick label style={font=\fontsize{14}{14}\selectfont},
ytick style={color=black}
]
\draw[draw=black,fill=steelblue31119180,fill opacity=0.75] (axis cs:20,0) rectangle (axis cs:1027.92,116);
\draw[draw=black,fill=steelblue31119180,fill opacity=0.75] (axis cs:1027.92,0) rectangle (axis cs:2035.84,22);
\draw[draw=black,fill=steelblue31119180,fill opacity=0.75] (axis cs:2035.84,0) rectangle (axis cs:3043.76,13);
\draw[draw=black,fill=steelblue31119180,fill opacity=0.75] (axis cs:3043.76,0) rectangle (axis cs:4051.68,9);
\draw[draw=black,fill=steelblue31119180,fill opacity=0.75] (axis cs:4051.68,0) rectangle (axis cs:5059.6,2);
\draw[draw=black,fill=steelblue31119180,fill opacity=0.75] (axis cs:5059.6,0) rectangle (axis cs:6067.52,3);
\draw[draw=black,fill=steelblue31119180,fill opacity=0.75] (axis cs:6067.52,0) rectangle (axis cs:7075.44,0);
\draw[draw=black,fill=steelblue31119180,fill opacity=0.75] (axis cs:7075.44,0) rectangle (axis cs:8083.36,2);
\draw[draw=black,fill=steelblue31119180,fill opacity=0.75] (axis cs:8083.36,0) rectangle (axis cs:9091.28,5);
\draw[draw=black,fill=steelblue31119180,fill opacity=0.75] (axis cs:9091.28,0) rectangle (axis cs:10099.2,1);
\draw[draw=black,fill=steelblue31119180,fill opacity=0.75] (axis cs:10099.2,0) rectangle (axis cs:11107.12,2);
\draw[draw=black,fill=steelblue31119180,fill opacity=0.75] (axis cs:11107.12,0) rectangle (axis cs:12115.04,2);
\draw[draw=black,fill=steelblue31119180,fill opacity=0.75] (axis cs:12115.04,0) rectangle (axis cs:13122.96,2);
\draw[draw=black,fill=steelblue31119180,fill opacity=0.75] (axis cs:13122.96,0) rectangle (axis cs:14130.88,0);
\draw[draw=black,fill=steelblue31119180,fill opacity=0.75] (axis cs:14130.88,0) rectangle (axis cs:15138.8,0);
\draw[draw=black,fill=steelblue31119180,fill opacity=0.75] (axis cs:15138.8,0) rectangle (axis cs:16146.72,2);
\draw[draw=black,fill=steelblue31119180,fill opacity=0.75] (axis cs:16146.72,0) rectangle (axis cs:17154.64,0);
\draw[draw=black,fill=steelblue31119180,fill opacity=0.75] (axis cs:17154.64,0) rectangle (axis cs:18162.56,1);
\draw[draw=black,fill=steelblue31119180,fill opacity=0.75] (axis cs:18162.56,0) rectangle (axis cs:19170.48,0);
\draw[draw=black,fill=steelblue31119180,fill opacity=0.75] (axis cs:19170.48,0) rectangle (axis cs:20178.4,2);
\draw[draw=black,fill=steelblue31119180,fill opacity=0.75] (axis cs:20178.4,0) rectangle (axis cs:21186.32,1);
\draw[draw=black,fill=steelblue31119180,fill opacity=0.75] (axis cs:21186.32,0) rectangle (axis cs:22194.24,1);
\draw[draw=black,fill=steelblue31119180,fill opacity=0.75] (axis cs:22194.24,0) rectangle (axis cs:23202.16,1);
\draw[draw=black,fill=steelblue31119180,fill opacity=0.75] (axis cs:23202.16,0) rectangle (axis cs:24210.08,0);
\draw[draw=black,fill=steelblue31119180,fill opacity=0.75] (axis cs:24210.08,0) rectangle (axis cs:25218,1);
\end{axis}

\end{tikzpicture}
  \end{center}%
  \vspace{-10px}%
  \caption{%
  \textbf{Pixel-level vs Instance-level AP.}
  The commonly used pixel-level AP metric favors large objects.
  We sort the $188$ instances in Fishyscapes L\&F in descending order of size and compute both pixel-level and instance-level metrics.
  By correctly predicting only $25$ objects, we can achieve a pAP of 69.03\%, while only getting an iAP of 13.82\% .
  Since the majority of the objects in the dataset are small (as seen on the right), it is important to use a metric that does not favor large objects near to the ego-vehicle.
  }
  \label{fig:papiap}
\end{figure}

Therefore, to evaluate unknown instance segmentation we use the common AP and AP50 from the Cityscapes evaluation protocol, here referred to as iAP and iAP50.
We argue that the instance-level AP metric is a harder metric since objects of all sizes are treated the same, while the pixel-level AP is dominated by large objects.
This is visually demonstrated in Figure \ref{fig:papiap}.
Unlike the default Cityscapes evaluation, we evaluate any instances larger than 10 pixels, using all objects in the Fishyscapes L\&F dataset for evaluation.

\PAR{Models and Training.}
We follow the setup by Nayal \etal \cite{nayal2023rba} for training and fine-tuning the Mask2Former.
Initially, we train the model on the Cityscapes dataset.
Then, we fine-tune the model for $5000$ iterations on Cityscapes images with pasted COCO \cite{lin2014microsoft} objects as simulated anomalies.
Differently to RbA, we fine-tune the full transformer decoder and use the entire COCO \cite{lin2014microsoft} dataset.
We do not fine-tune or alter the SAM model.

\subsection{Results}
\begin{table}[t]
\footnotesize
\centering
\caption{\textbf{Comparison to State-of-the-Art.}
We report pixel-level scores for UGainS on the Road Anomaly and the Fishyscapes Lost and Found datasets.
As can be seen, we obtain state-of-the-art performances, both with and without additional OoD training data.
($^*$: RbA scores based on our re-implementation.)
}

\begin{tabu}{lccccccccc} 
\toprule
\multirow{2}{*}{Method} & \multirow{2}{*}{\parbox{1cm}{OoD Data}} & \multirow{2}{*}{\parbox{1.2cm}{Extra Network}} &~~~& \multicolumn{2}{c}{Road Anomaly} &~~~~& \multicolumn{2}{c}{FS L\&F}  \\
\cmidrule{5-6} \cmidrule{8-9}
& & && pAP$\uparrow$ & $\text{FPR}_{95}$$\downarrow$ && pAP$\uparrow$ & $\text{FPR}_{95}$$\downarrow$ \\
\midrule
MSP \cite{hendrycks2018baseline}               & \xmark & \xmark && $20.59$ & $68.44$ && $6.02 $ & $45.63$ \\
Mahalanobis \cite{lee2018simple}          & \xmark & \xmark && $22.85$ & $59.20$ && $27.83$ & $30.17$ \\
SML \cite{jung2021standardized}                    & \xmark & \xmark && $25.82$ & $49.74$ && $36.55$ & $14.53$ \\ 
GMMSeg \cite{liang2022gmmseg}               & \xmark & \xmark && $57.65$ & $44.34$ && $50.03$ & $12.55$ \\
SynthCP \cite{xia2020synthesize}                 & \xmark & \cmark && $24.86$ & $64.69$ && $6.54 $ & $45.95$ \\
RbA \cite{nayal2023rba}               & \xmark & \xmark && $\underline{78.45}$ & $\underline{11.83}$ && $\underline{60.96}$ & $\underline{10.63}$ \\
\rowfont{\color{gray}}
RbA$^*$               & \xmark & \xmark && $74.78$ & $17.83$ && $59.51$ & $11.34$ \\
\rowcolor{LightCyan}UGainS (Ours)    & \xmark & \cmark && $\bm{81.32}$ & $\bm{11.59}$ && $\bm{70.90}$ & $\bm{10.38}$\\
\cdashlinelr{1-9}
SynBoost \cite{dibiase2021pixelwise}                       & \cmark & \cmark && $38.21$ & $64.75$ && $60.58$ & $31.02$ \\
Maximized Entropy \cite{chan2021entropy}               & \cmark & \xmark && $-$ & $-$ && $41.31$ & $37.69$ \\
PEBAL \cite{tian2022pixelwise}                           & \cmark & \xmark && $44.41$ & $37.98$ && $64.43$ & $6.56$ \\ 
DenseHybrid \cite{grcic2022densehybrid}             & \cmark & \xmark && $-$ & $-$ && $63.8$ & $\bm{6.1}$ \\
RbA \cite{nayal2023rba}               & \cmark & \xmark && $85.42$ & $ \bm{6.92}$ && $70.81$ & $\underline{6.30}$ \\
\rowfont{\color{gray}}
RbA$^*$               & \cmark & \xmark && $\underline{87.12}$ & $12.06$ && $\underline{75.61}$ & $7.46$ \\
\rowcolor{LightCyan}UGainS (Ours)    & \cmark & \cmark && $\bm{88.98}$ & $\underline{10.42}$ && $\bm{80.08}$ & $6.61$\\
\bottomrule   
\end{tabu}
\label{tab:main}
\end{table}

\PARbegin{Comparison to State-of-the-art.} %
To validate the overall performance of our approach, we first compare it on a pixel-level to existing anomaly segmentation methods.
We report performance on the FS L\&F and RoadAnomaly datasets according to the common protocol, evaluating both a model without fine-tuning, as well as a fine-tuned version.
These are shown above and below the dashed line in Table \ref{tab:main}, respectively.
On both the RoadAnomaly and Fishyscapes Lost and Found datasets we outperform other approaches in the pixel-level average precision (pAP) metric, with and without the use of additional training data.
We attribute this performance to strong object priors from our OoD instance prediction branch.

\begin{table}[t]
\centering
  \caption{
    \textbf{Upper and lower bound comparison.}
    We show how our method compares to two important baselines, sampling points in an oracle way from the ground truth and simply using dense grid sampling as proposed by SAM.
    As expected we cannot match the oracle sampling, however, we do outperform the simple dense sampling strategy significantly.
  }
  \label{tab:oracle}
  \begin{tabularx}{\textwidth}{lcccYYcYY}
    \toprule 
    \multirow{2}{*}{Method}& \multirow{2}{*}{$N$ samples} & \multirow{2}{*}{GT Data} &~~& \multicolumn{5}{c}{Fishyscapes L\&F}\tabularnewline
    \cmidrule{5-9}
    & & && pAP$\uparrow$ & $\text{FPR}_{95}$$\downarrow$ &~~~&
    iAP$\uparrow$ & iAP$50$$\uparrow$ \tabularnewline
    \midrule 
    Center Sampling (Oracle) & 1 per object & \cmark && 8.39 & 100.0 && 49.94 & 77.56\tabularnewline
    \hspace{10px} $+$ RbA\cite{nayal2023rba}~scoring & 1 per object & \cmark && 70.46 & 48.55 && 36.18 & 60.60\tabularnewline
    \hline 
    FPS Sampling $+$ RbA Scoring & 50 points & \xmark && 63.70 & 99.91 && 6.12 & 9.20\tabularnewline
    \hspace{10px} $+$ NMS & 50 points & \xmark && 73.32 & 99.54 && \multirow{2}{*}{29.75} & \multirow{2}{*}{48.35} \tabularnewline
    \hspace{10px} $+$ RbA ensemble & 50 points & \xmark && 80.08 & 6.61 && & \tabularnewline
    \midrule
    SAM \cite{kirillov2023segment} dense $+$ RbA scoring & 1024 points & \xmark && 73.85 & 5.15 && 17.02 & 24.60 \tabularnewline
    \bottomrule 
    \end{tabularx}
\end{table}

\PAR{Lower and Upper Performance Bounds of UGainS.} %
Since no other method has been properly evaluated for anomaly instance segmentation on the FS L\&F or RoadAnomaly datasets using a standard average precision metric, we unfortunately cannot compare our method to other works, but we can create two baselines using the strong SAM model.
For the first baseline, we can obtain instance predictions using the perfect sampling to define an upper boundary. 
For the second, we can get instance predictions from SAM based on the simple dense grid sampling approach to get instance masks.
These two baselines can be seen as an upper and lower bound on the performance of our approach.

The upper part of Table \ref{tab:oracle} shows the upper-bound performance when we use ground-truth annotations to sample a single point per anomaly object.
Note that such an oracle setting can only be considered for finding theoretical bounds and should not be considered as an actual performance of our approach.
While this setting achieves a strong instance AP of nearly $50\%$, the pAP score is low.
To address this, instead of using the IoU scores predicted by SAM, we use the average of the underlying RbA scores.
This significantly improves the pixel AP, but also reduces the instance-level AP (iAP) by assigning a lower score to anomalies that were not well captured by RbA.

The other end of the spectrum would be to use SAM with the default automatic mask generation module, which samples $32 \times 32$ points in a dense grid and predicts three masks for each point, followed by a series of post-processing steps \cite{kirillov2023segment}.
Since this sampling approach generates masks for the majority of objects, masks require anomaly scoring.
Similar to the upper-bound baseline, we assign the average uncertainty score of the pixels inside a mask, since SAM does not provide any type of anomaly score out of the box.
While such an approach can be used in practice, it achieves a significantly lower iAP compared to our approach, indicating  that sampling proposal points for the mask prediction in an informed way is beneficial (see \reftab{oracle}).
In addition, while low scores of dense sampling could be addressed by sampling a denser grid, the computational complexity increases quadratically.

To compare our approach, we look at the individual steps and indicate how they compare to the two upper and lower bounds.
In the simplest case, where we sample only $50$ points from the RbA pixel-level predictions, we cannot match the dense predictions of SAM.
However, since we typically sample multiple points from the same object, a significant improvement can be gained with a simple non-maximum suppression step, which is also applied to dense point sampling.
At this point, we can obtain an iAP that is significantly better than dense sampling, although there is still a gap to the oracle performance that could be attributed to incorrect sampling.
As a final step, we ensemble the average uncertainty scores under each anomaly mask with the initial RbA scores, significantly increasing the final pixel-level AP of our approach, while also being able to predict meaningful instance segmentation of the anomalous objects in the scene.

\begin{figure}[t]
  \centering
  \setlength{\tabcolsep}{1pt}
  \begin{tabular}{cccc}
    \includegraphics[width=0.244\textwidth]{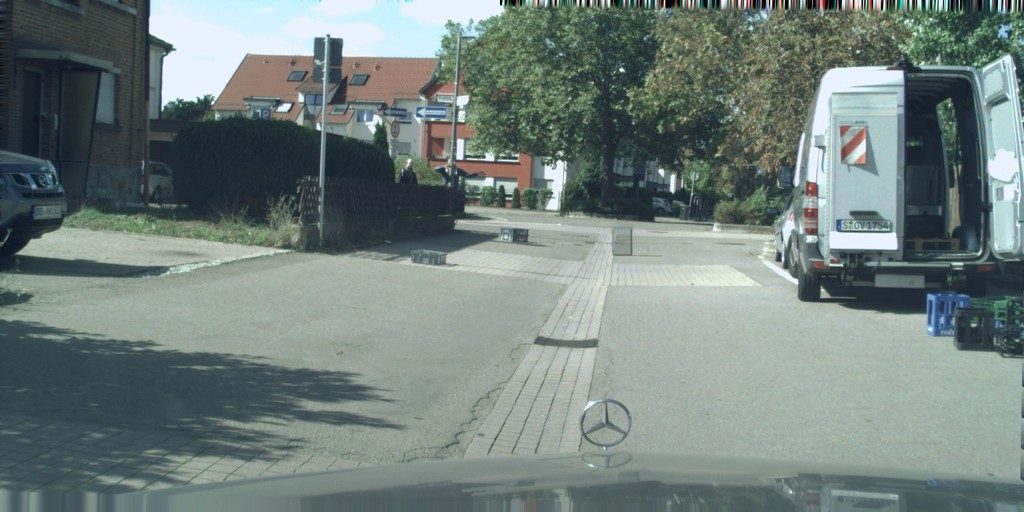}&%
    \includegraphics[width=0.244\textwidth]{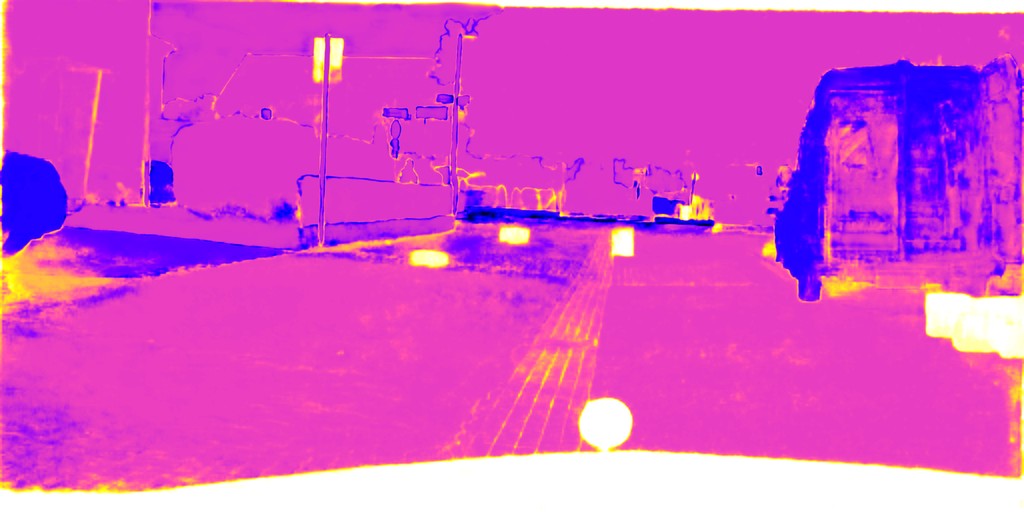}&%
    \includegraphics[width=0.244\textwidth]{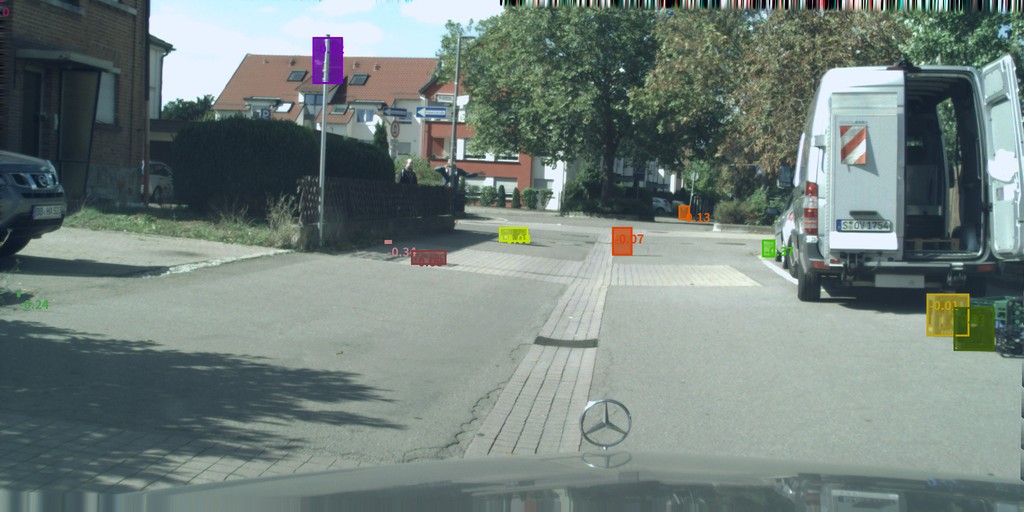}&%
    \includegraphics[width=0.244\textwidth]{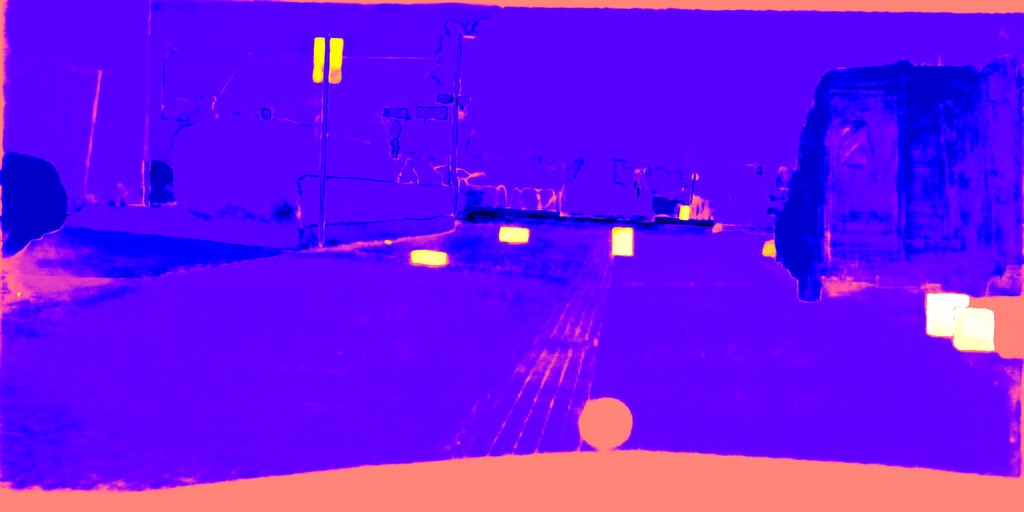}\\[-1pt]%
    \includegraphics[width=0.244\textwidth]{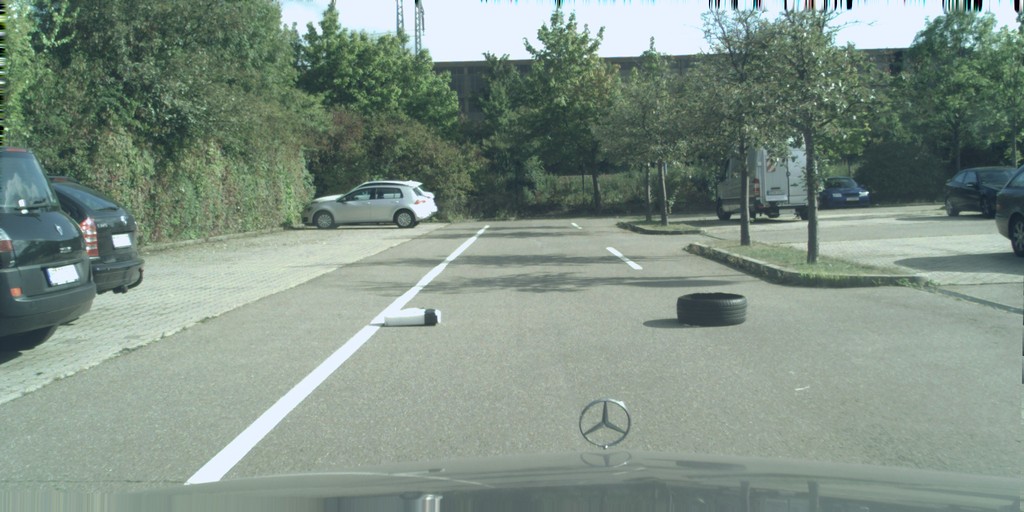}&%
    \includegraphics[width=0.244\textwidth]{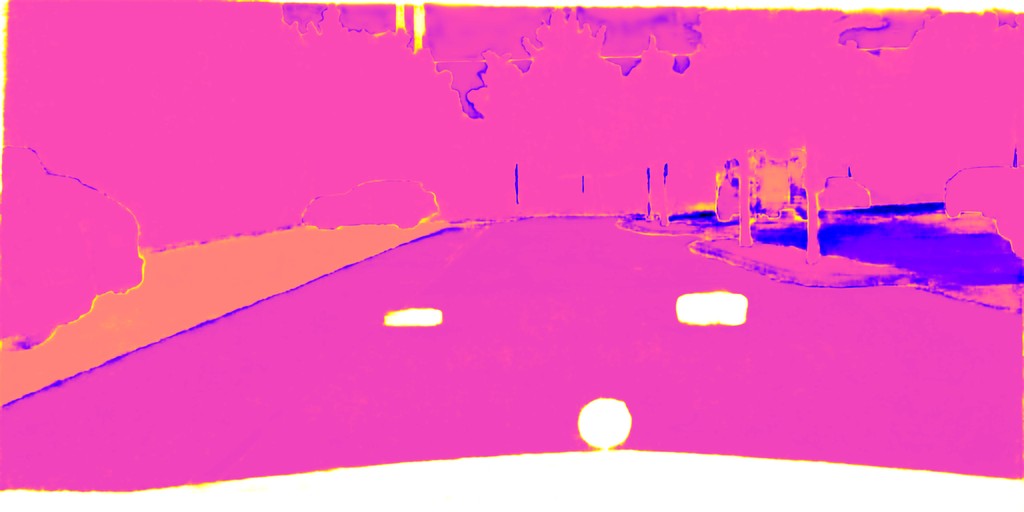}&%
    \includegraphics[width=0.244\textwidth]{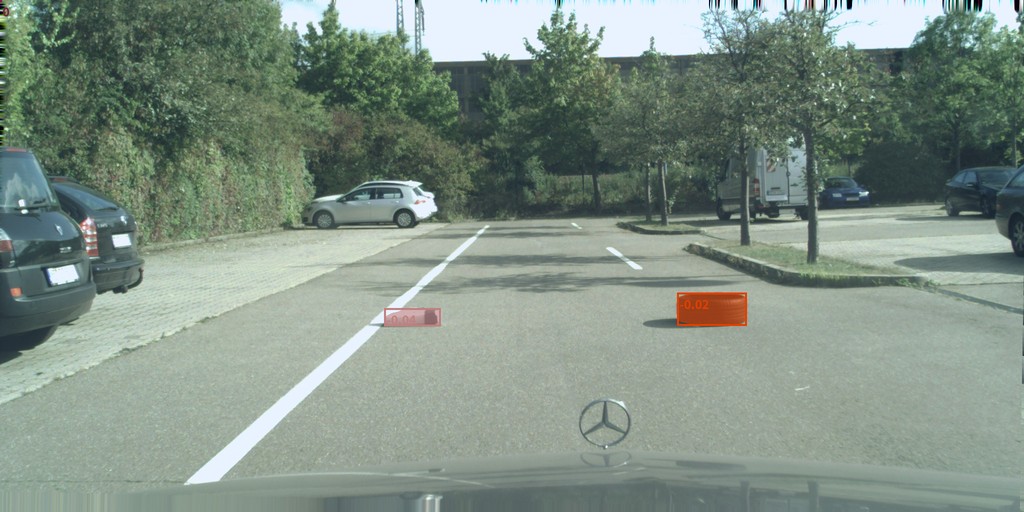}&%
    \includegraphics[width=0.244\textwidth]{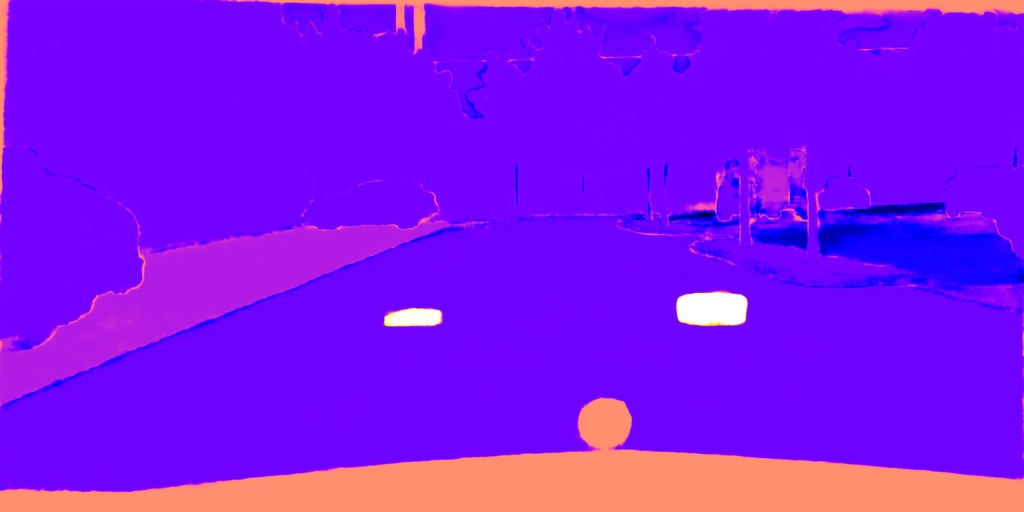}\\%
    \footnotesize Image & \footnotesize RbA scores & \footnotesize Instance segmentation & \footnotesize Updated scores 
  \end{tabular}
  \vspace{-5pt}
  \caption{
      \textbf{Qualitative results on Fishyscapes L\&F}.
      The third column shows the instance segmentations generated using RbA predictions in the second column. 
      Note the prominent anomalous objects in the refined per-pixel anomaly predictions in the fourth column.
    }
  \label{fig:qualitative_fslaf}
\end{figure}

\begin{figure}[t]
    \includegraphics[width=0.325\textwidth]{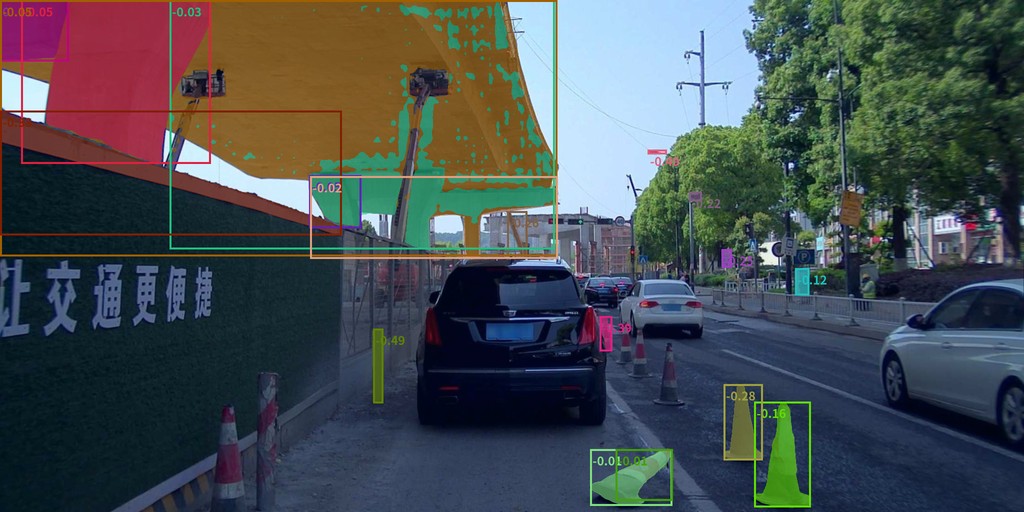}\hfill%
    \includegraphics[width=0.325\textwidth]{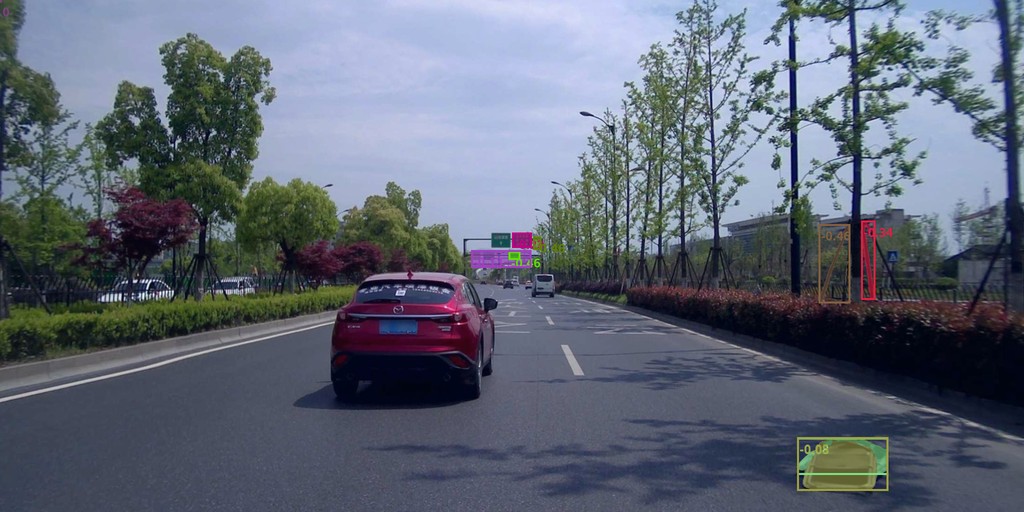}\hfill%
    \includegraphics[width=0.325\textwidth]{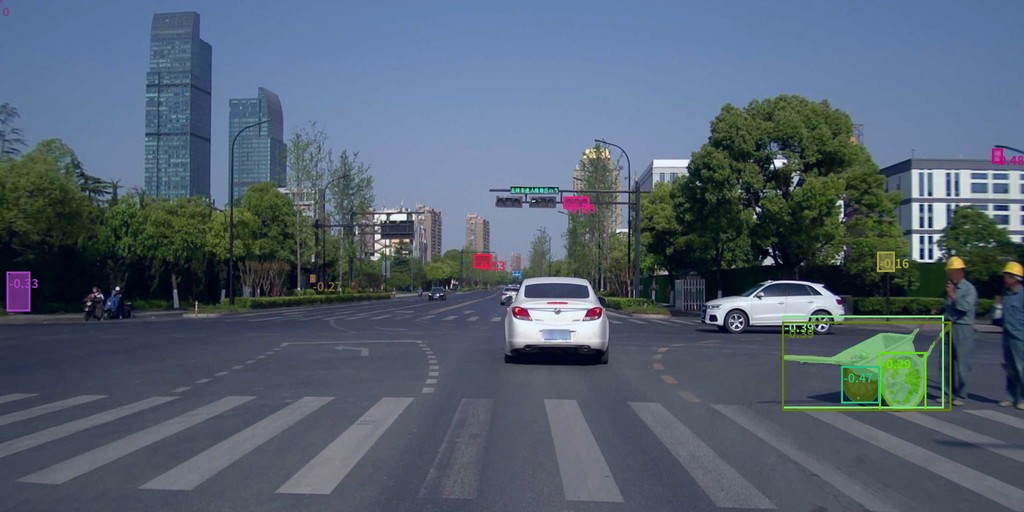}
  \vspace{-5pt}
  \caption{
      \textbf{Qualitative results on CODA.} 
      Our model also performs well even under a larger domain shift.
      We are able to segment the typical unknown dangerous objects on the road, but it also reveals other instances for which the model has no proper semantic knowledge, \eg bridges.
  }
  \label{fig:qualitative_coda}
\end{figure}

\PAR{Qualitative Results.} 
We show qualitative results of our approach on the Fishyscapes L\&F dataset in Figure~\ref{fig:qualitative_fslaf}.
It can be seen how our approach segments instances from the pixel-level anomaly score predictions.
In addition, UGainS can further refine the initial pixel-level predictions using instance predictions.
In Figure~\ref{fig:qualitative_coda} we show additional qualitative results on the more challenging CODA dataset \cite{li2022coda}.
Here we show that our method can capture interesting objects that are not well-covered in the training set.
This ability is valuable for model introspection, providing insight into instances that the model may struggle with.

\subsection{Ablations}
\label{sec:ablations}
In this section, we ablate some of our design choices.
We focus on how pixel-level performance, the proposal sampling method, and the number of sampled points affect instance-level performance.

\begin{table}[t]
  \caption{
    \textbf{UGainS improvements of several anomaly segmentation methods.}
    Our method notably improves pixel-level AP results of several methods in addition to providing anomaly instance segmentations.
  }
  \label{tab:different_per_pixel_methods}
  \centering
\begin{tabular}{lccccccc}
\toprule 
\multirow{2}{*}{Method} & \multirow{2}{*}{UGainS} &~~& \multicolumn{5}{c}{Fishyscapes L\&F} \tabularnewline
\cmidrule{4-8}
&  && \multicolumn{2}{c}{pAP$\uparrow$} & $\text{FPR}_{95}$$\downarrow$ & iAP$\uparrow$ & iAP$50$$\uparrow$\tabularnewline
\midrule
\multirow{2}{*}{Max Softmax \cite{hendrycks2018baseline}} & \xmark && $4.58$ & \hspace{10px}\multirow{2}{*}{\ArrowDown{$+5.73$}} & $40.59$ & $-$ & $-$ \tabularnewline
 & \cmark && $10.31$ & & $46.43$ & $5.61$ & $9.11$ \tabularnewline
\midrule 
\multirow{2}{*}{Mahalanobis \cite{lee2018simple}} & \xmark && $28.75$ & \hspace{10px}\multirow{2}{*}{\ArrowDown{$+1.51$}} & $9.52$ & $-$ & $-$ \tabularnewline
 & \cmark && $30.26$ & & $13.27$ & $2.69$ & $4.18$ \tabularnewline
\midrule 
\multirow{2}{*}{Maximzed Entropy \cite{chan2021entropy}} & \xmark && $41.31$ & \hspace{10px}\multirow{2}{*}{\ArrowDown{$+6.67$}} & $37.67$ & $-$ & $-$ \tabularnewline
 & \cmark && $47.98$ & & $37.86$ & $12.16$ & $18.72$  \tabularnewline
\midrule
\multirow{2}{*}{DenseHybrid \cite{grcic2022densehybrid}} & \xmark && $69.78$ & \hspace{10px}\multirow{2}{*}{\ArrowDown{$+1.41$}} & $5.08$ & $-$ & $-$ \tabularnewline
 & \cmark && $71.19$ & & $5.77$ & $27.49$ & $44.07$\tabularnewline 
\midrule
\multirow{2}{*}{RbA\cite{nayal2023rba}} & \xmark && $75.61$ & \hspace{10px}\multirow{2}{*}{\ArrowDown{$+4.47$}} & $7.46$ & $-$ & $-$ \tabularnewline
 & \cmark && $80.08$ & & $6.61$ & $29.75$ & $48.35$ \tabularnewline \bottomrule 
\end{tabular}
\end{table}

\PAR{Evaluation of Other Per-Pixel Methods.}
Although we get the best performance with the RbA approach, we want to note that we can get instance segmentation and per-pixel score improvements for other anomaly segmentation methods as well.
Table \ref{tab:different_per_pixel_methods} shows the performance differences and anomaly instance segmentation performance of our approach for several other per-pixel anomaly segmentation methods.
In all cases, we can improve the per-pixel AP by quite a large margin, and we observe that higher pixel-level AP generally leads to higher instance-level AP.
Since none of the other approaches achieve a similar or higher resulting instance-level AP, this indicates that all tested methods still have problems with small/far objects.

\begin{figure}[t]
  \begin{center}
  \begin{tabular}{ccc}
    \includegraphics[width=0.328\textwidth]{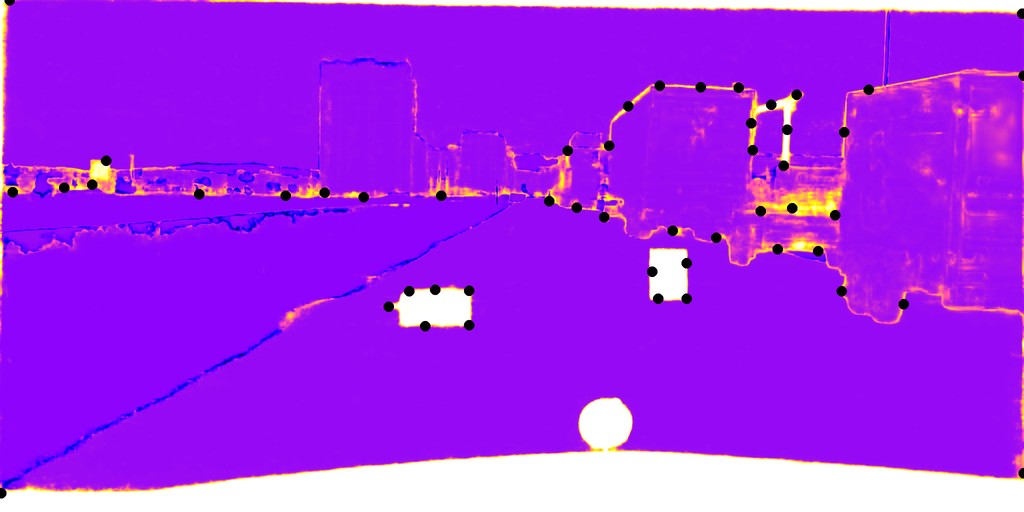} & 
    \includegraphics[width=0.328\textwidth]{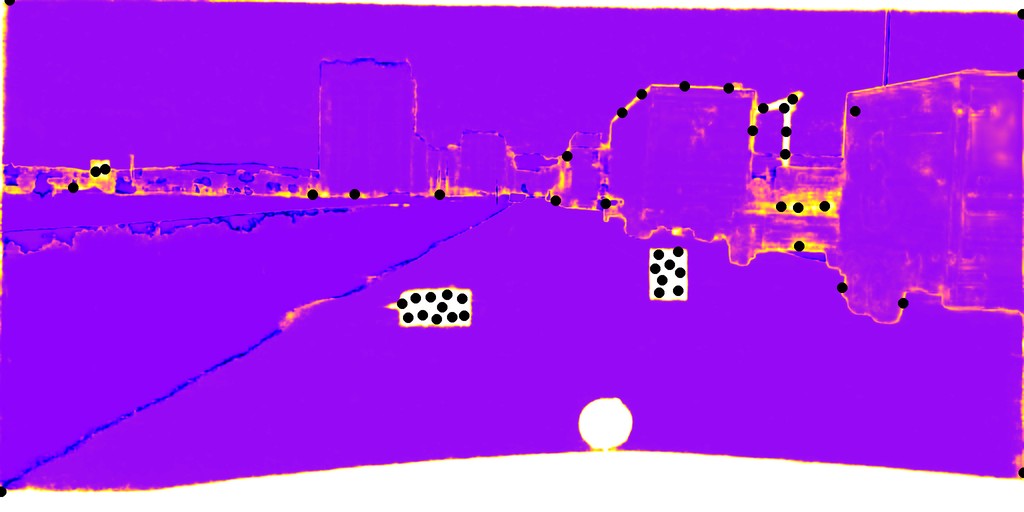} &
    \includegraphics[width=0.328\textwidth]{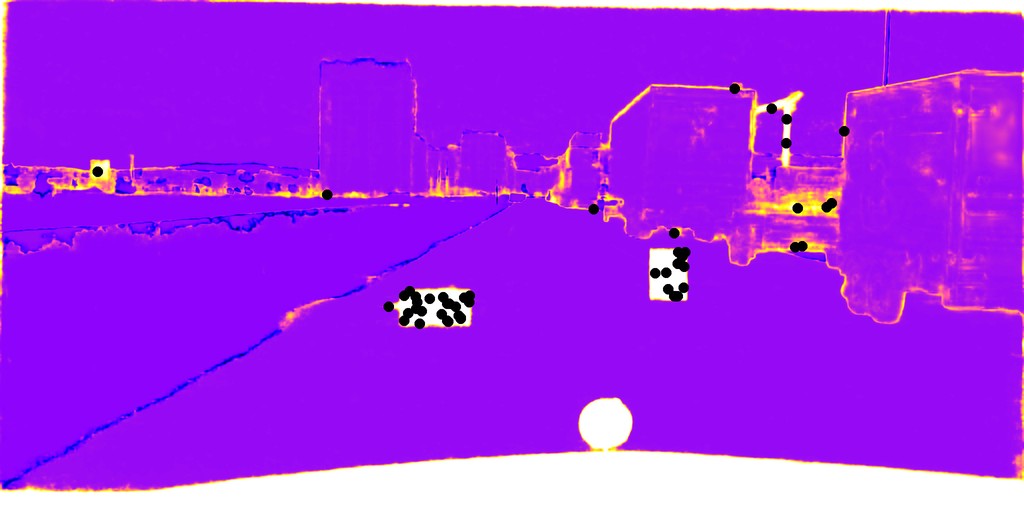} \\[-7pt]
    \footnotesize FPS & \footnotesize K-means & \footnotesize Random \\
  \end{tabular}
  \end{center}
  \vspace{-20px}
  \caption{
  \textbf{Visual comparison of different anomaly point sampling methods.}
  Bright areas correspond to high RbA scores and black dots represent the sampled points.
  K-means estimates points mostly in the center of uncertain regions, while FPS generates most points at the boundary.
  Random sampling randomly selects points from the high uncertainty regions and thus often misses smaller objects.
  }
  \label{fig:kmeansfps}
\end{figure}

\begin{table}[t]
  \centering
  \caption{\textbf{Performance impact of the proposal sampling methods.}
    On the left we report the typical metrics, whereas on the right we additionally report metrics of the sampled points, \ie how many points fall on an anomalous object (precision) and how many anomalous objects are sampled at least once (recall).
    FPS and K-means performs roughly on par, while random sampling performs worse due to a lower recall on the point level.
  }
  \label{tab:sampling_methods}
    \begin{tabular}{lccccccccccc}
      \toprule
        \multirow{2}{*}{Sampling method} &~~& \multicolumn{4}{c}{Full Metrics} &~~~& \multicolumn{2}{c}{Point Level}\\
        \cmidrule{3-6} \cmidrule{8-9}
        && pAP$\uparrow$ & $\text{FPR}_{95}$$\downarrow$ & iAP$\uparrow$ & iAP$50$$\uparrow$ && Precision & Recall \\
      \midrule
        Random Sampling && $79.81$ & $5.97$ & $25.82$ & $40.45$  && $57.22$ & $60.11$ \\
        K-means && $79.92$ & $6.60$ & $29.04$ & $46.23$ && $47.51$ & $70.74$ \\
        FPS && $80.08$ & $6.61$ & $29.75$ & $48.35$ && $35.30$ & $71.81$  \\
      \bottomrule
    \end{tabular}
\end{table}

\begin{table}[t]
  \caption{
    \textbf{Performance impact of the number of sampled points.}
    Considering there are often a few larger anomalous instances per frame, we see a diminishing return after sampling more than 50 points. 
    While the recall of smaller instances might go up, we are also bound to segment false positive instances, thus reducing the precision.
  }
  \label{tab:numberofsamples}
  \centering
    \begin{tabular}{ccccccc}
      \toprule
        \# Points &~~& pAP$\uparrow$ & $\text{FPR}_{95}$$\downarrow$ &~~& iAP$\uparrow$ & iAP$50$$\uparrow$ \\
      \midrule
        $10$  && $76.73$ & $\bm{5.89}$ && $26.81$ & $41.33$\\
        $20$  && $78.83$ & $\underline{5.90}$ && $\bm{30.82}$ & $\mathbf{49.14}$ \\
        $50$  && $\bm{80.08}$ & $6.61$ && $\underline{29.75}$ & $\underline{48.35}$ \\
        $100$ && $78.96$ & $7.01$ && $27.16$ & $45.39$ \\
        $200$ && $78.99$ & $7.09$ && $25.77$ & $43.82$ \\
        $300$ && $78.55$ & $7.16$ && $24.44$ & $41.98$ \\
      \bottomrule
    \end{tabular}
\end{table}

\PAR{Sampling methods.} In Table \ref{tab:sampling_methods}, we evaluate farthest point sampling (FPS), K-means, and random sampling from uncertainty regions.
We observe that K-means predicts points closer to the centroids of the objects (see \reffig{kmeansfps}).
In contrast, FPS generates proposals at boundaries and achieves higher recall.
Due to the maximum recall achieved with the same number of points, we chose farthest point sampling as the best option.
We also evaluate random sampling, but samples favor larger objects, resulting in a higher point-level precision, but a significantly lower recall compared to FPS and K-means.

\PAR{Number of samples.} 
In Table \ref{tab:numberofsamples} we evaluate the impact of the number of sampled point proposals used to generate instance masks.
Since the FS L\&F dataset typically contains several instances per scene, we observe diminishing returns as the number of samples increases.
With increased numbers of samples, most of the predicted masks end up getting removed by non-maximum suppression, and the approach tends to predict multiple small boundary regions increasing the number of false positives.

\section{Conclusion}
We have proposed a novel approach for uncertainty-guided segmentation of anomalous instances.
By combining anomaly object scores with initial per-pixel anomaly scores our approach shows consistent improvement on the task of anomaly segmentation.
We show that UGainS achieves state-of-the-art average precision performance for pixel-level predictions on the Fishyscapes L\&F and RoadAnomaly datasets.
While anomaly segmentation on existing datasets seems to slowly saturate, we find that the more challenging task of anomaly instance segmentation is far from this point.
Our method provides anomaly instance masks that can be used for model introspection and collection of rare instances and results can further be improved both through better anomaly point proposals and better instance segmentation. 

\parag{Acknowledgements.} 
This project was partially funded by the Ford University Alliance Program (project FA0347).
Compute resources were granted by RWTH Aachen under project ‘supp0003’.
We thank Ali Athar, George Lydakis, Idil Esen Zulfikar, Fabian Rempfer and the reviewers for helpful feedback.

\bibliographystyle{splncs04}
\bibliography{egbib}

\begin{thebibliography}{10}
\providecommand{\url}[1]{\texttt{#1}}
\providecommand{\urlprefix}{URL }
\providecommand{\doi}[1]{https://doi.org/#1}

\bibitem{besnier2021triggering}
Besnier, V., Bursuc, A., Picard, D., Briot, A.: {Triggering Failures: Out-Of-Distribution detection by learning from local adversarial attacks in Semantic Segmentation}. In: International Conference on Computer Vision (ICCV) (2021)

\bibitem{bevandic2019simultaneous}
Bevandić, P., Krešo, I., Oršić, M., Šegvić, S.: {Simultaneous Semantic Segmentation and Outlier Detection in Presence of Domain Shift}. In: German Conference on Pattern Recognition (GCPR) (2019)

\bibitem{blum2021fishyscapes}
Blum, H., Sarlin, P.E., Nieto, J., Siegwart, R., Cadena, C.: {The Fishyscapes Benchmark: Measuring Blind Spots in Semantic Segmentation}. International Journal of Computer Vision  \textbf{129}(11),  3119–3135 (2021)

\bibitem{cao2023segment}
Cao, Y., Xu, X., Sun, C., Cheng, Y., Du, Z., Gao, L., Shen, W.: {Segment Any Anomaly without Training via Hybrid Prompt Regularization}. arXiv preprint arXiv:2305.10724  (2023)

\bibitem{chan2021segmentmeifyoucan}
Chan, R., Lis, K., Uhlemeyer, S., Blum, H., Honari, S., Siegwart, R., Fua, P., Salzmann, M., Rottmann, M.: {SegmentMeIfYouCan: A Benchmark for Anomaly Segmentation}. In: Proceedings of the Neural Information Processing Systems Track on Datasets and Benchmarks (2021)

\bibitem{chan2021entropy}
Chan, R., Rottmann, M., Gottschalk, H.: {Entropy Maximization and Meta Classification for Out-Of-Distribution Detection in Semantic Segmentation}. In: International Conference on Computer Vision (ICCV) (2021)

\bibitem{chen2018encoderdecoder}
Chen, L.C., Zhu, Y., Papandreou, G., Schroff, F., Adam, H.: {Encoder-Decoder with Atrous Separable Convolution for Semantic Image Segmentation}. In: European Conference on Computer Vision (ECCV) (2018)

\bibitem{cheng2022maskedattention}
Cheng, B., Misra, I., Schwing, A.G., Kirillov, A., Girdhar, R.: {Masked-attention Mask Transformer for Universal Image Segmentation}. In: Conference on Computer Vision and Pattern Recognition (CVPR) (2022)

\bibitem{cordts2016cityscapes}
Cordts, M., Omran, M., Ramos, S., Rehfeld, T., Enzweiler, M., Benenson, R., Franke, U., Roth, S., Schiele, B.: {The Cityscapes Dataset for Semantic Urban Scene Understanding}. In: Conference on Computer Vision and Pattern Recognition (CVPR) (2016)

\bibitem{dibiase2021pixelwise}
Di~Biase, G., Blum, H., Siegwart, R., Cadena, C.: {Pixel-wise Anomaly Detection in Complex Driving Scenes}. In: Conference on Computer Vision and Pattern Recognition (CVPR) (2021)

\bibitem{ding2023openvocabulary}
Ding, Z., Wang, J., Tu, Z.: {Open-Vocabulary Universal Image Segmentation with MaskCLIP}. In: International Conference on Machine Learning (ICML) (2023)

\bibitem{dosovitskiy2021vit}
Dosovitskiy, A., Beyer, L., Kolesnikov, A., Weissenborn, D., Zhai, X., Unterthiner, T., Dehghani, M., Minderer, M., Heigold, G., Gelly, S., Uszkoreit, J., Houlsby, N.: {An Image is Worth 16x16 Words: Transformers for Image Recognition at Scale}. In: International Conference on Learning Representations (ICLR) (2021)

\bibitem{gasperini2022holistic}
Gasperini, S., Marcos-Ramiro, A., Schmidt, M., Navab, N., Busam, B., Tombari, F.: {Holistic Segmentation}. arXiv preprint arXiv:2209.05407  (2022)

\bibitem{grcic2021dense}
Grcić, M., Bevandić, P., Kalafatić, Z., Šegvić, S.: {Dense anomaly detection by robust learning on synthetic negative data}. arXiv preprint arXiv:2112.12833  (2021)

\bibitem{grcic2022densehybrid}
Grcić, M., Bevandić, P., Šegvić, S.: {DenseHybrid: Hybrid Anomaly Detection for Dense Open-set Recognition}. In: European Conference on Computer Vision (ECCV) (2022)

\bibitem{guo2017calibration}
Guo, C., Pleiss, G., Sun, Y., Weinberger, K.Q.: {On calibration of modern neural networks}. In: International Conference on Machine Learning (ICML) (2017)

\bibitem{gupta2018mergenet}
Gupta, K., Javed, S.A., Gandhi, V., Krishna, K.M.: {MergeNet: A Deep Net Architecture for Small Obstacle Discovery}. In: International Conference on Robotics and Automation (ICRA) (2018)

\bibitem{hendrycks2018baseline}
Hendrycks, D., Gimpel, K.: {A Baseline for Detecting Misclassified and Out-of-Distribution Examples in Neural Networks}. In: International Conference on Learning Representations (ICLR) (2018)

\bibitem{hendrycks2019deep}
Hendrycks, D., Mazeika, M., Dietterich, T.: {Deep Anomaly Detection with Outlier Exposure}. In: International Conference on Learning Representations (ICLR) (2019)

\bibitem{hwang2021exemplarbased}
Hwang, J., Oh, S.W., Lee, J.Y., Han, B.: {Exemplar-Based Open-Set Panoptic Segmentation Network}. In: Conference on Computer Vision and Pattern Recognition (CVPR) (2021)

\bibitem{ji2023segment}
Ji, W., Li, J., Bi, Q., Liu, T., Li, W., Cheng, L.: {Segment Anything Is Not Always Perfect: An Investigation of SAM on Different Real-world Applications}. In: Conference on Computer Vision and Pattern Recognition Workshop (CVPR'W) (2023)

\bibitem{jia2021scaling}
Jia, C., Yang, Y., Xia, Y., Chen, Y.T., Parekh, Z., Pham, H., Le, Q.V., Sung, Y., Li, Z., Duerig, T.: {Scaling Up Visual and Vision-Language Representation Learning With Noisy Text Supervision}. In: International Conference on Machine Learning (ICML) (2021)

\bibitem{jiang2022improving}
Jiang, C.M., Najibi, M., Qi, C.R., Zhou, Y., Anguelov, D.: {Improving the Intra-class Long-tail in 3D Detection via Rare Example Mining}. In: European Conference on Computer Vision (ECCV) (2022)

\bibitem{jung2021standardized}
Jung, S., Lee, J., Gwak, D., Choi, S., Choo, J.: {Standardized Max Logits: A Simple yet Effective Approach for Identifying Unexpected Road Obstacles in Urban-Scene Segmentation}. In: International Conference on Computer Vision (ICCV) (2021)

\bibitem{kendall2017what}
Kendall, A., Gal, Y.: {What Uncertainties Do We Need in Bayesian Deep Learning for Computer Vision?} In: Neural Information Processing Systems (NIPS) (2017)

\bibitem{kirillov2023segment}
Kirillov, A., Mintun, E., Ravi, N., Mao, H., Rolland, C., Gustafson, L., Xiao, T., Whitehead, S., Berg, A.C., Lo, W.Y., Dollár, P., Girshick, R.: {Segment Anything}. In: International Conference on Computer Vision (ICCV) (2023)

\bibitem{lakshminarayanan2017simple}
Lakshminarayanan, B., Pritzel, A., Blundell, C.: {Simple and Scalable Predictive Uncertainty Estimation using Deep Ensembles}. In: Neural Information Processing Systems (NIPS) (2017)

\bibitem{lee2018simple}
Lee, K., Lee, K., Lee, H., Shin, J.: {A Simple Unified Framework for Detecting Out-of-Distribution Samples and Adversarial Attacks}. In: Neural Information Processing Systems (NIPS) (2018)

\bibitem{li2022coda}
Li, K., Chen, K., Wang, H., Hong, L., Ye, C., Han, J., Chen, Y., Zhang, W., Xu, C., Yeung, D.Y., Liang, X., Li, Z., Xu, H.: {CODA: A Real-World Road Corner Case Dataset for Object Detection in Autonomous Driving}. In: European Conference on Computer Vision (ECCV) (2022)

\bibitem{liang2022gmmseg}
Liang, C., Wang, W., Miao, J., Yang, Y.: {GMMSeg: Gaussian Mixture based Generative Semantic Segmentation Models}. In: Neural Information Processing Systems (NIPS) (2022)

\bibitem{liang2018enhancing}
Liang, S., Li, Y., Srikant, R.: {Enhancing The Reliability of Out-of-distribution Image Detection in Neural Networks}. In: International Conference on Learning Representations (ICLR) (2018)

\bibitem{lin2014microsoft}
Lin, T.Y., Maire, M., Belongie, S., Bourdev, L., Girshick, R., Hays, J., Perona, P., Ramanan, D., Zitnick, C.L., Dollár, P.: {Microsoft COCO: Common Objects in Context}. In: European Conference on Computer Vision (ECCV) (2014)

\bibitem{lis2021detecting}
Lis, K., Honari, S., Fua, P., Salzmann, M.: {Detecting Road Obstacles by Erasing Them}. arXiv preprint arXiv:2012.13633  (2021)

\bibitem{lis2019detecting}
Lis, K., Nakka, K., Fua, P., Salzmann, M.: {Detecting the Unexpected via Image Resynthesis}. In: International Conference on Computer Vision (ICCV) (2019)

\bibitem{liu2022opening}
Liu, Y., Zulfikar, I.E., Luiten, J., Dave, A., Ramanan, D., Leibe, B., Ošep, A., Leal-Taixé, L.: {Opening up Open-World Tracking}. In: Conference on Computer Vision and Pattern Recognition (CVPR) (2022)

\bibitem{liu2021swin}
Liu, Z., Lin, Y., Cao, Y., Hu, H., Wei, Y., Zhang, Z., Lin, S., Guo, B.: {Swin transformer: Hierarchical vision transformer using shifted windows}. In: Conference on Computer Vision and Pattern Recognition (CVPR) (2021)

\bibitem{ma2023segment}
Ma, J., He, Y., Li, F., Han, L., You, C., Wang, B.: {Segment Anything in Medical Images}. arXiv preprint arXiv:2304.12306  (2023)

\bibitem{mukhoti2019evaluating}
Mukhoti, J., Gal, Y.: {Evaluating Bayesian Deep Learning Methods for Semantic Segmentation}. arXiv preprint arXiv:1811.12709  (2019)

\bibitem{nayal2023rba}
Nayal, N., Yavuz, M., Henriques, J.F., Güney, F.: {RbA: Segmenting Unknown Regions Rejected by All}. In: International Conference on Computer Vision (ICCV) (2023)

\bibitem{ohgushi2020road}
Ohgushi, T., Horiguchi, K., Yamanaka, M.: {Road Obstacle Detection Method Based on an Autoencoder with Semantic Segmentation}. In: Asian Conference on Computer Vision (ACCV) (2020)

\bibitem{pinggera2016lost}
Pinggera, P., Ramos, S., Gehrig, S., Franke, U., Rother, C., Mester, R.: {Lost and Found: Detecting Small Road Hazards for Self-Driving Vehicles}. In: International Conference on Intelligent Robots and Systems (IROS) (2016)

\bibitem{radford2021learning}
Radford, A., Kim, J.W., Hallacy, C., Ramesh, A., Goh, G., Agarwal, S., Sastry, G., Askell, A., Mishkin, P., Clark, J., Krueger, G., Sutskever, I.: {Learning Transferable Visual Models From Natural Language Supervision}. In: International Conference on Machine Learning (ICML) (2021)

\bibitem{rao2022denseclip}
Rao, Y., Zhao, W., Chen, G., Tang, Y., Zhu, Z., Huang, G., Zhou, J., Lu, J.: {DenseCLIP: Language-Guided Dense Prediction with Context-Aware Prompting}. In: Conference on Computer Vision and Pattern Recognition (CVPR) (2022)

\bibitem{singh2020lidar}
Singh, A., Kamireddypalli, A., Gandhi, V., Krishna, K.M.: {LiDAR guided Small obstacle Segmentation}. In: International Conference on Intelligent Robots and Systems (IROS) (2020)

\bibitem{tang2023can}
Tang, L., Xiao, H., Li, B.: {Can SAM Segment Anything? When SAM Meets Camouflaged Object Detection}. arXiv preprint arXiv:2304.04709  (2023)

\bibitem{tian2022pixelwise}
Tian, Y., Liu, Y., Pang, G., Liu, F., Chen, Y., Carneiro, G.: {Pixel-wise Energy-biased Abstention Learning for Anomaly Segmentation on Complex Urban Driving Scenes}. In: European Conference on Computer Vision (ECCV) (2022)

\bibitem{valdenegro-toro2021find}
Valdenegro-Toro, M.: {I Find Your Lack of Uncertainty in Computer Vision Disturbing}. In: Conference on Computer Vision and Pattern Recognition Workshop (CVPR'W) (2021)

\bibitem{wong2019identifying}
Wong, K., Wang, S., Ren, M., Liang, M., Urtasun, R.: {Identifying Unknown Instances for Autonomous Driving}. In: Conference on Robot Learning (CoRL) (2019)

\bibitem{wu2023medical}
Wu, J., Zhang, Y., Fu, R., Fang, H., Liu, Y., Wang, Z., Xu, Y., Jin, Y.: {Medical SAM Adapter: Adapting Segment Anything Model for Medical Image Segmentation}. arXiv preprint arXiv:2304.12620  (2023)

\bibitem{xia2020synthesize}
Xia, Y., Zhang, Y., Liu, F., Shen, W., Yuille, A.: {Synthesize then Compare: Detecting Failures and Anomalies for Semantic Segmentation}. In: European Conference on Computer Vision (ECCV) (2020)

\bibitem{xie2021segformer}
Xie, E., Wang, W., Yu, Z., Anandkumar, A., Alvarez, J.M., Luo, P.: {SegFormer: Simple and Efficient Design for Semantic Segmentation with Transformers}. In: Neural Information Processing Systems (NIPS) (2021)

\bibitem{xu2022groupvit}
Xu, J., De~Mello, S., Liu, S., Byeon, W., Breuel, T., Kautz, J., Wang, X.: {GroupViT: Semantic Segmentation Emerges from Text Supervision}. In: Conference on Computer Vision and Pattern Recognition (CVPR) (2022)

\bibitem{xue2020tiny}
Xue, F., Ming, A., Zhou, Y.: {Tiny Obstacle Discovery by Occlusion-Aware Multilayer Regression}. IEEE Transactions on Image Processing (TIP)  \textbf{29},  9373–9386 (2020)

\bibitem{yu2023inpaint}
Yu, T., Feng, R., Feng, R., Liu, J., Jin, X., Zeng, W., Chen, Z.: {Inpaint Anything: Segment Anything Meets Image Inpainting}. arXiv preprint arXiv:2304.06790  (2023)

\bibitem{zou2022generalized}
Zou, X., Dou, Z.Y., Yang, J., Gan, Z., Li, L., Li, C., Dai, X., Behl, H., Wang, J., Yuan, L., Peng, N., Wang, L., Lee, Y.J., Gao, J.: {Generalized Decoding for Pixel, Image, and Language}. In: Conference on Computer Vision and Pattern Recognition (CVPR) (2022)

\bibitem{zou2023segment}
Zou, X., Yang, J., Zhang, H., Li, F., Li, L., Wang, J., Wang, L., Gao, J., Lee, Y.J.: {Segment Everything Everywhere All at Once}. arXiv preprint arXiv:2304.06718  (2023)

\end{thebibliography}

\clearpage

{ \centering \LARGE \textbf{Supplementary Material\\[1cm]}}

\setcounter{equation}{0}
\setcounter{figure}{0}
\setcounter{table}{0}
\setcounter{page}{1}
\setcounter{section}{0}
\makeatletter
\renewcommand{\theequation}{S\arabic{equation}}
\renewcommand{\thefigure}{S\arabic{figure}}
\renewcommand{\thetable}{S\arabic{table}}
\renewcommand{\thesection}{S\arabic{section}}

In this supplementary we provide additional qualitative results, a comparison to the U3HS method, and visualizations of point sampling methods using different numbers of points.

\section{Comparison to U3HS \cite{gasperini2022holistic}}
\vspace{-5pt}
For the sake of completeness, we provide a comparison to the U3HS method using the panoptic metrics they use.
We do not use this as our main metric, since the pantopic metric for a single class gives little additional information compared to the AP metric.
UGainS achieves higher scores in panoptic quality and pixel-level average precision.
However, it is worth noting that U3HS is not exposed to any anomalies during training and does not use a strong generalist model.
In addition, U3HS uses strict probabilistic uncertainty modeling, which typically performs worse on a task of anomaly segmentation.

\begin{table}[h]
  \caption{
    Pixel-level and panoptic results on the Lost and Found test set.
  }
  \label{tab:lafpap}
  \begin{center}
    \begin{tabular}[c]{cccccc}
      \toprule
      Method & pAP$\uparrow$ & $\text{FPR}_{95}$$\downarrow$ & PQ & RQ & SQ \\
      \midrule
      U3HS & $25.44$ & $19.10$ & $7.94$ & $12.37$ & $64.24$ \\
      UGainS & $82.99$ & $22.65$ & $55.44$ & $68.14$ & $81.36$ \\
      \bottomrule
    \end{tabular}
  \end{center}
\end{table}

\section{Number of Points for Sampling}
\begin{figure}[ht]
  \centering
  \begin{tabular}{cccc}
    \rotatebox{90}{\hspace{17pt} 10} &
    \includegraphics[trim={0px 0px 0px 0px},clip,width=0.319\textwidth]{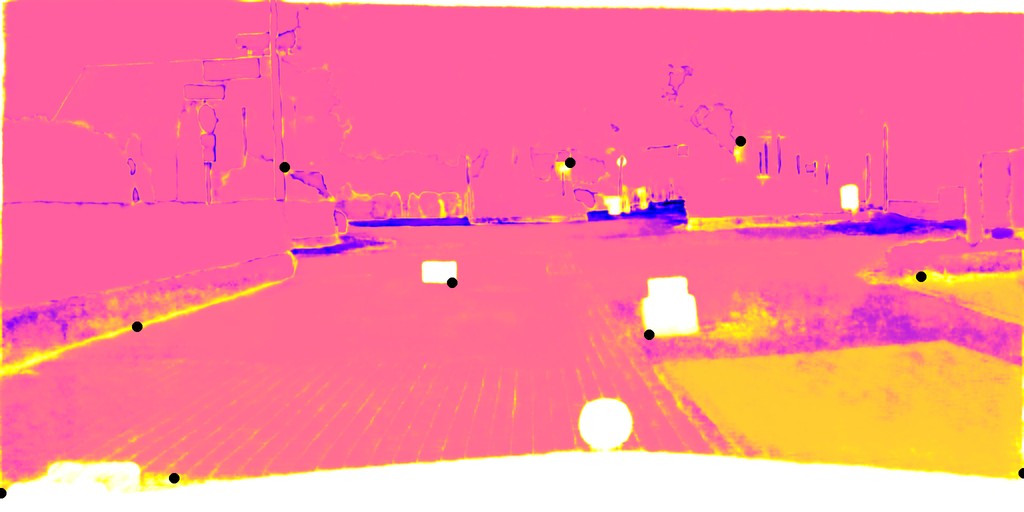}&%
    \includegraphics[trim={0px 0px 0px 0px},clip,width=0.319\textwidth]{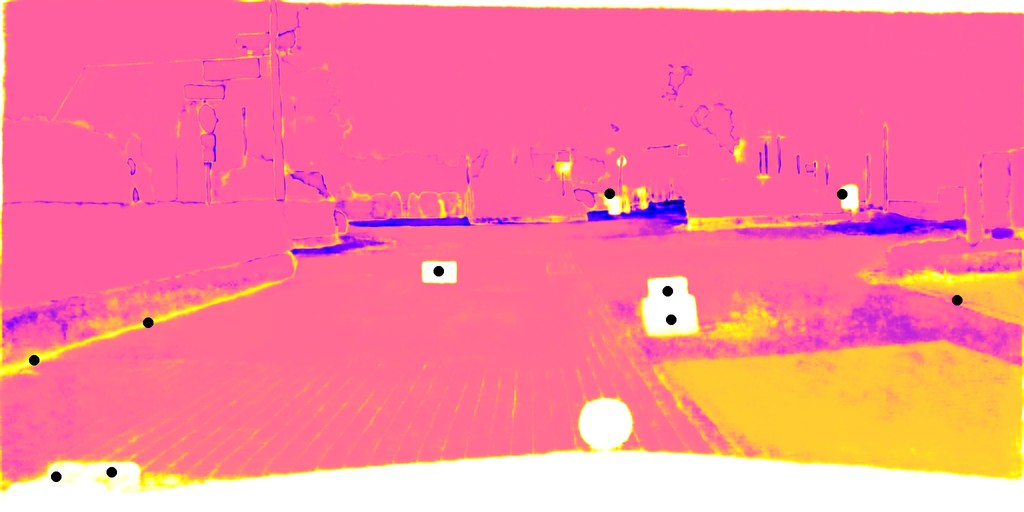}&%
    \includegraphics[trim={0px 0px 0px 0px},clip,width=0.319\textwidth]{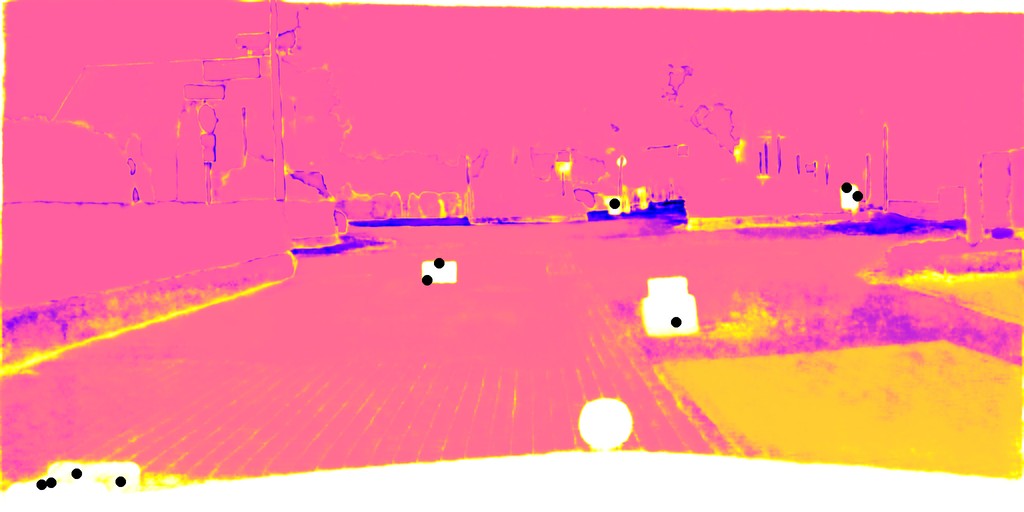}\\%

    \rotatebox{90}{\hspace{17pt} 20} &
    \includegraphics[trim={0px 0px 0px 0px},clip,width=0.319\textwidth]{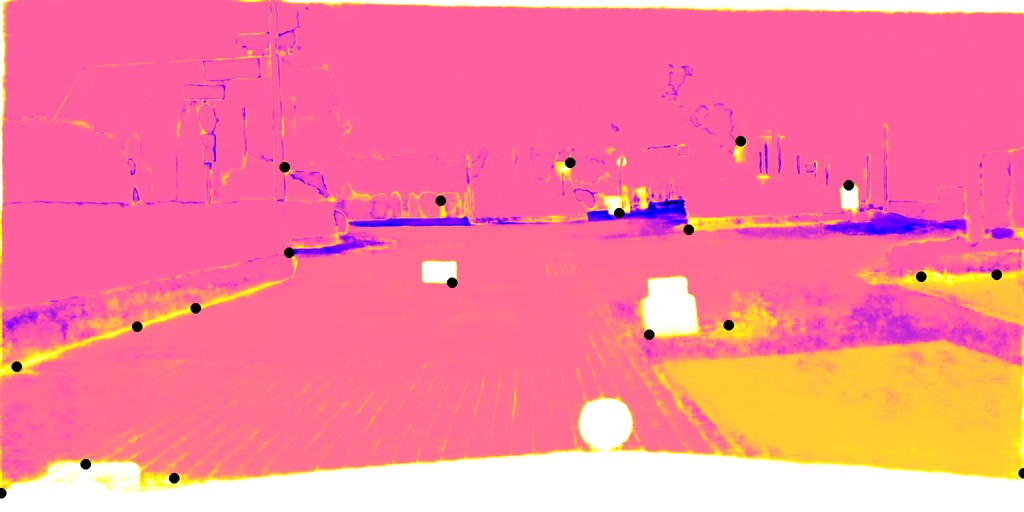}&%
    \includegraphics[trim={0px 0px 0px 0px},clip,width=0.319\textwidth]{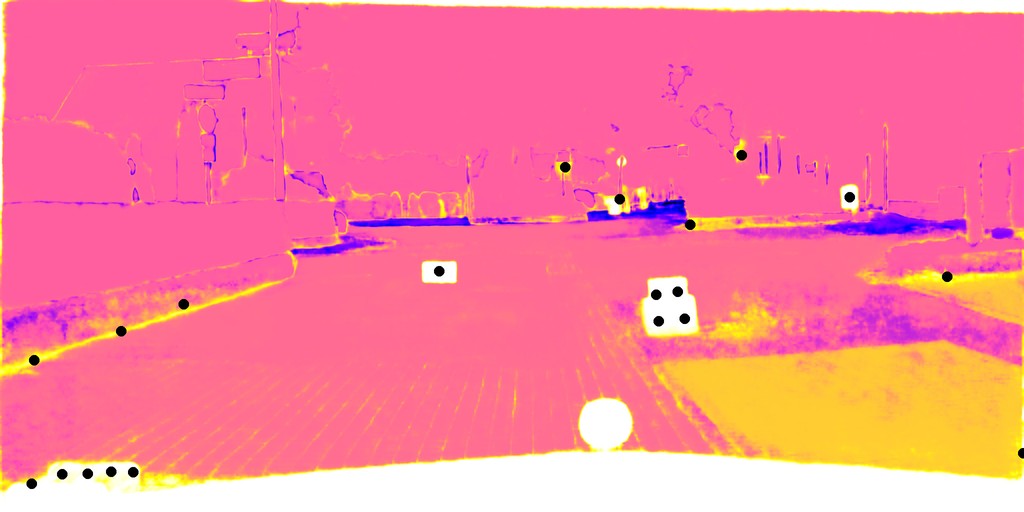}&%
    \includegraphics[trim={0px 0px 0px 0px},clip,width=0.319\textwidth]{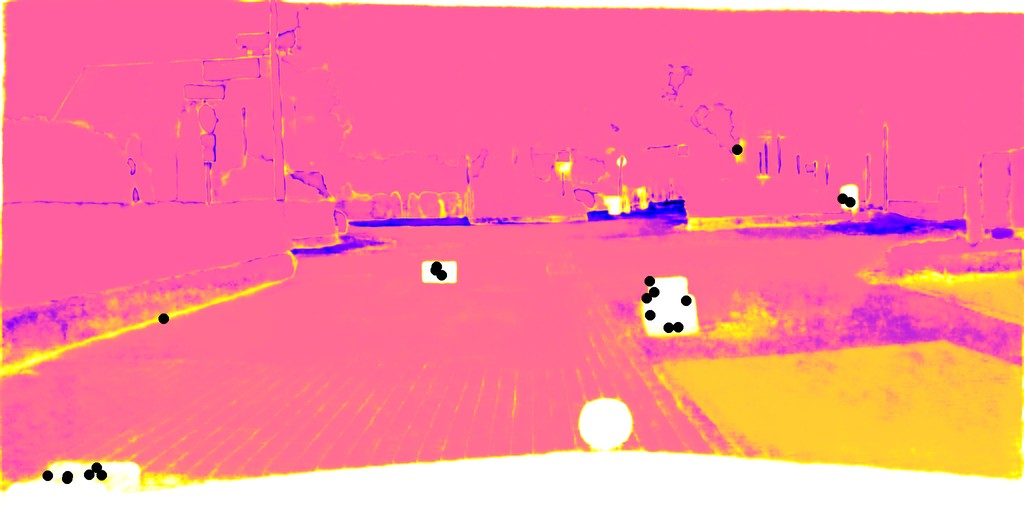}\\%

    \rotatebox{90}{\hspace{17pt} 50} &
    \includegraphics[trim={0px 0px 0px 0px},clip,width=0.319\textwidth]{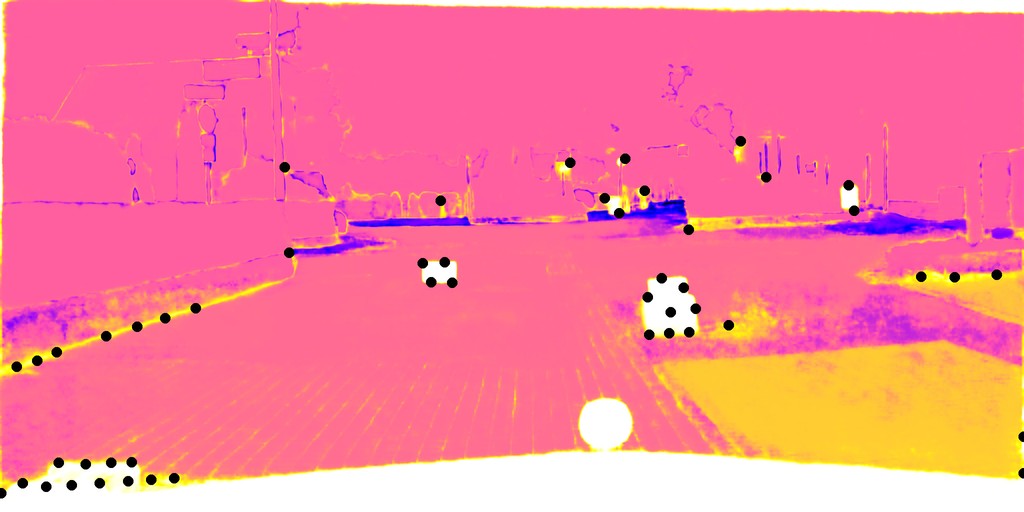}&%
    \includegraphics[trim={0px 0px 0px 0px},clip,width=0.319\textwidth]{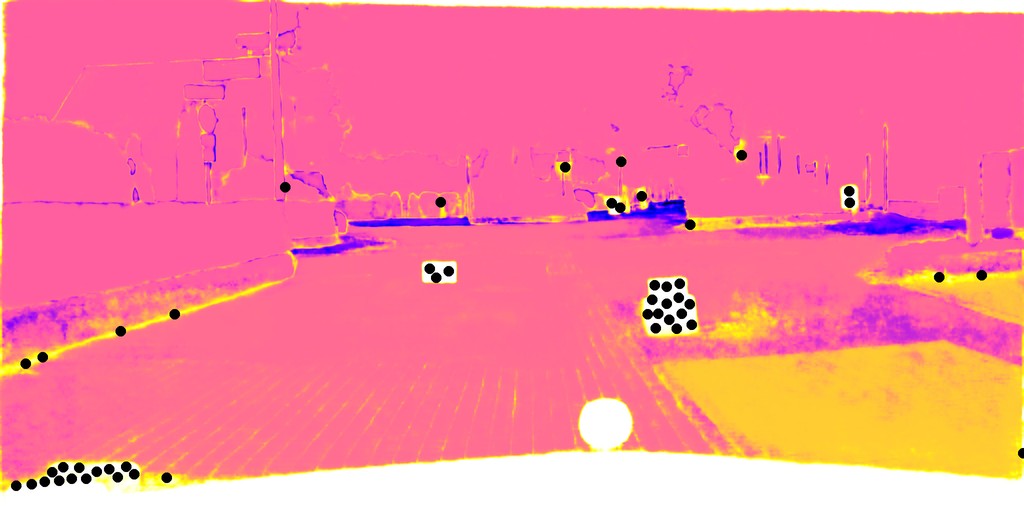}&%
    \includegraphics[trim={0px 0px 0px 0px},clip,width=0.319\textwidth]{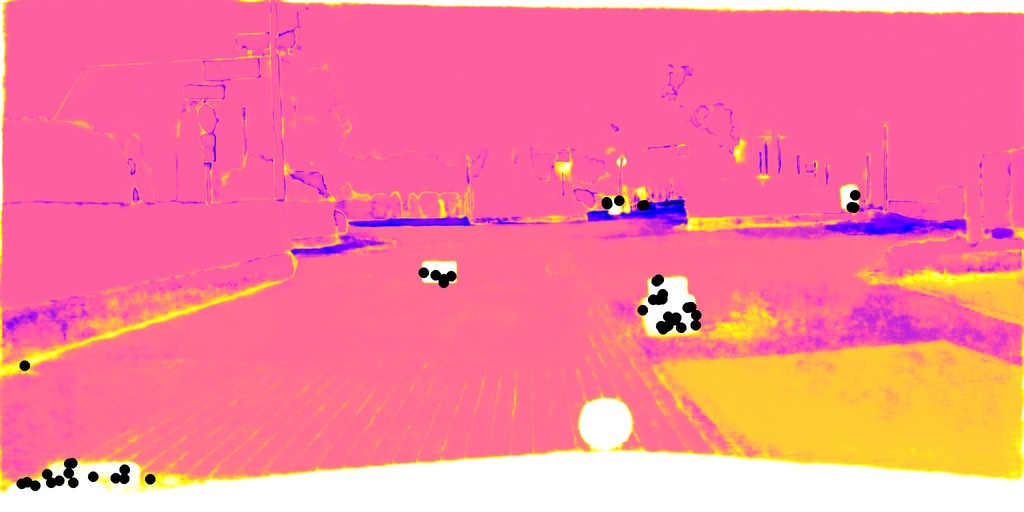}\\%

    \rotatebox{90}{\hspace{17pt} 100} &
    \includegraphics[trim={0px 0px 0px 0px},clip,width=0.319\textwidth]{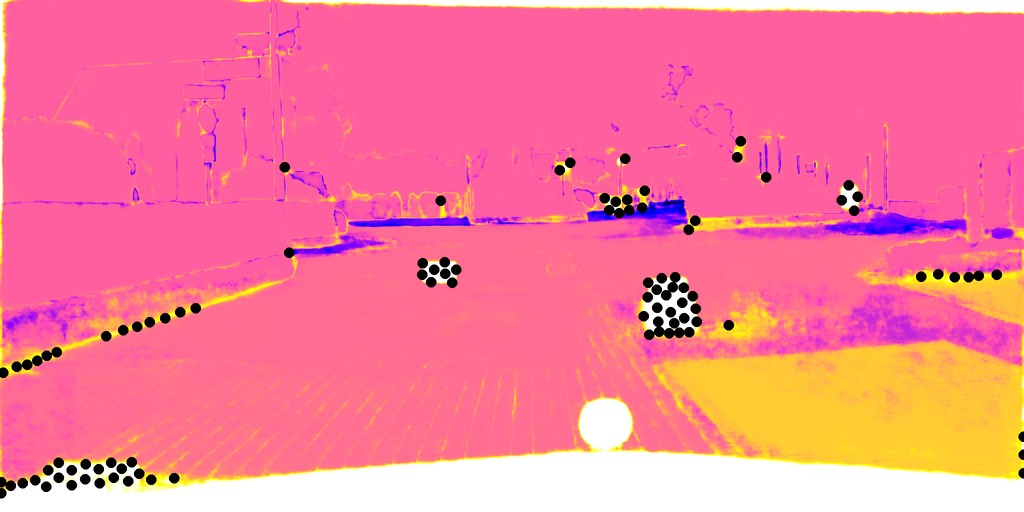}&%
    \includegraphics[trim={0px 0px 0px 0px},clip,width=0.319\textwidth]{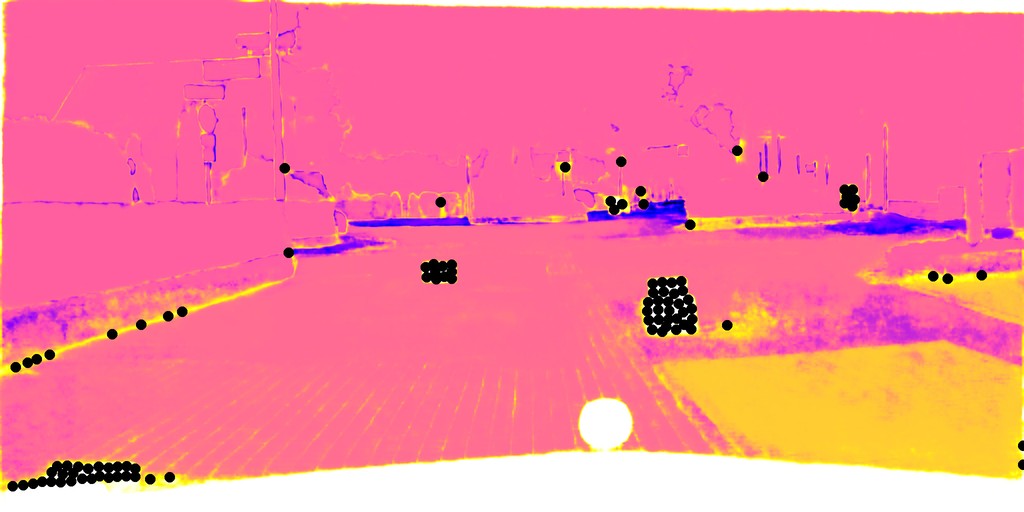}&%
    \includegraphics[trim={0px 0px 0px 0px},clip,width=0.319\textwidth]{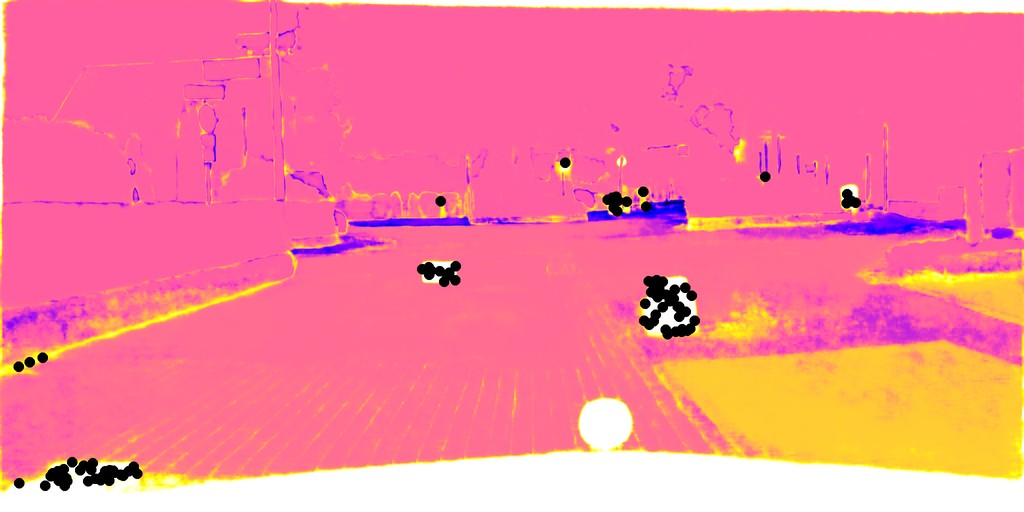}\\%
    
    & FPS & K-Means & Random Sampling \\
  \end{tabular}
  \caption{
      \textbf{Sampling Methods.}
      Using different numbers of sampled points, we show the visual performance of different sampling methods.
  }
  \label{fig:supp_sampling}
\end{figure}

\vspace{-5pt}
We provide visualizations of sampling methods with different numbers of points.
We show FPS, K-Means, and Random Sampling methods with $10$, $20$, $50$, and $100$ points on  the Figure \ref{fig:supp_sampling}.
Random Sampling generates points mostly in large regions, rarely on small objects;
K-Means provides mostly central sampling of anomalous regions;
while FPS provides a good balance between sampling on a large object and sampling from small regions.

\begin{figure}[t]
  \centering
  \begin{tabular}{ccc}
    \includegraphics[trim={0px 0px 0px 0px},clip,width=0.333\textwidth]{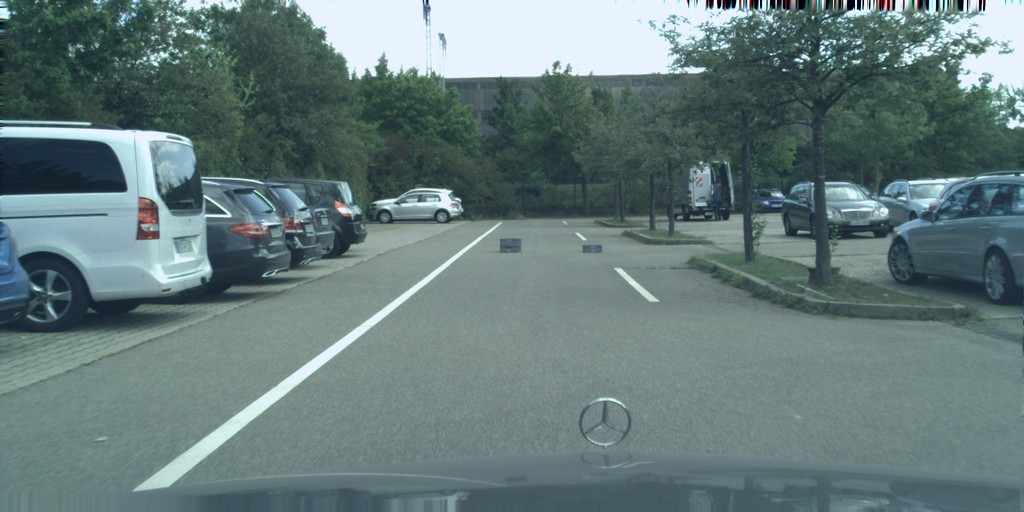}&%
    \includegraphics[trim={0px 0px 0px 0px},clip,width=0.333\textwidth]{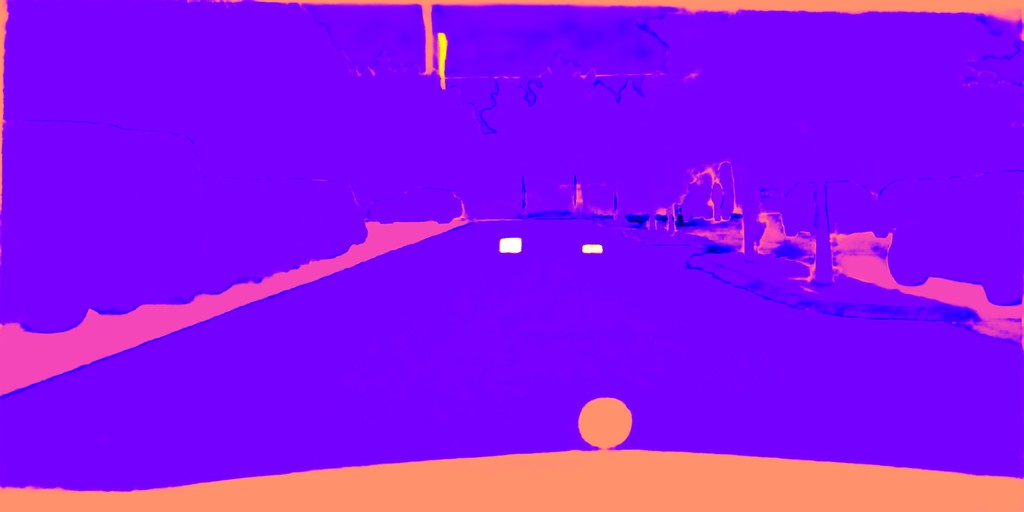}&%
    \includegraphics[trim={0px 0px 0px 0px},clip,width=0.333\textwidth]{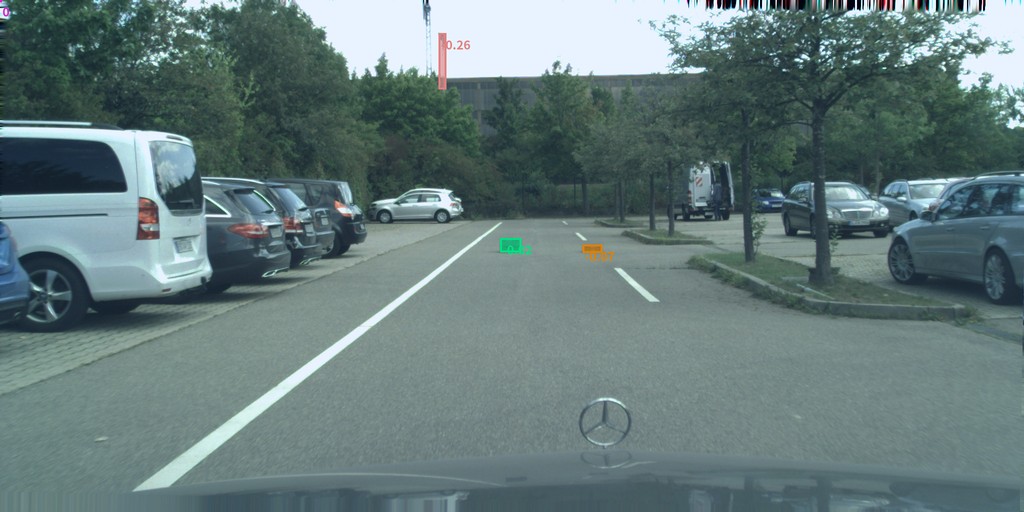}\\%

    \includegraphics[trim={0px 0px 0px 0px},clip,width=0.333\textwidth]{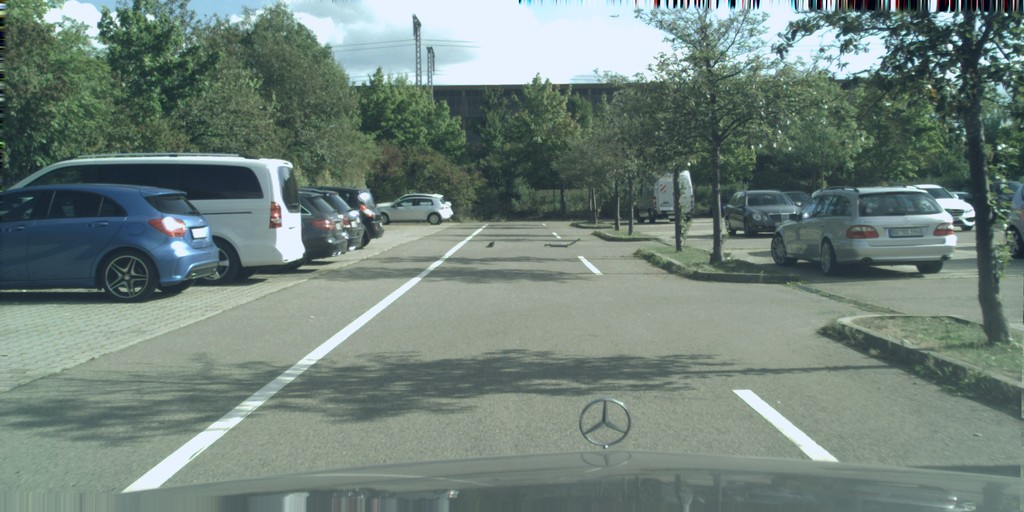}&%
    \includegraphics[trim={0px 0px 0px 0px},clip,width=0.333\textwidth]{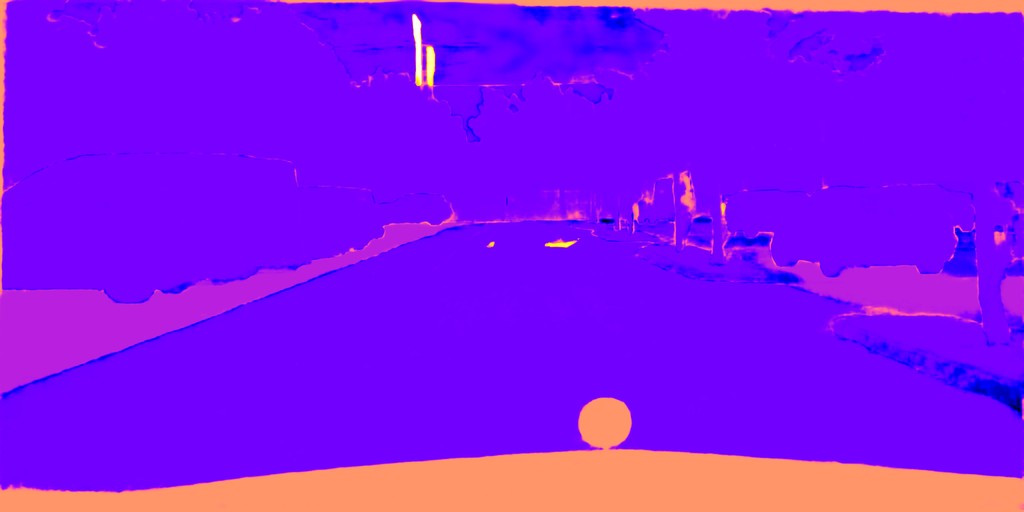}&%
    \includegraphics[trim={0px 0px 0px 0px},clip,width=0.333\textwidth]{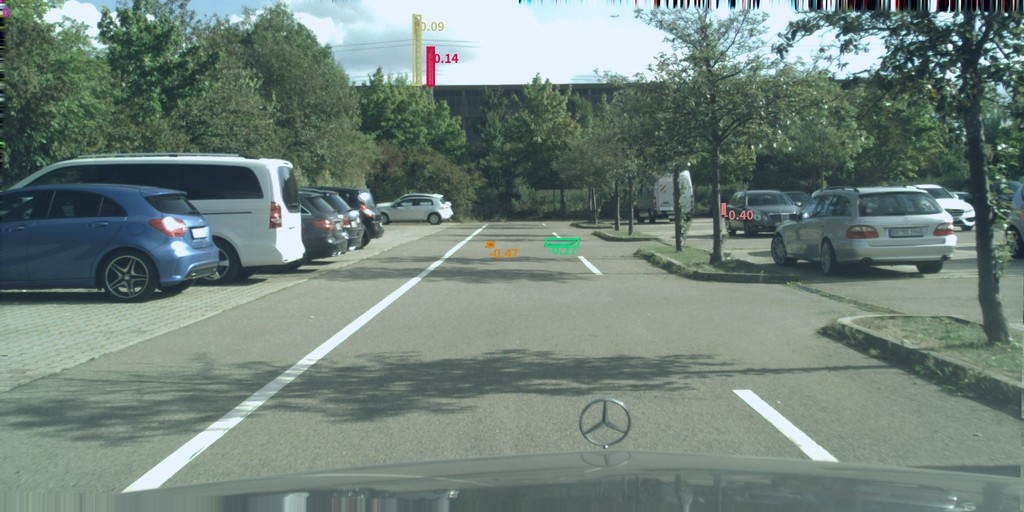}\\%
    
    \includegraphics[trim={0px 0px 0px 0px},clip,width=0.333\textwidth]{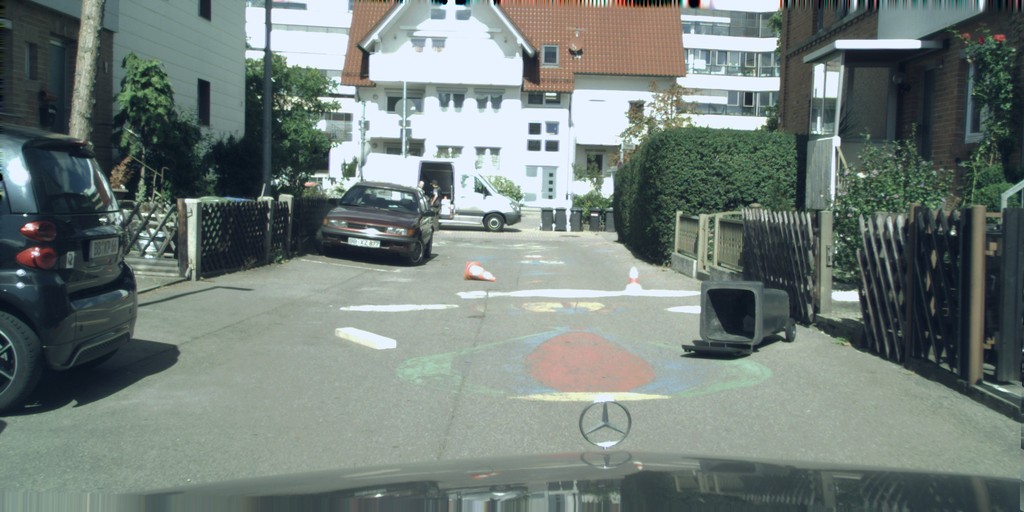}&%
    \includegraphics[trim={0px 0px 0px 0px},clip,width=0.333\textwidth]{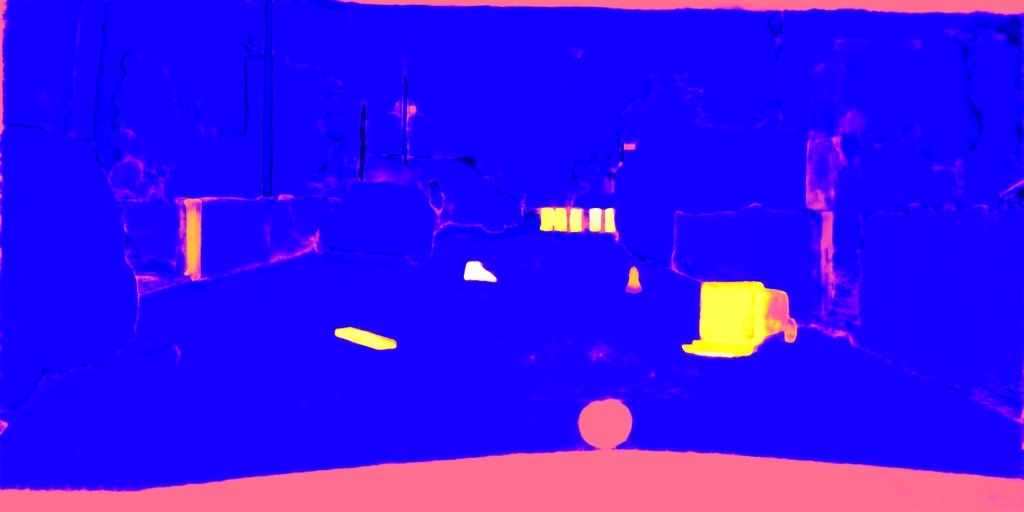}&%
    \includegraphics[trim={0px 0px 0px 0px},clip,width=0.333\textwidth]{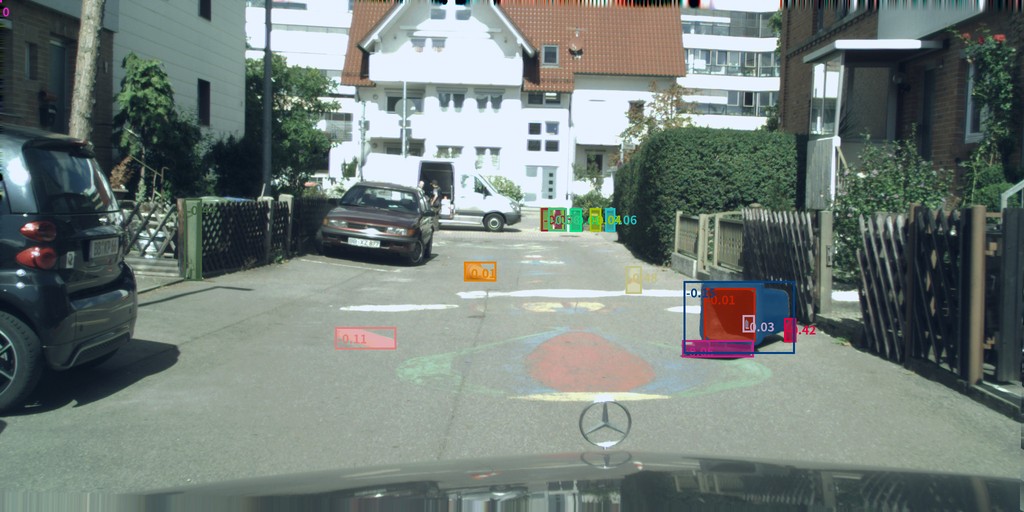}\\%
    
    \includegraphics[trim={0px 0px 0px 0px},clip,width=0.333\textwidth]{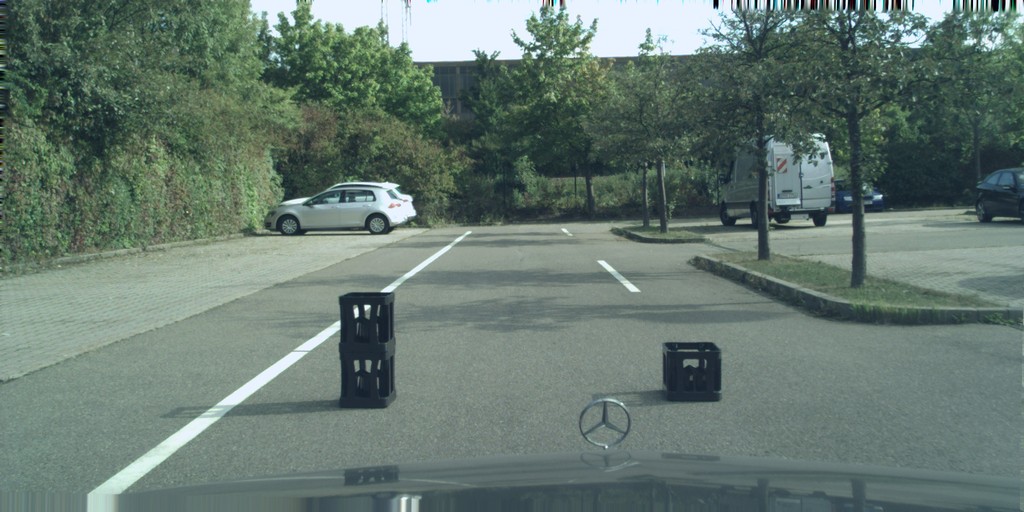}&%
    \includegraphics[trim={0px 0px 0px 0px},clip,width=0.333\textwidth]{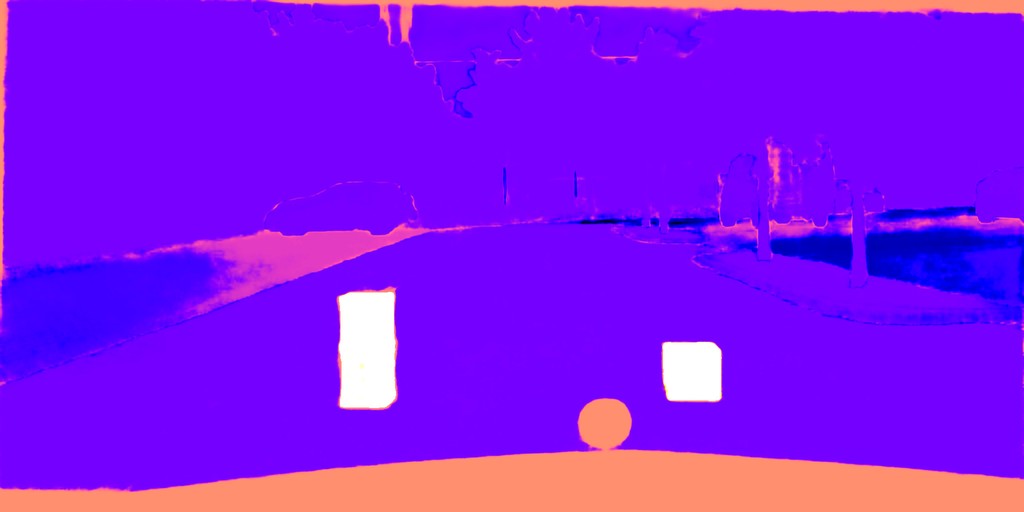}&%
    \includegraphics[trim={0px 0px 0px 0px},clip,width=0.333\textwidth]{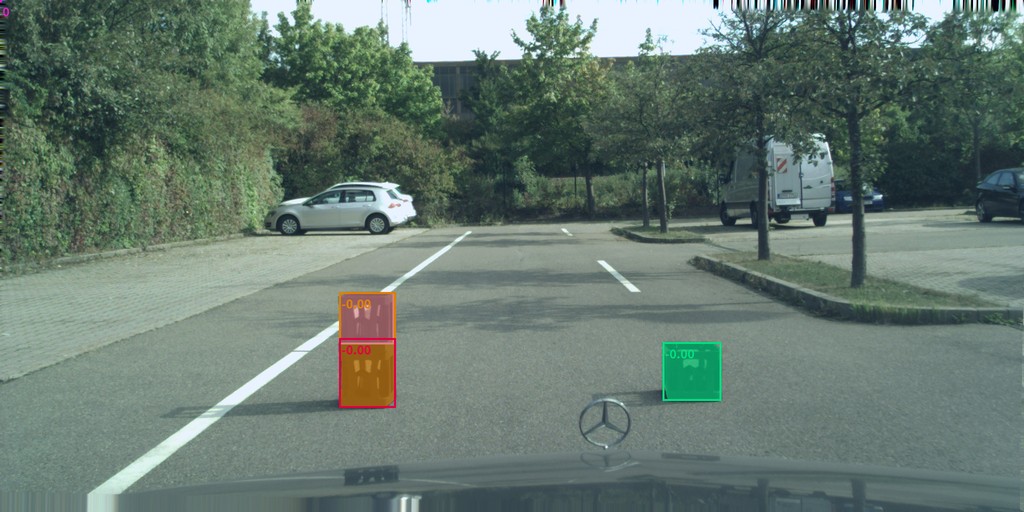}\\%

    \includegraphics[trim={0px 0px 0px 0px},clip,width=0.333\textwidth]{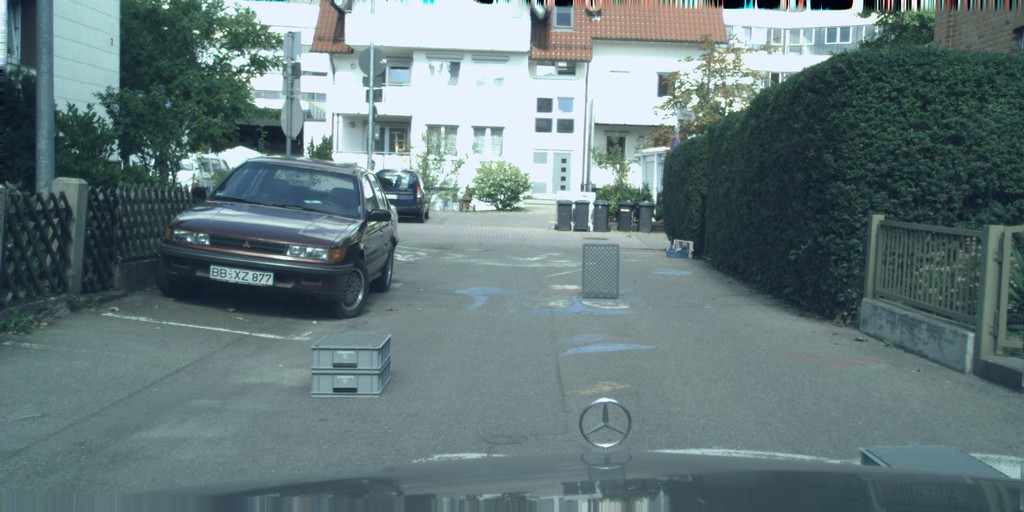}&%
    \includegraphics[trim={0px 0px 0px 0px},clip,width=0.333\textwidth]{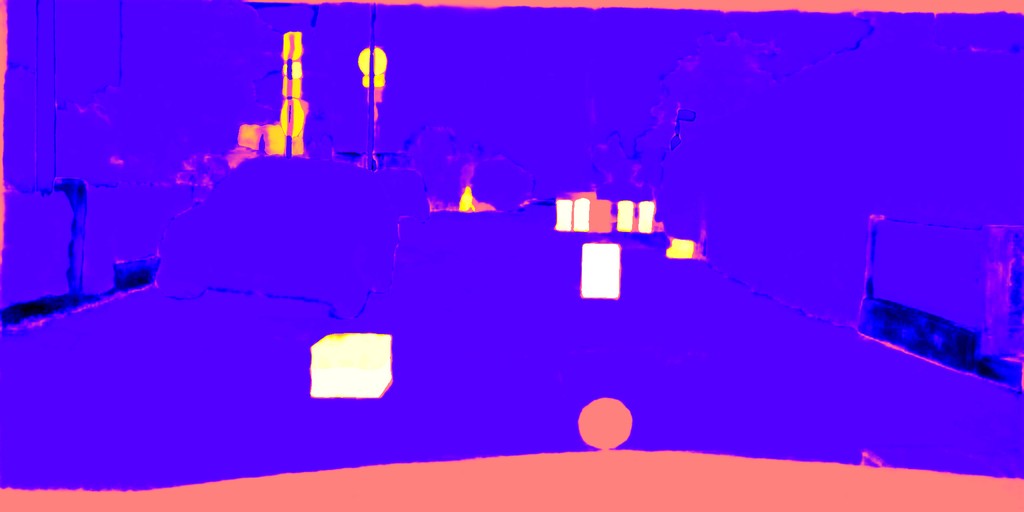}&%
    \includegraphics[trim={0px 0px 0px 0px},clip,width=0.333\textwidth]{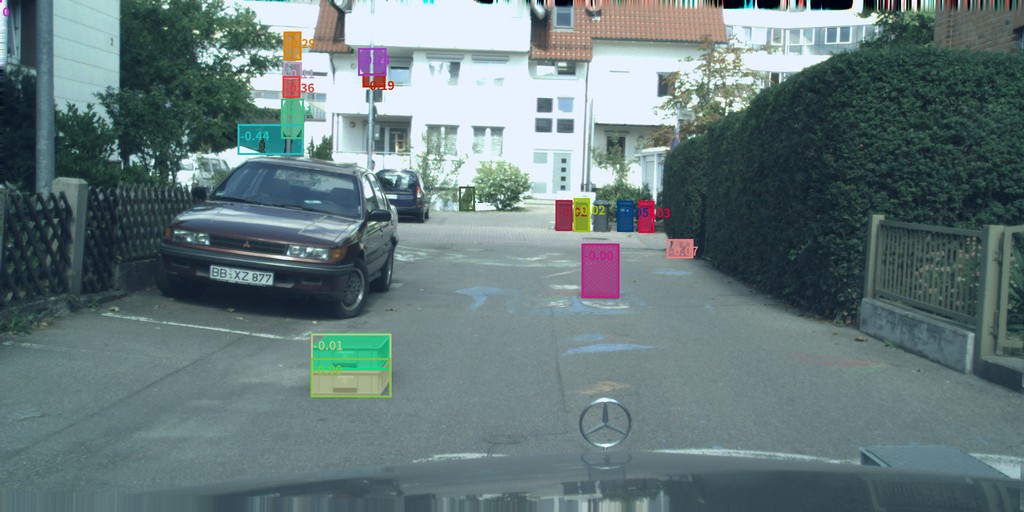}\\%

    \includegraphics[trim={0px 0px 0px 0px},clip,width=0.333\textwidth]{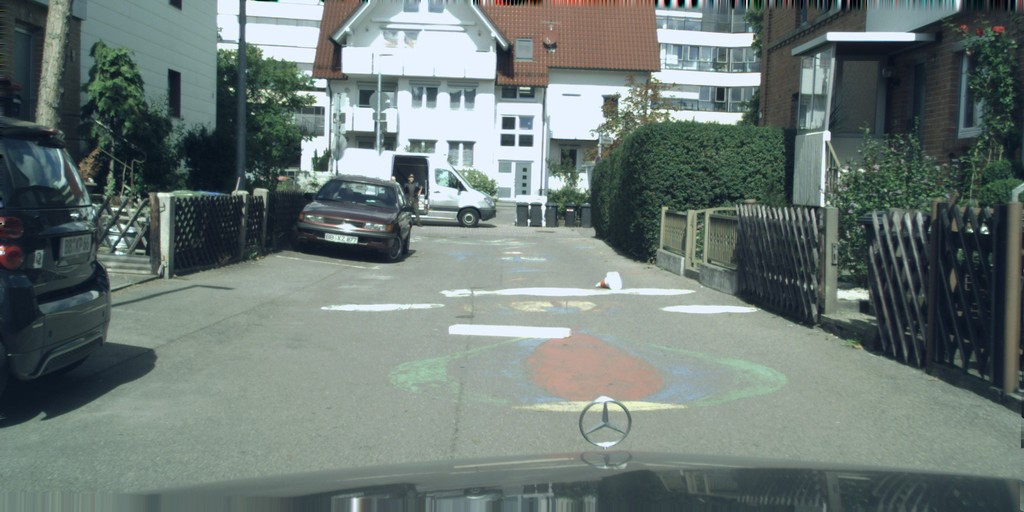}&%
    \includegraphics[trim={0px 0px 0px 0px},clip,width=0.333\textwidth]{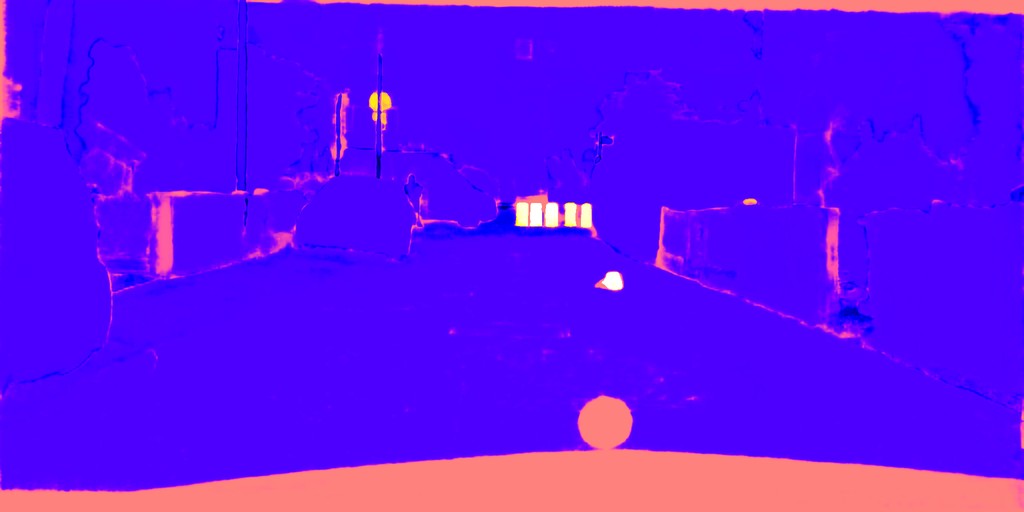}&%
    \includegraphics[trim={0px 0px 0px 0px},clip,width=0.333\textwidth]{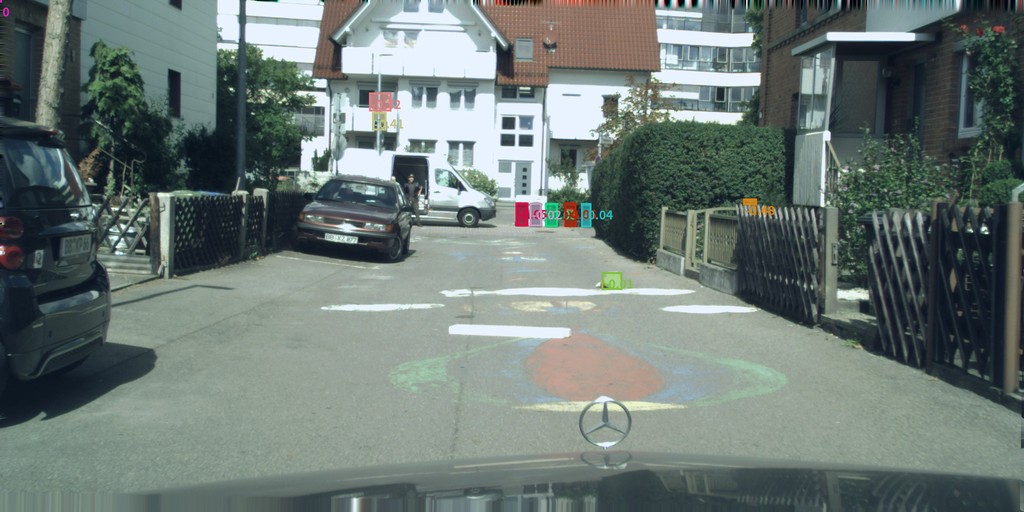}\\%

    \includegraphics[trim={0px 0px 0px 0px},clip,width=0.333\textwidth]{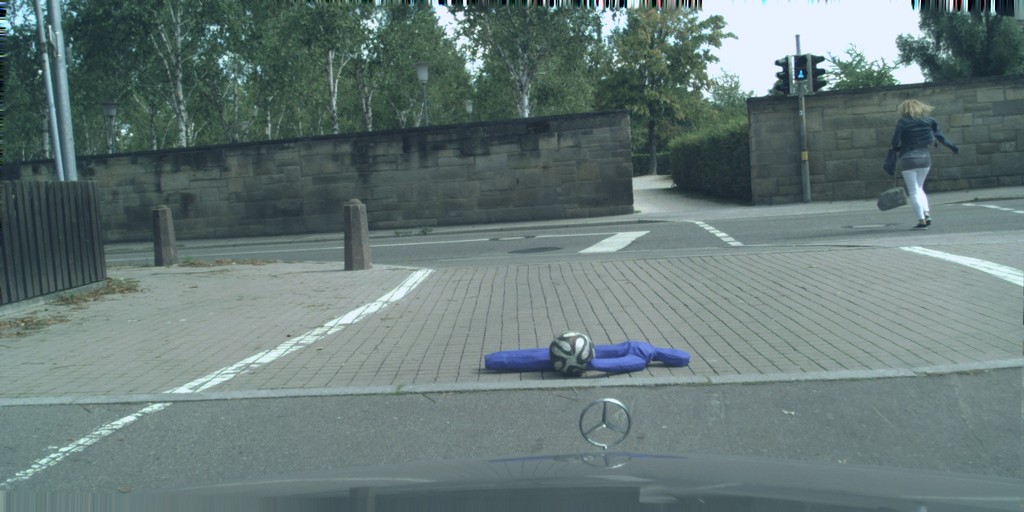}&%
    \includegraphics[trim={0px 0px 0px 0px},clip,width=0.333\textwidth]{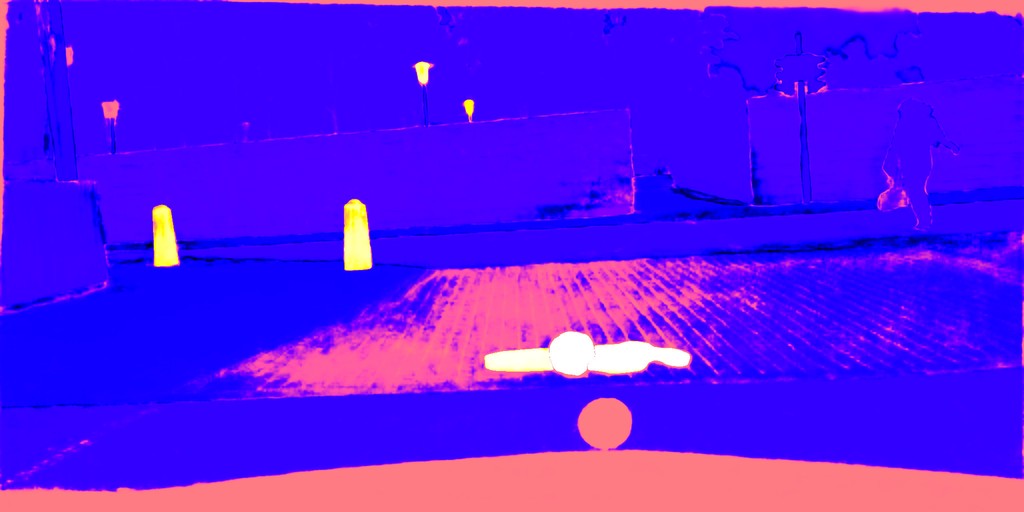}&%
    \includegraphics[trim={0px 0px 0px 0px},clip,width=0.333\textwidth]{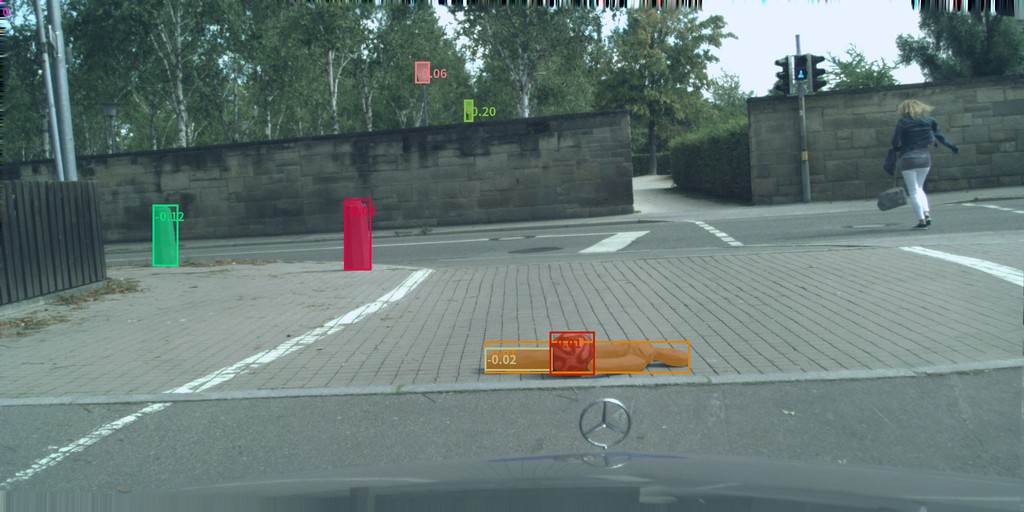}\\%

    Image & Anomaly scores & Instance prediction \\
  \end{tabular}
  \caption{
  \textbf{Qualitative Results} on the Fishyscapes Lost and Found dataset \cite{blum2021fishyscapes}.
  }
  \label{fig:supp_fslaf}
\end{figure}

\section{Qualitative Results} 
\begin{figure}[t]
  \centering
  \begin{tabular}{ccc}
    \includegraphics[trim={0px 0px 0px 0px},clip,width=0.333\textwidth]{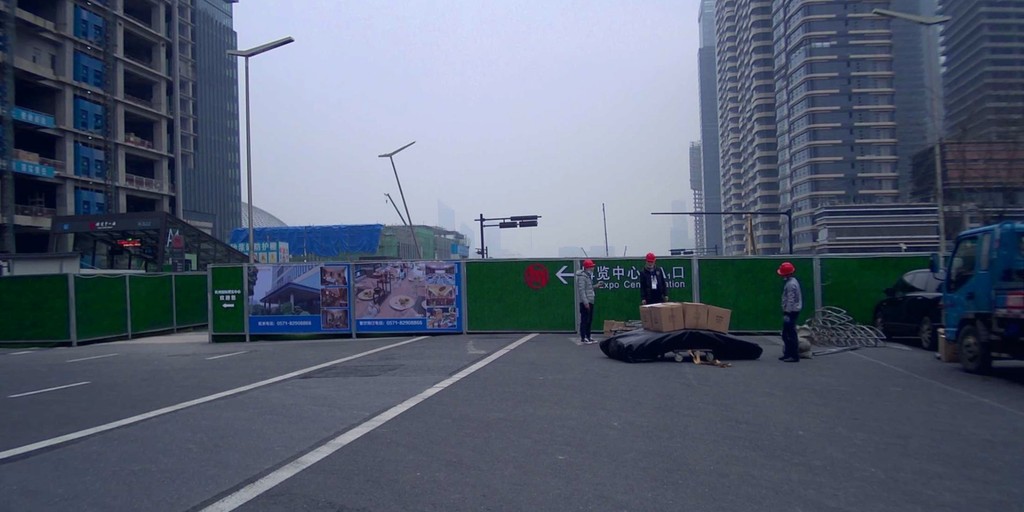}&%
    \includegraphics[trim={0px 0px 0px 0px},clip,width=0.333\textwidth]{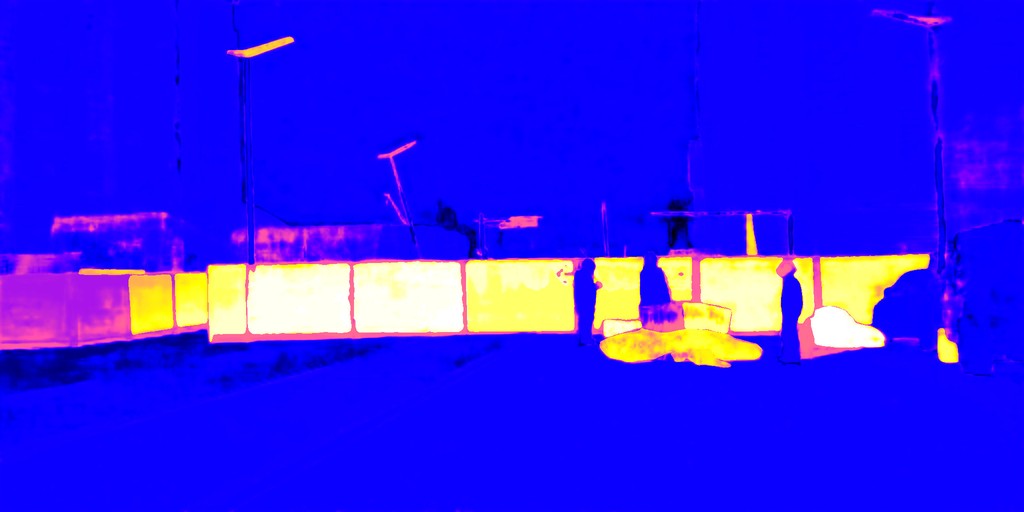}&%
    \includegraphics[trim={0px 0px 0px 0px},clip,width=0.333\textwidth]{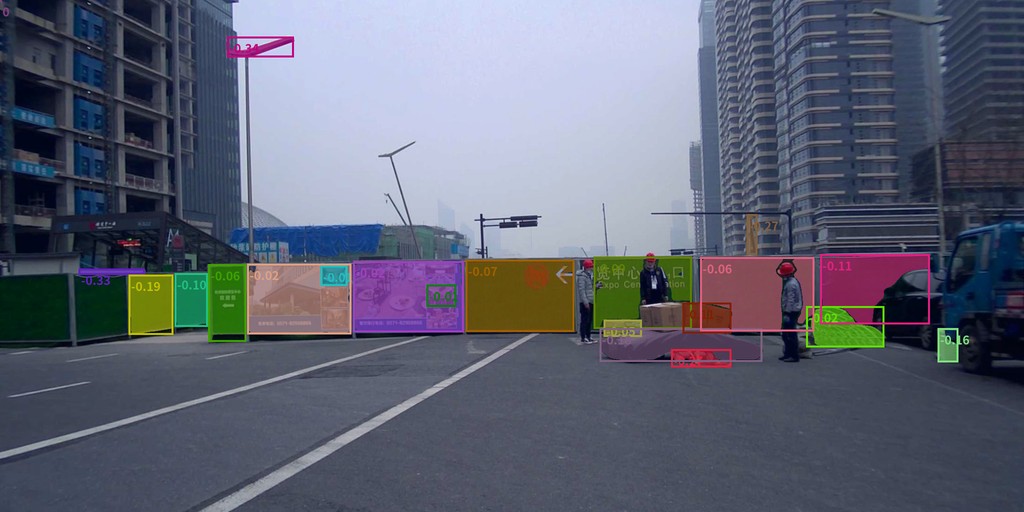}\\%

    \includegraphics[trim={0px 0px 0px 0px},clip,width=0.333\textwidth]{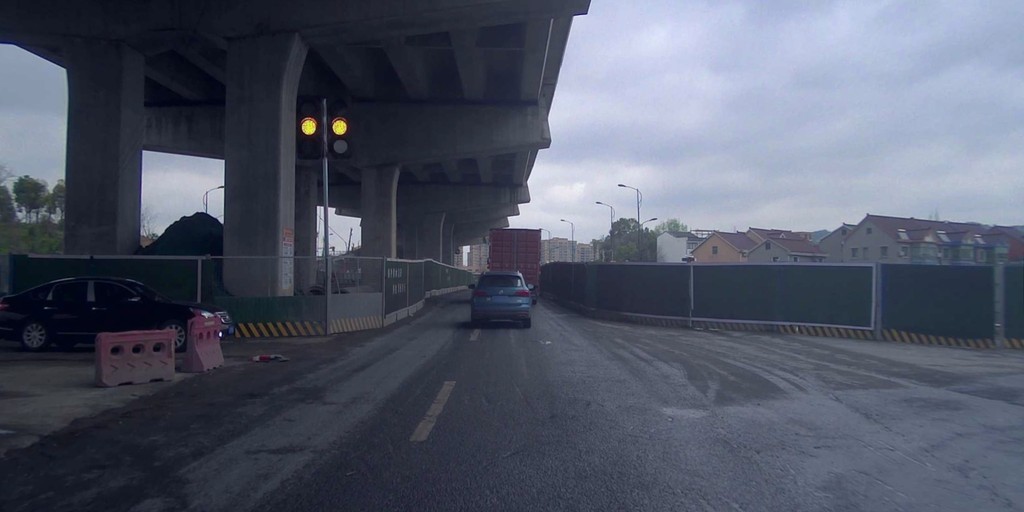}&%
    \includegraphics[trim={0px 0px 0px 0px},clip,width=0.333\textwidth]{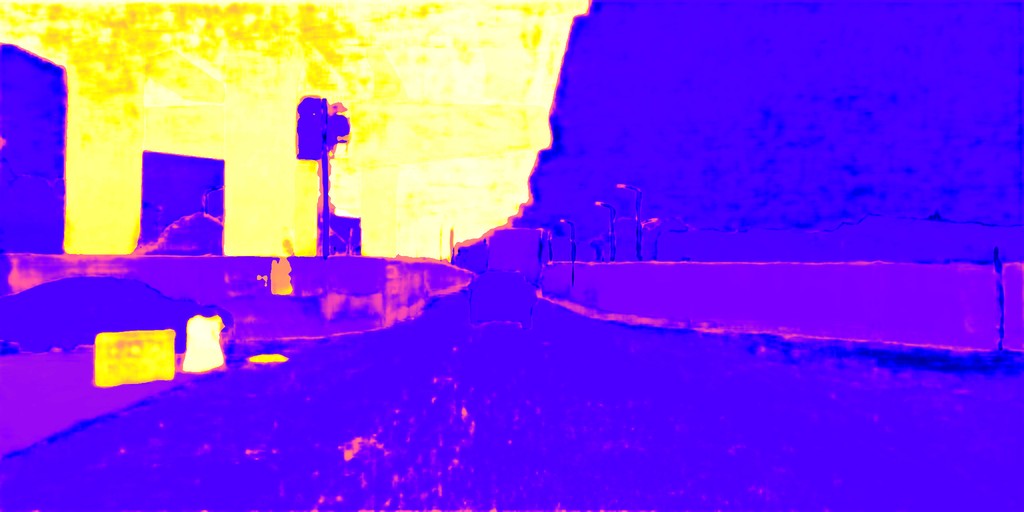}&%
    \includegraphics[trim={0px 0px 0px 0px},clip,width=0.333\textwidth]{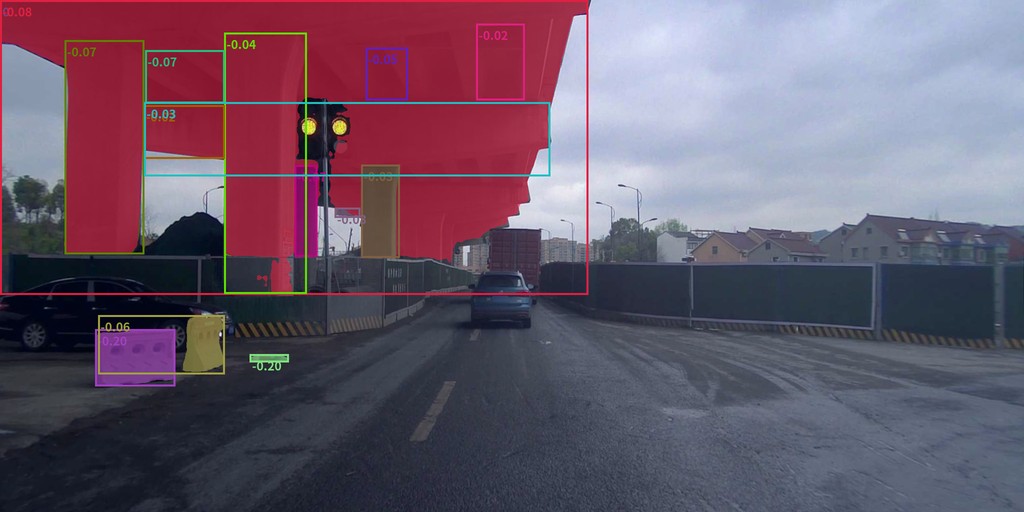}\\%

    \includegraphics[trim={0px 0px 0px 0px},clip,width=0.333\textwidth]{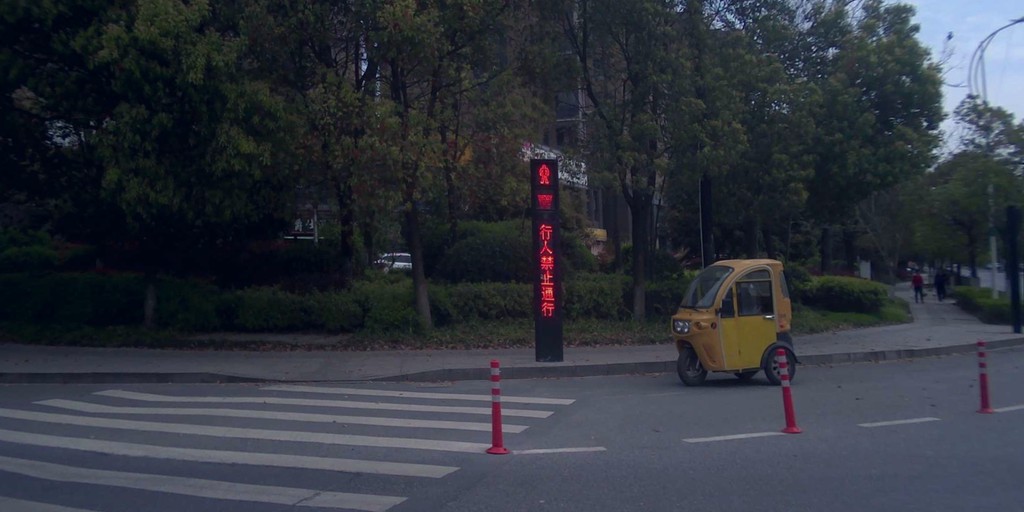}&%
    \includegraphics[trim={0px 0px 0px 0px},clip,width=0.333\textwidth]{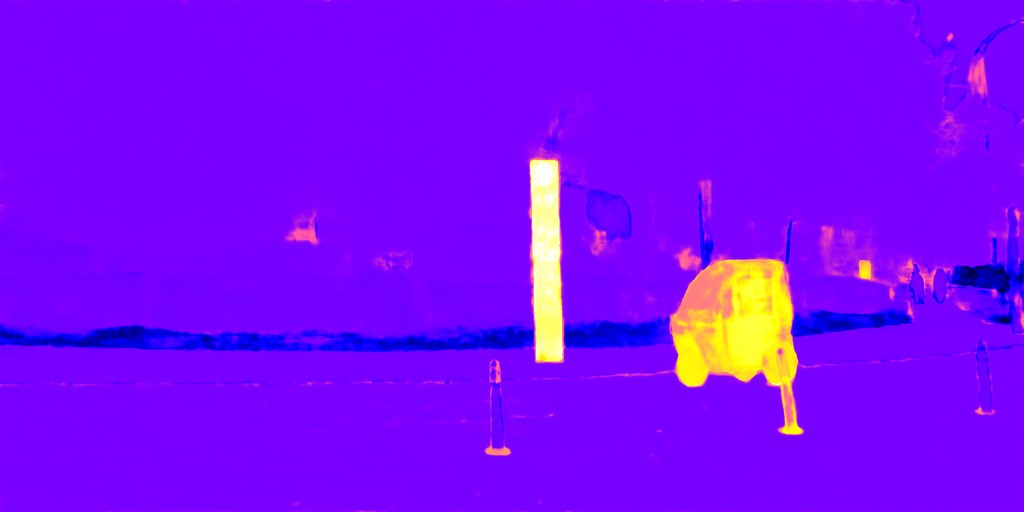}&%
    \includegraphics[trim={0px 0px 0px 0px},clip,width=0.333\textwidth]{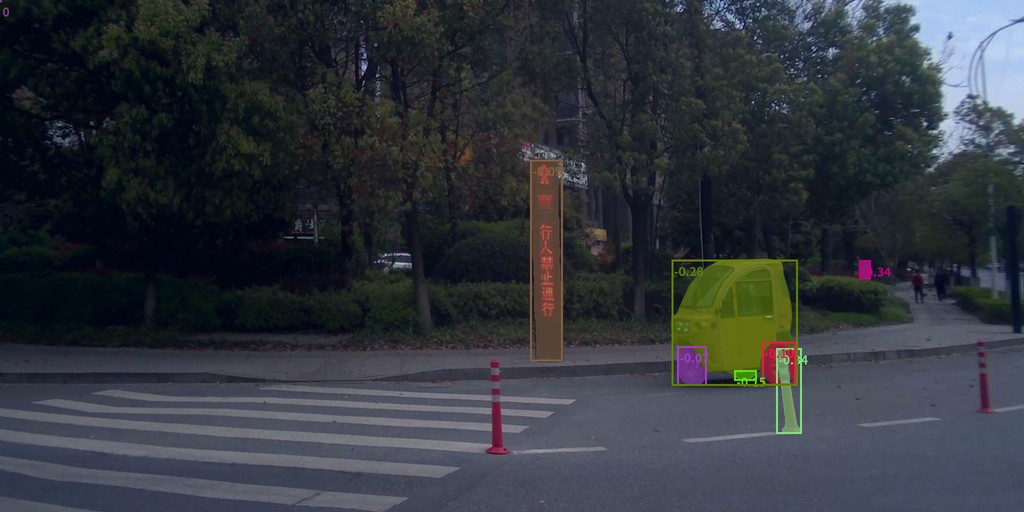}\\%

    \includegraphics[trim={0px 0px 0px 0px},clip,width=0.333\textwidth]{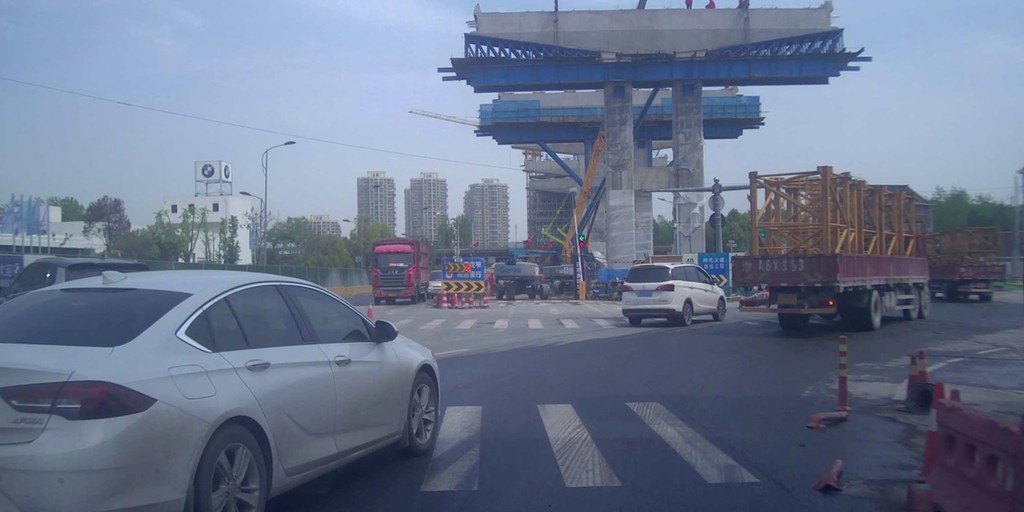}&%
    \includegraphics[trim={0px 0px 0px 0px},clip,width=0.333\textwidth]{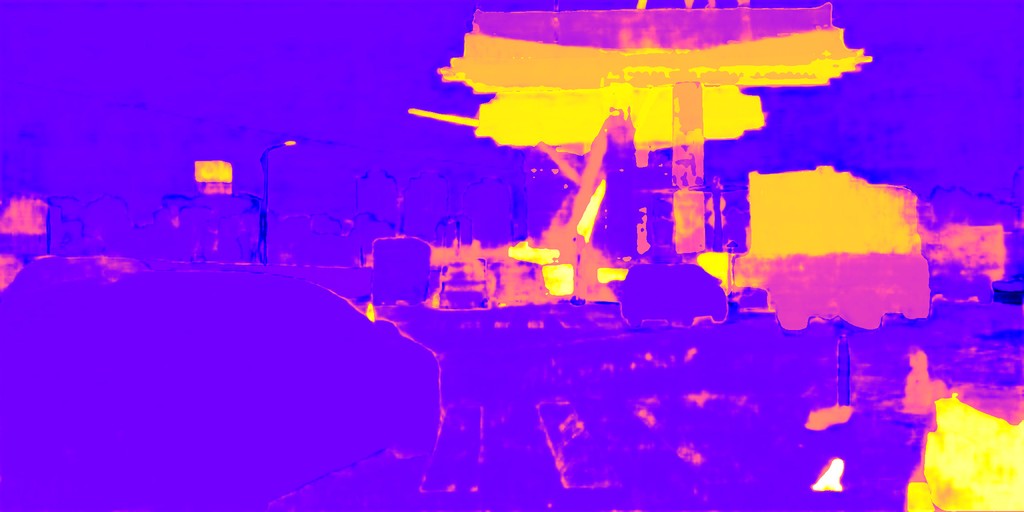}&%
    \includegraphics[trim={0px 0px 0px 0px},clip,width=0.333\textwidth]{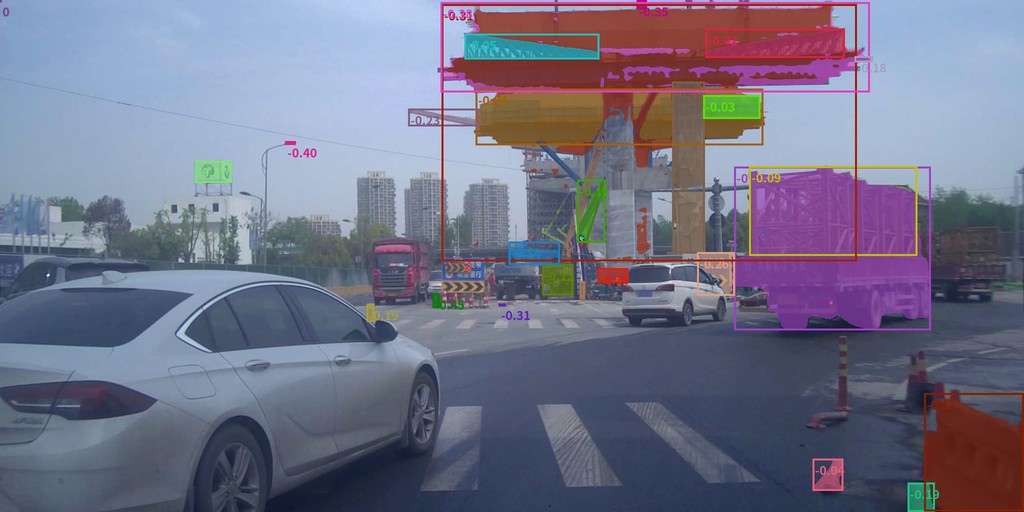}\\%

    \includegraphics[trim={0px 0px 0px 0px},clip,width=0.333\textwidth]{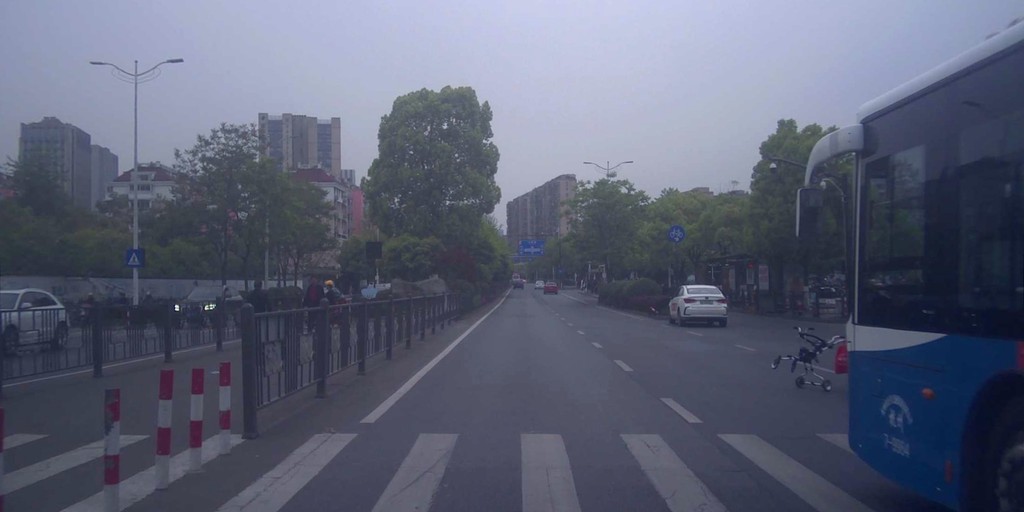}&%
    \includegraphics[trim={0px 0px 0px 0px},clip,width=0.333\textwidth]{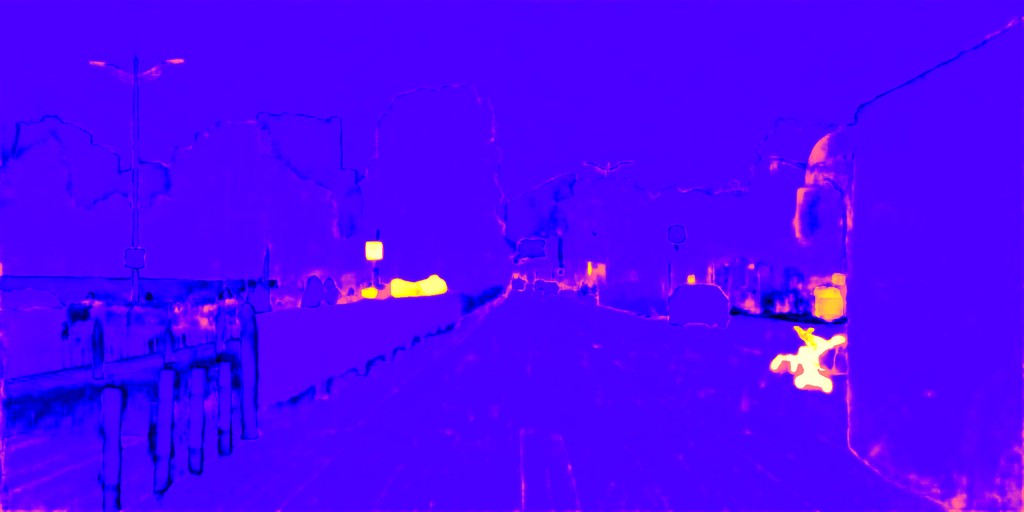}&%
    \includegraphics[trim={0px 0px 0px 0px},clip,width=0.333\textwidth]{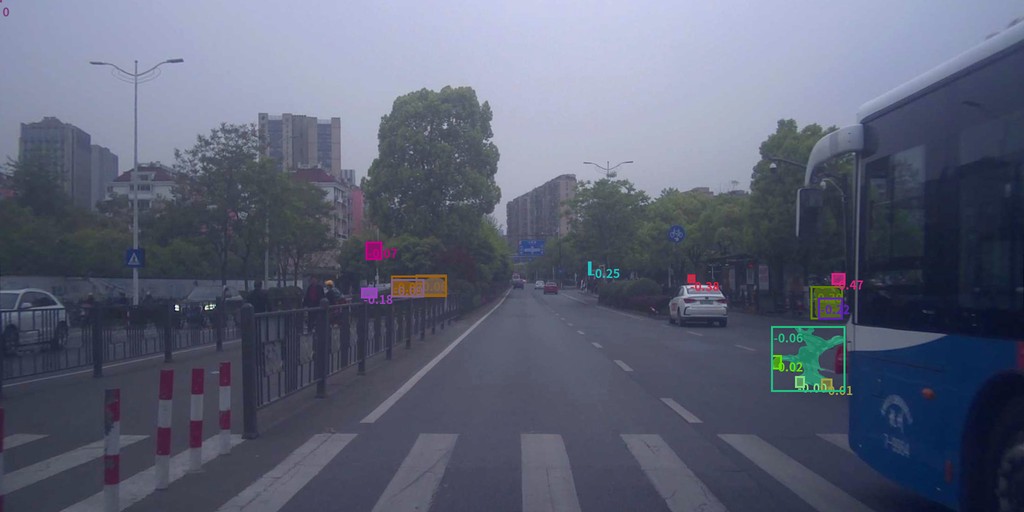}\\%

    \includegraphics[trim={0px 0px 0px 0px},clip,width=0.333\textwidth]{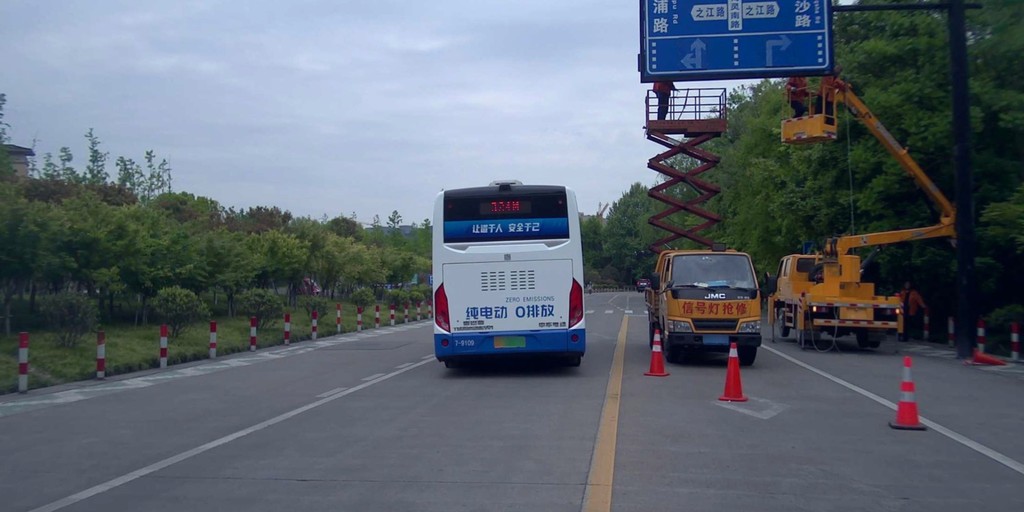}&%
    \includegraphics[trim={0px 0px 0px 0px},clip,width=0.333\textwidth]{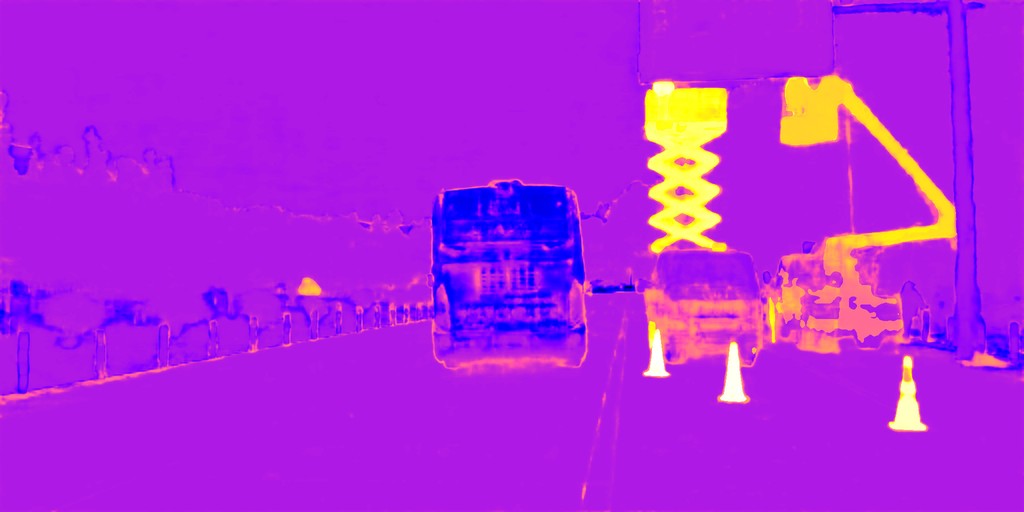}&%
    \includegraphics[trim={0px 0px 0px 0px},clip,width=0.333\textwidth]{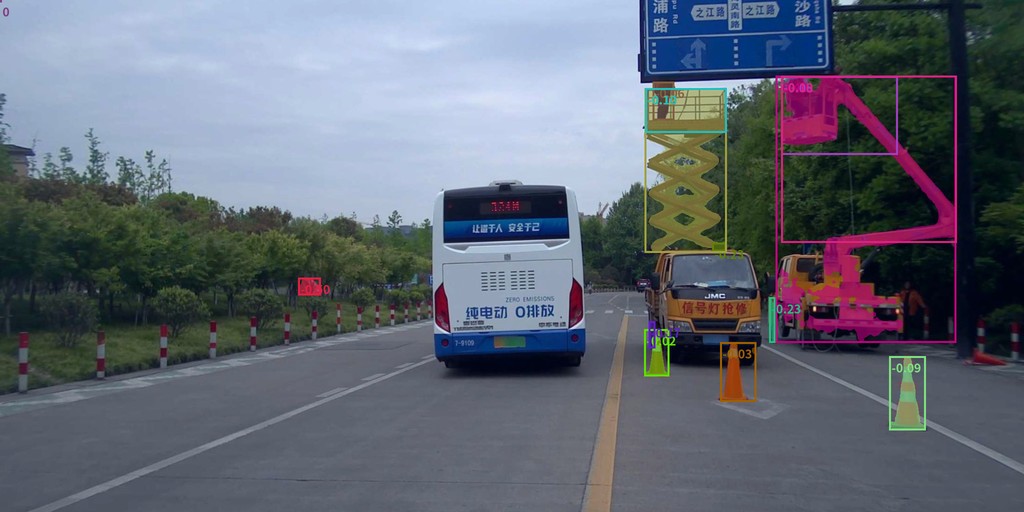}\\%

    \includegraphics[trim={0px 0px 0px 0px},clip,width=0.333\textwidth]{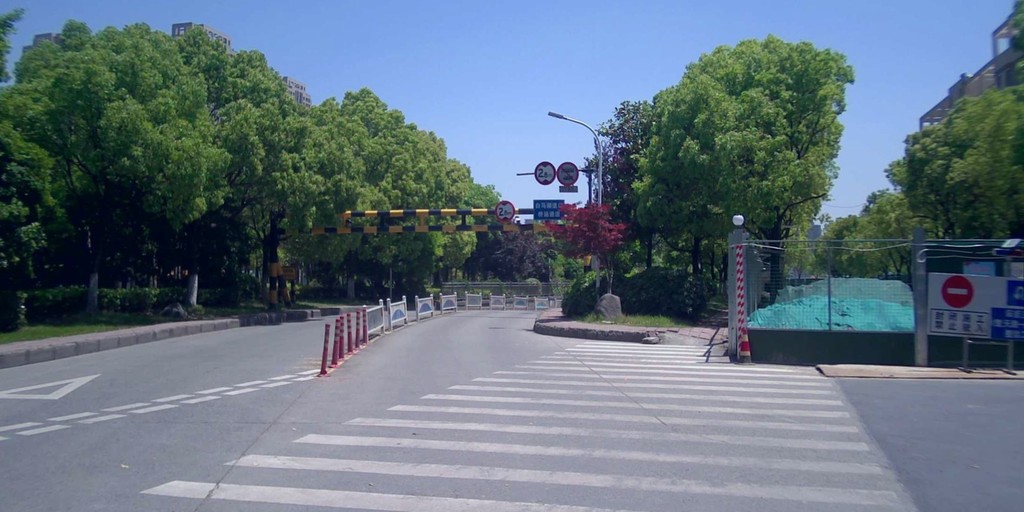}&%
    \includegraphics[trim={0px 0px 0px 0px},clip,width=0.333\textwidth]{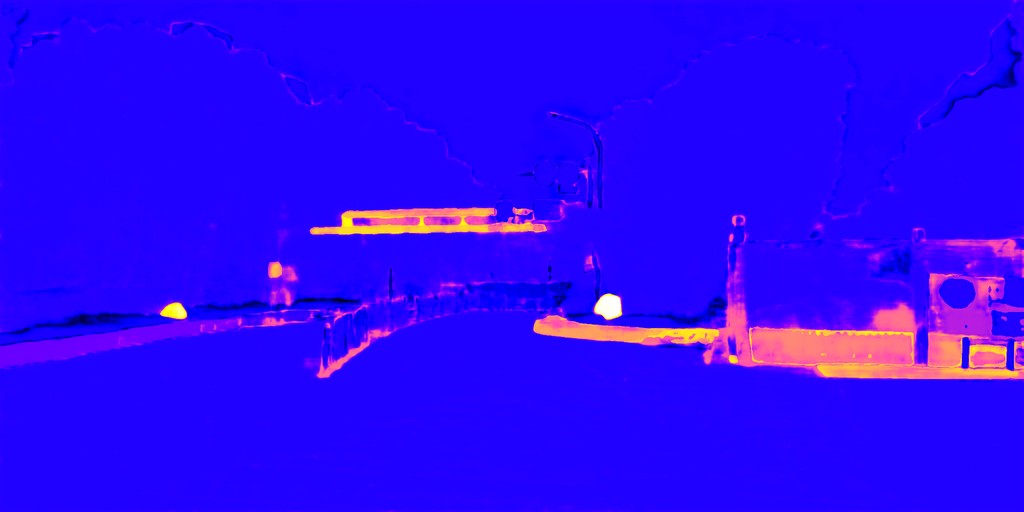}&%
    \includegraphics[trim={0px 0px 0px 0px},clip,width=0.333\textwidth]{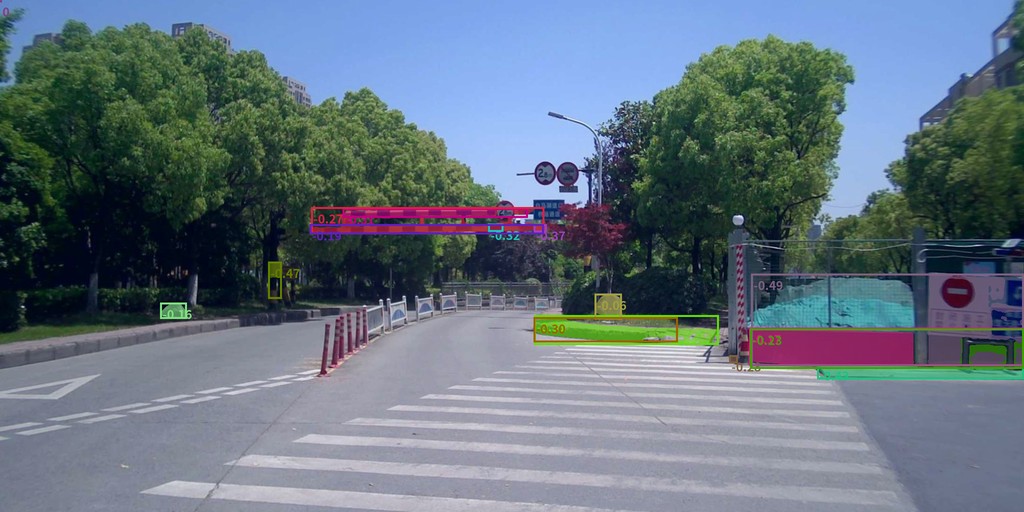}\\%

    Image & Anomaly scores & Instance prediction \\
  \end{tabular}
  \caption{
  \textbf{Qualitative Results} on the CODA dataset \cite{li2022coda}.
  }
  \label{fig:supp_coda}
\end{figure}

\begin{figure}[t]
  \centering
  \begin{tabular}{ccc}
    \includegraphics[trim={0px 0px 0px 0px},clip,width=0.333\textwidth]{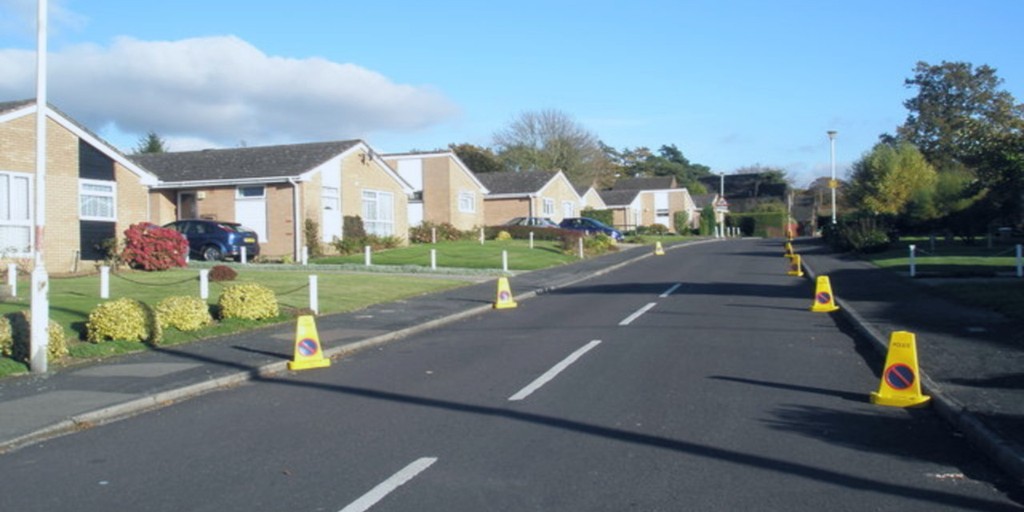}&%
    \includegraphics[trim={0px 0px 0px 0px},clip,width=0.333\textwidth]{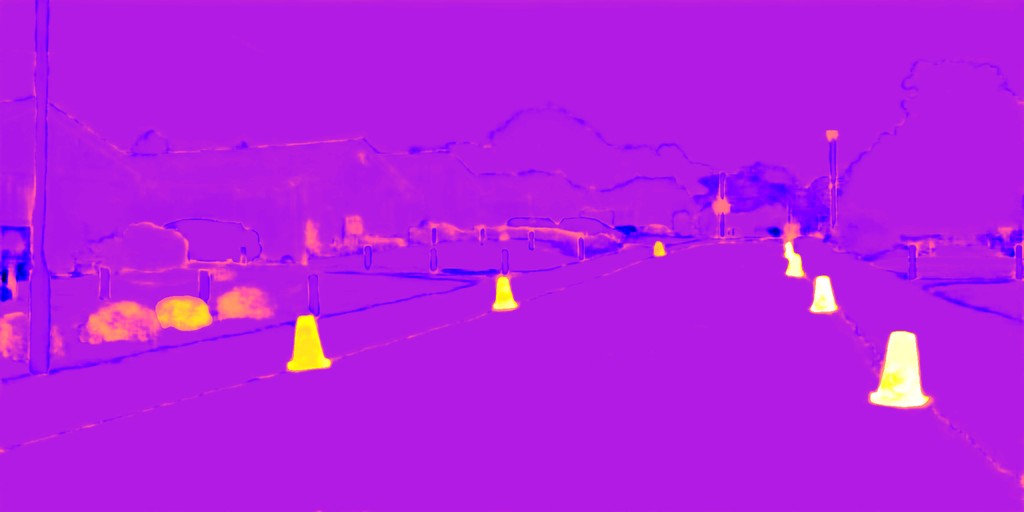}&%
    \includegraphics[trim={0px 0px 0px 0px},clip,width=0.333\textwidth]{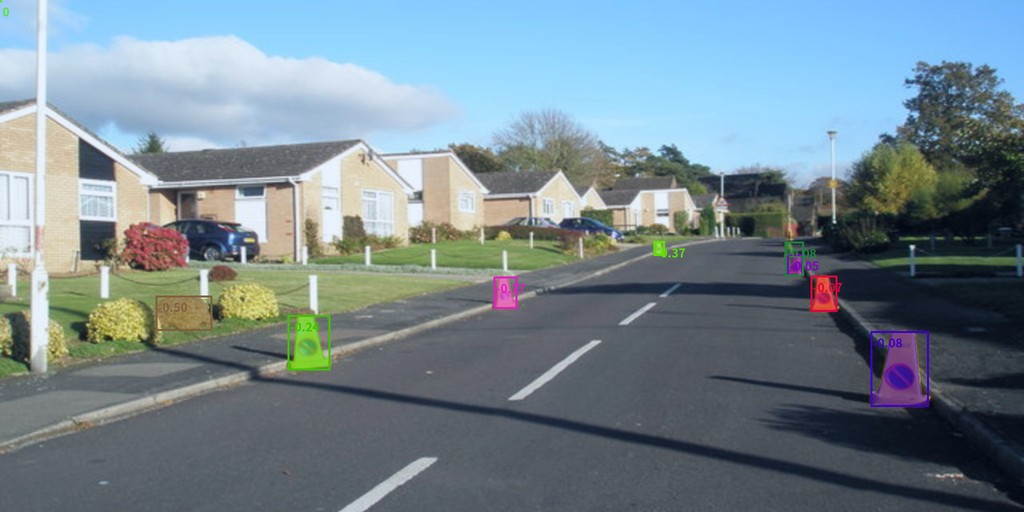}\\%

    \includegraphics[trim={0px 0px 0px 0px},clip,width=0.333\textwidth]{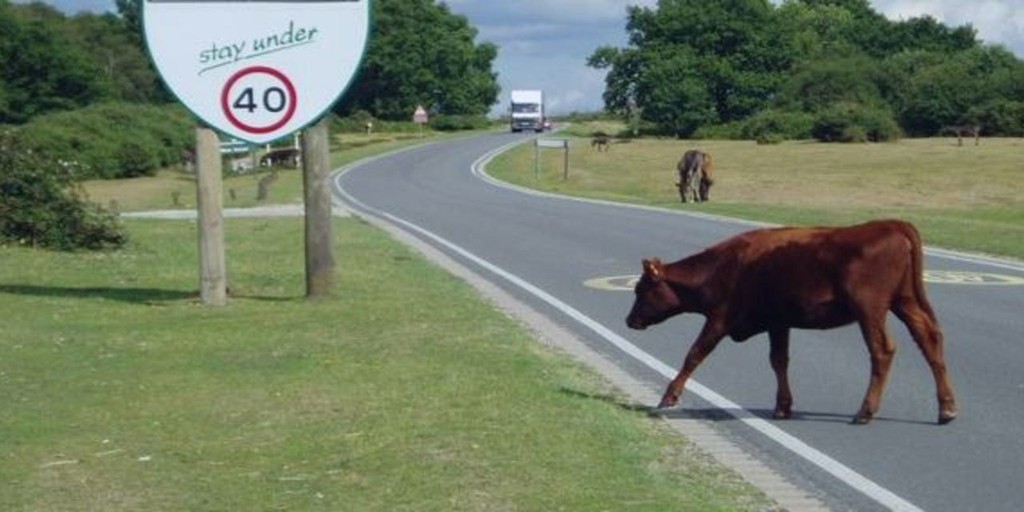}&%
    \includegraphics[trim={0px 0px 0px 0px},clip,width=0.333\textwidth]{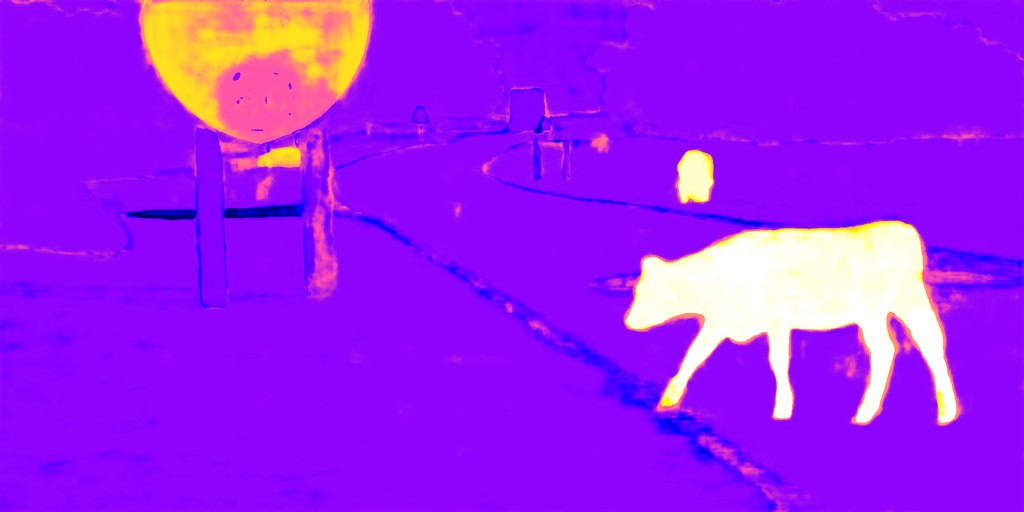}&%
    \includegraphics[trim={0px 0px 0px 0px},clip,width=0.333\textwidth]{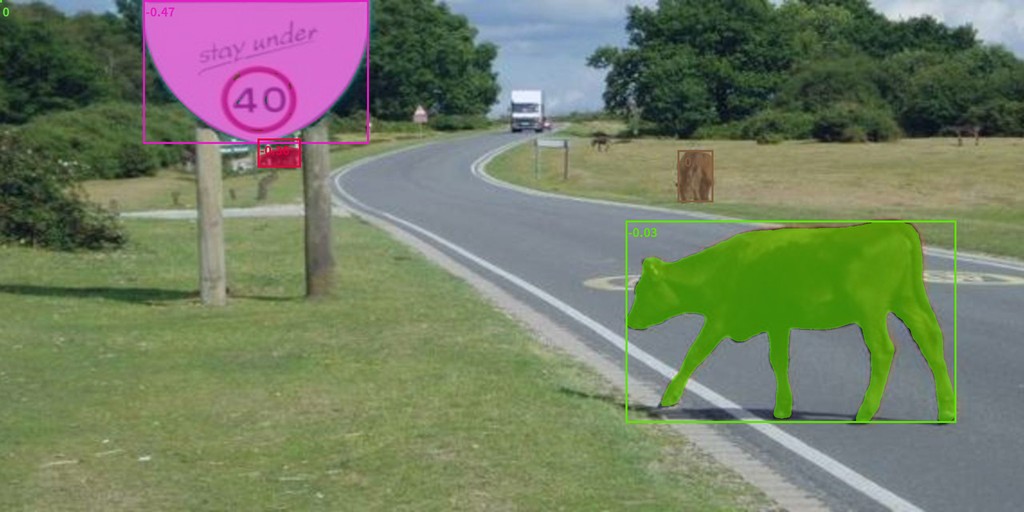}\\%

    \includegraphics[trim={0px 0px 0px 0px},clip,width=0.333\textwidth]{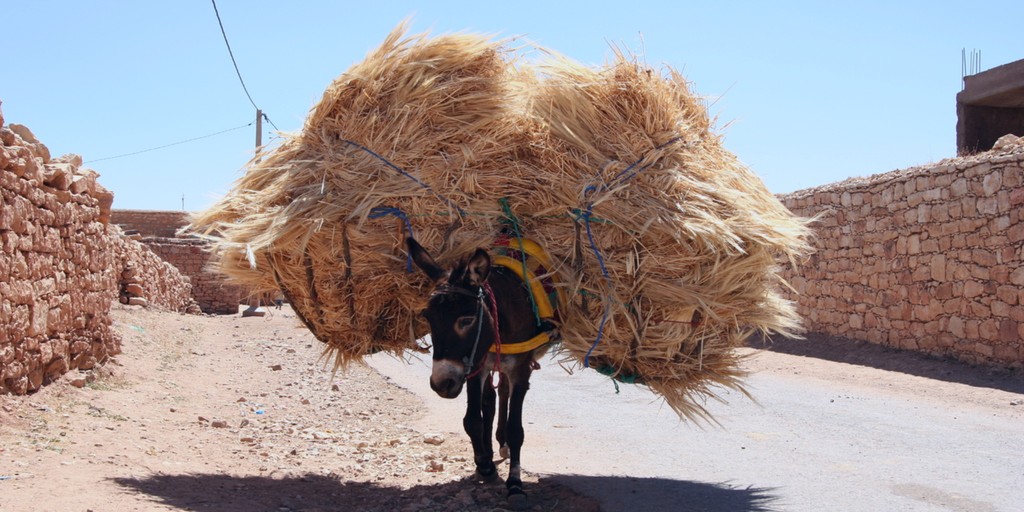}&%
    \includegraphics[trim={0px 0px 0px 0px},clip,width=0.333\textwidth]{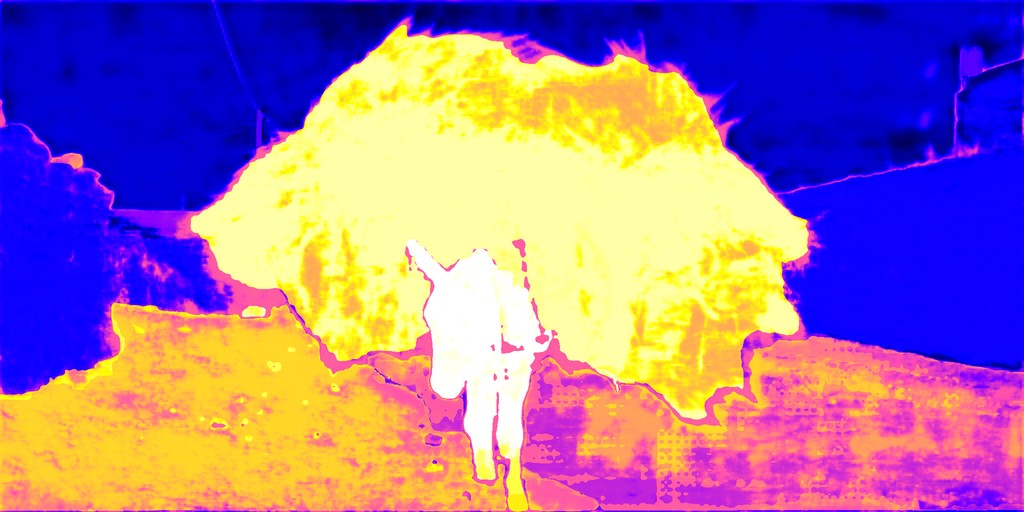}&%
    \includegraphics[trim={0px 0px 0px 0px},clip,width=0.333\textwidth]{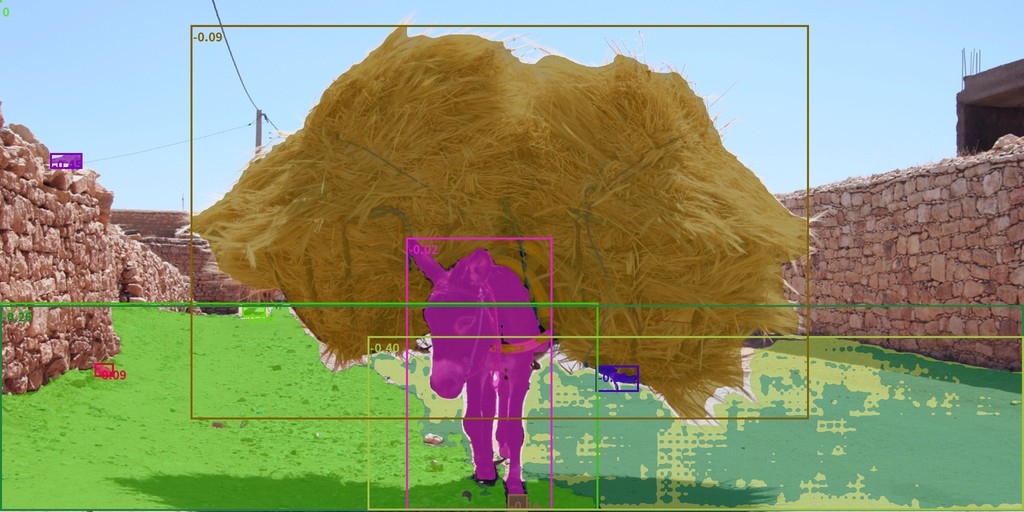}\\%

    \includegraphics[trim={0px 0px 0px 0px},clip,width=0.333\textwidth]{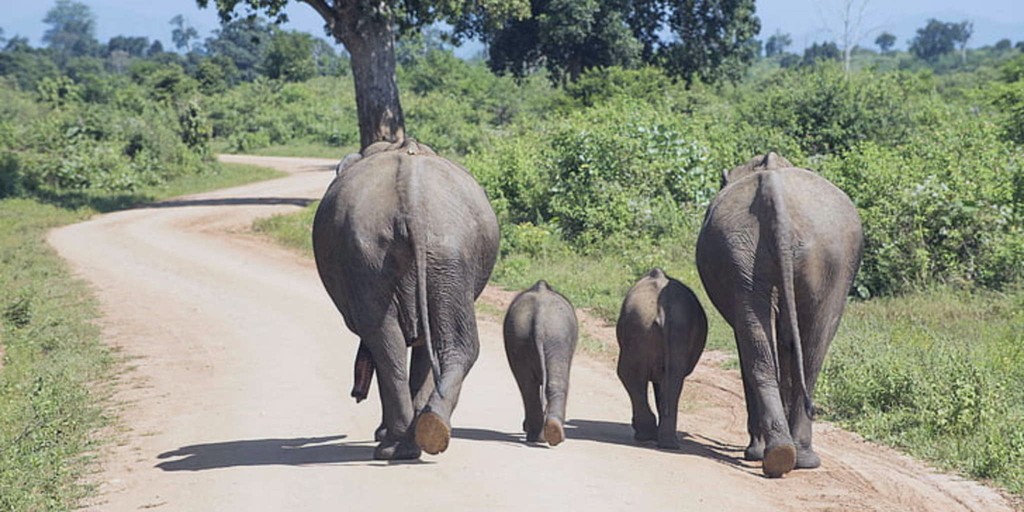}&%
    \includegraphics[trim={0px 0px 0px 0px},clip,width=0.333\textwidth]{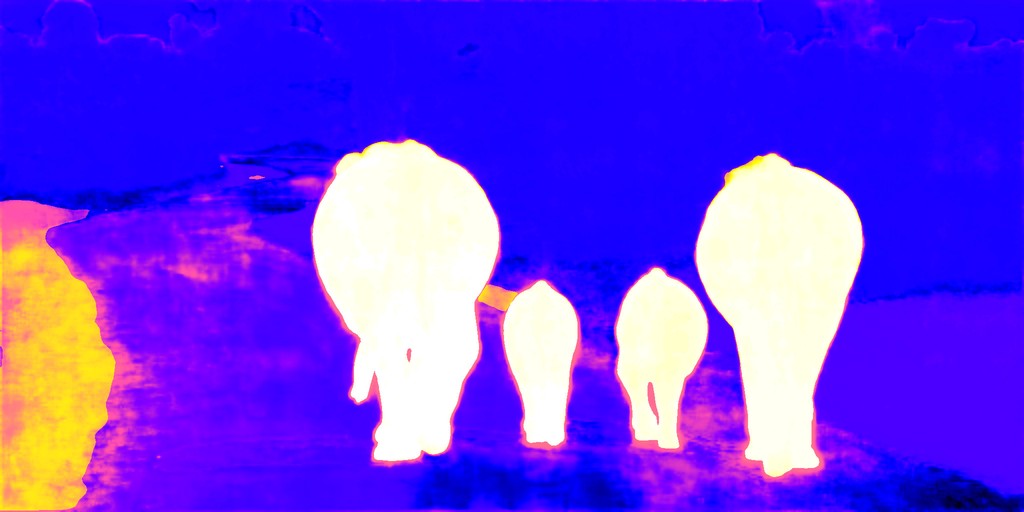}&%
    \includegraphics[trim={0px 0px 0px 0px},clip,width=0.333\textwidth]{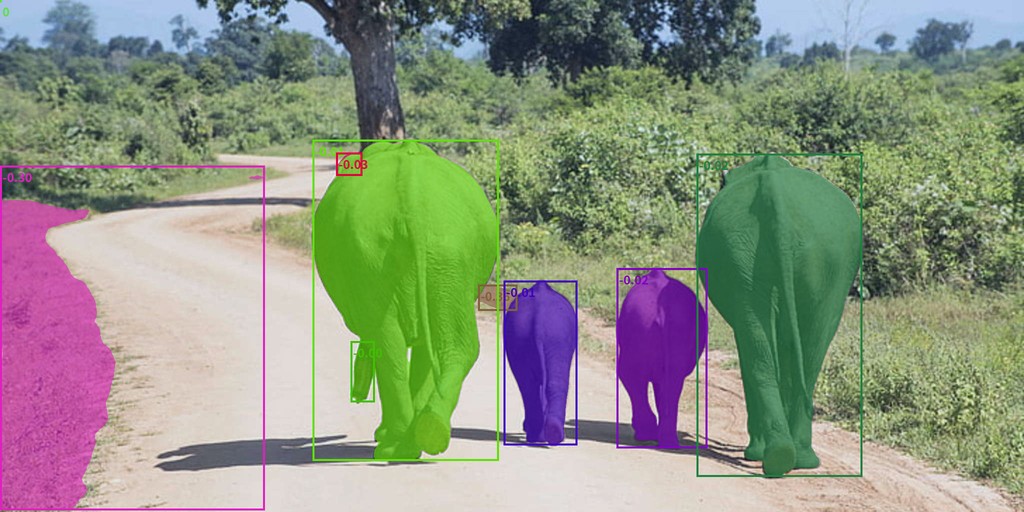}\\%

    \includegraphics[trim={0px 0px 0px 0px},clip,width=0.333\textwidth]{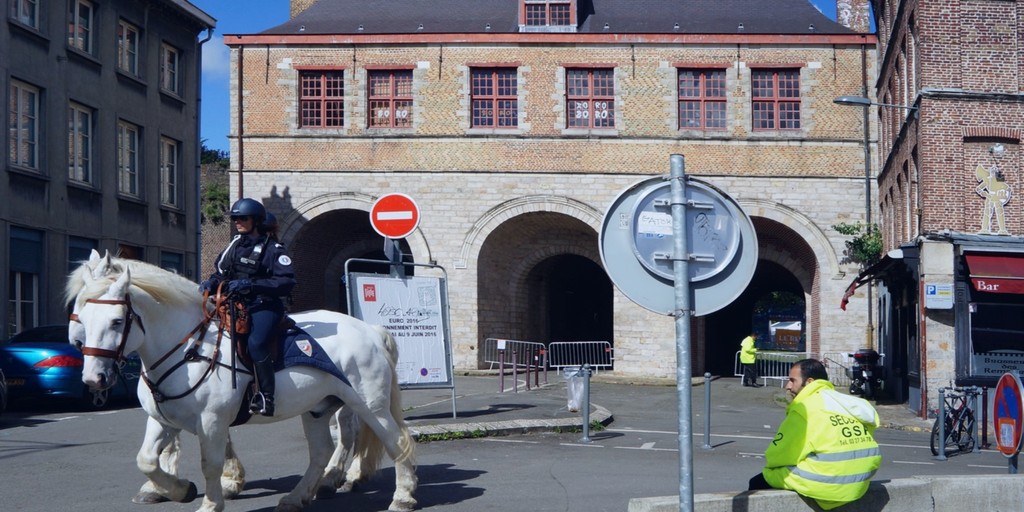}&%
    \includegraphics[trim={0px 0px 0px 0px},clip,width=0.333\textwidth]{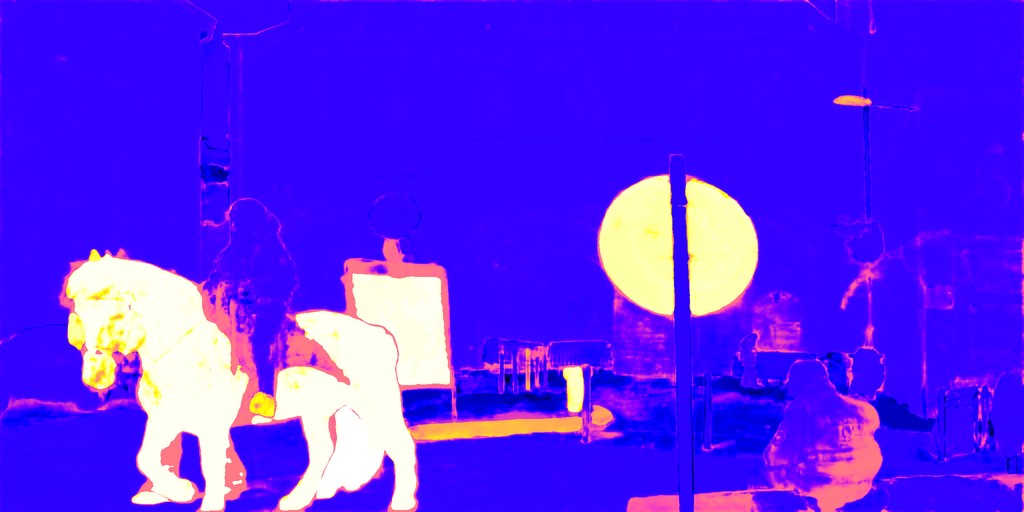}&%
    \includegraphics[trim={0px 0px 0px 0px},clip,width=0.333\textwidth]{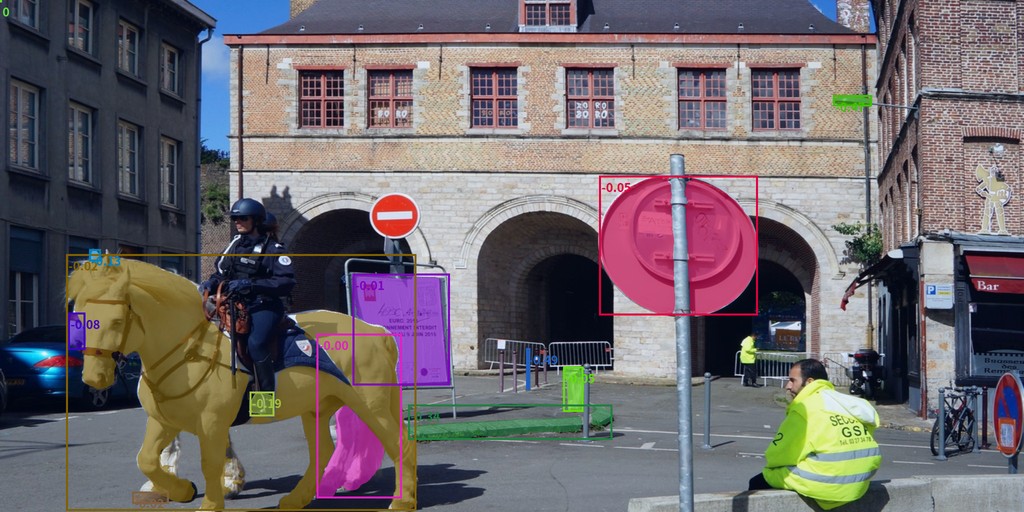}\\%

    \includegraphics[trim={0px 0px 0px 0px},clip,width=0.333\textwidth]{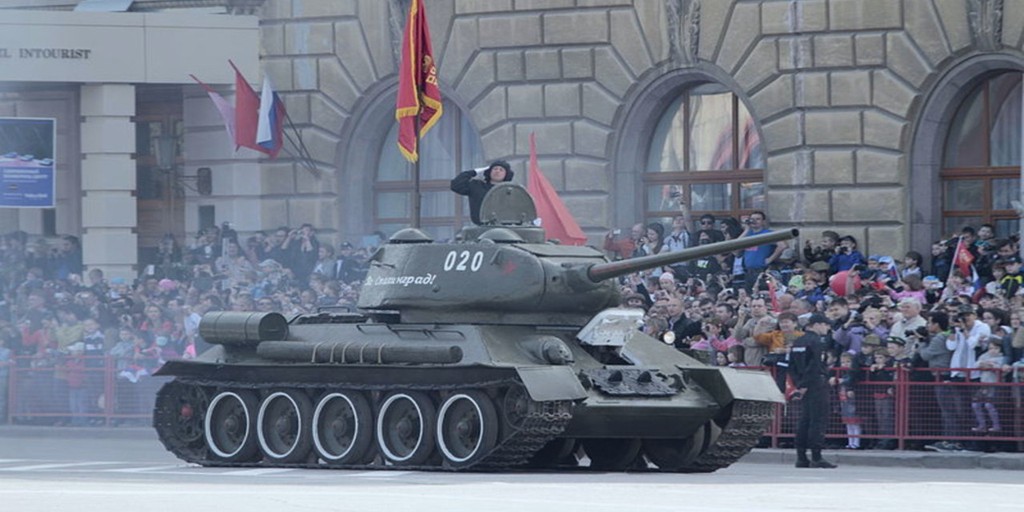}&%
    \includegraphics[trim={0px 0px 0px 0px},clip,width=0.333\textwidth]{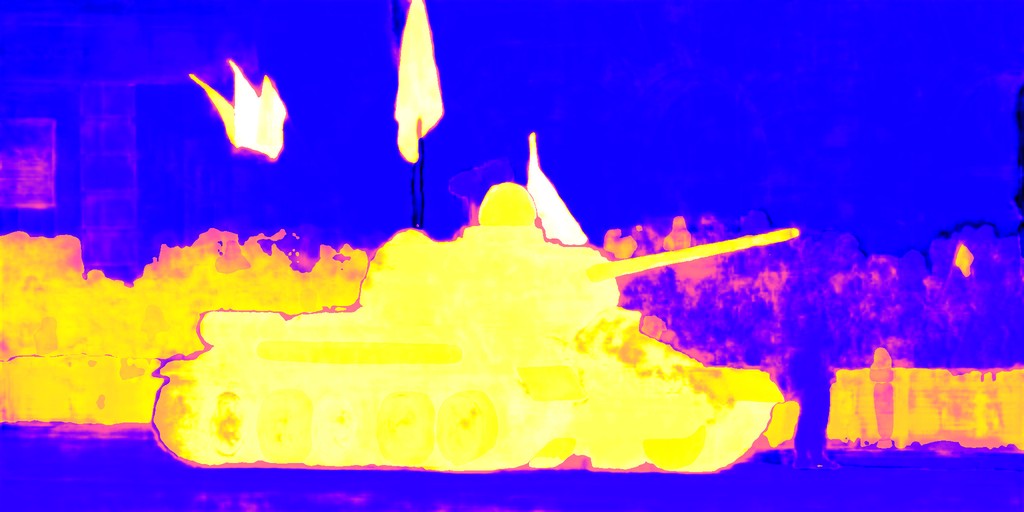}&%
    \includegraphics[trim={0px 0px 0px 0px},clip,width=0.333\textwidth]{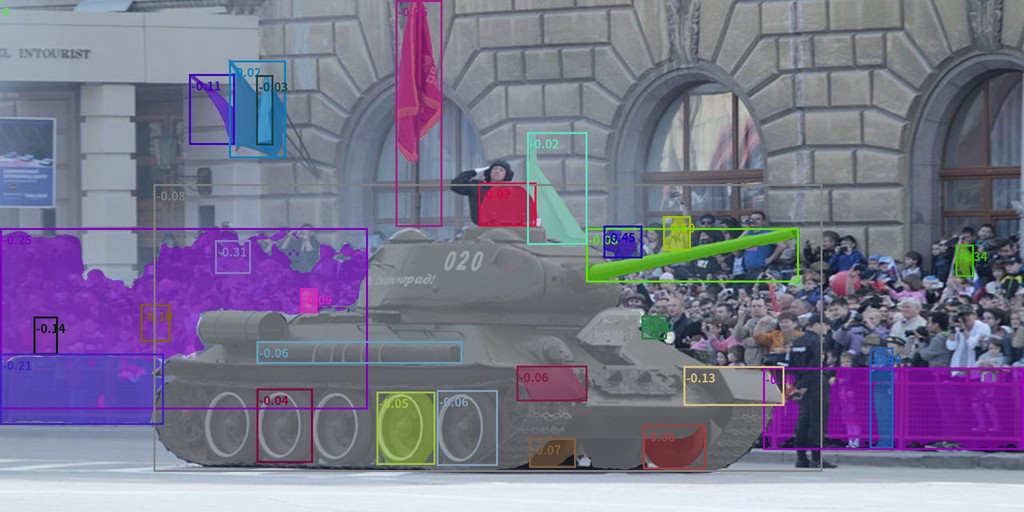}\\%

    \includegraphics[trim={0px 0px 0px 0px},clip,width=0.333\textwidth]{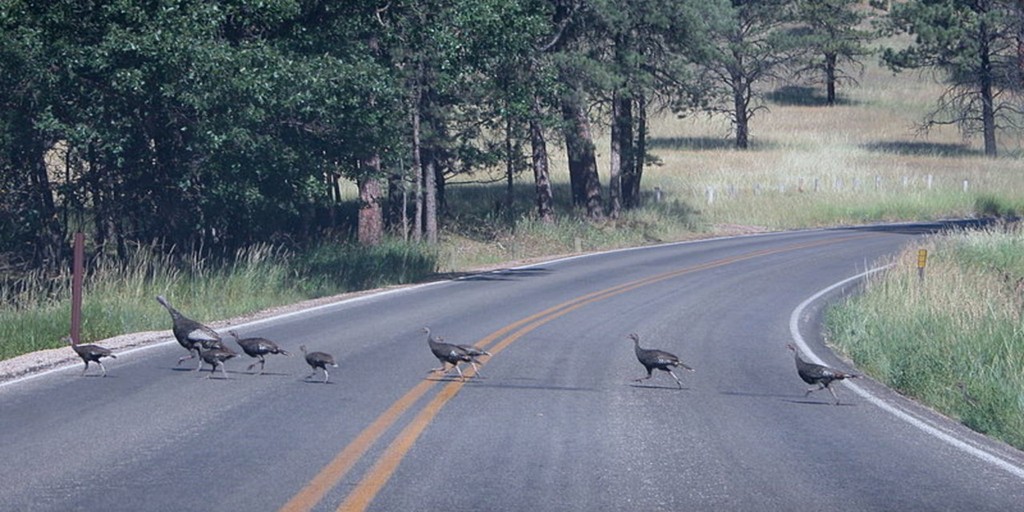}&%
    \includegraphics[trim={0px 0px 0px 0px},clip,width=0.333\textwidth]{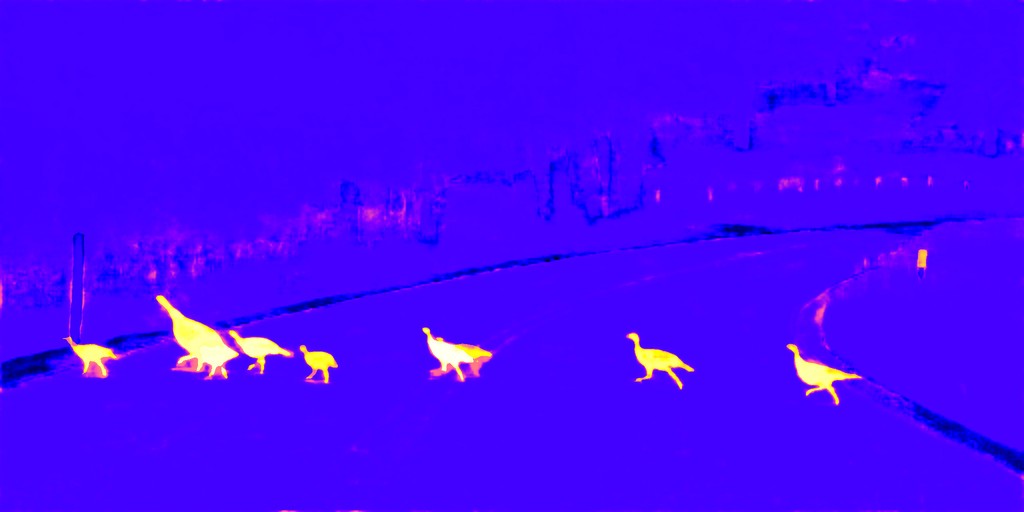}&%
    \includegraphics[trim={0px 0px 0px 0px},clip,width=0.333\textwidth]{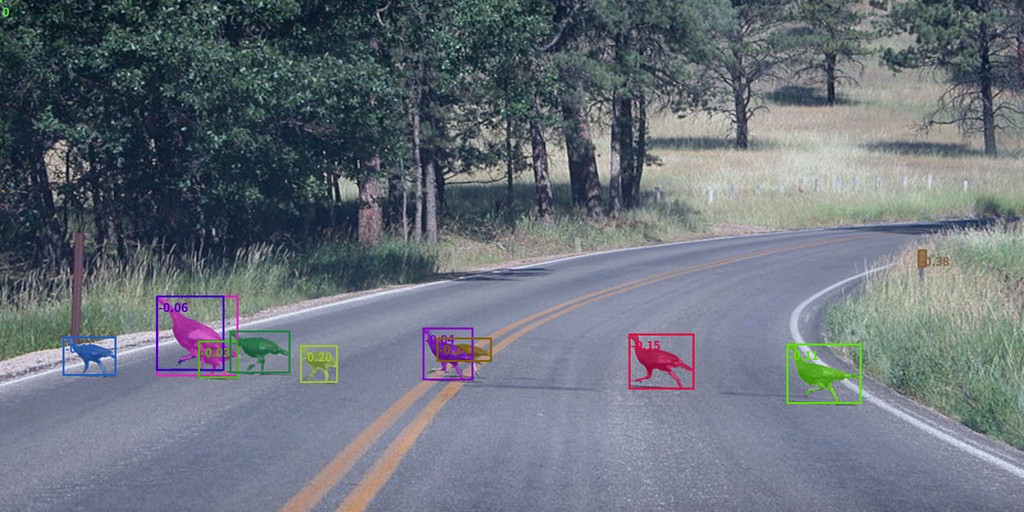}\\%

    Image & Anomaly scores & Instance prediction \\
  \end{tabular}
  \caption{
  \textbf{Qualitative Results} on the SegmentMeIfYouCan dataset \cite{chan2021segmentmeifyoucan}.
  }
  \label{fig:supp_smiyc}
\end{figure}

\vspace{-5pt}
We provide further qualitative results on the CODA \cite{li2022coda} (\reffig{supp_coda}), SegmentMeIfYouCan \cite{chan2021segmentmeifyoucan} (\reffig{supp_smiyc}), and Fishyscapes L\&F \cite{blum2021fishyscapes} (\reffig{supp_fslaf}) datasets.
The Fishyscapes L\&F results show that we can even separate stacked crates, although this negatively affects our instance segmentation performance, since these are annotated as a single instance.

The results on SegmentMeIfYouCan and CODA show more interesting anomalous instances that we segment, and here we can also observe more objects that seem connected due to overlaps.
However, neither of these datasets provide exhaustive instance annotations, so we cannot compute quantitative results there.

\end{document}